UNIVERSITÉ DE GENÈVE
Département d'informatique

FACULTÉ DES SCIENCES
Professeur Stéphane Marchand-Maillet
Docteur Alexandros Kalousis

# Sparse Learning for Variable Selection with Structures and Nonlinearities

## THÈSE

présentée à la Faculté des sciences de l'Université de Genève
pour obtenir le grade de Docteur ès sciences, mention informatique

par

**M**agda GREGOROVÁ

de

Prague, République Tchèque

Thèse No 5318

GENÈVE
2018

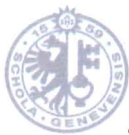

# UNIVERSITÉ DE GENÈVE

**FACULTÉ DES SCIENCES**

# DOCTORAT ÈS SCIENCES, MENTION INFORMATIQUE

## Thèse de Madame Magda GREGOROVA

intitulée :

## «Sparse Learning for Variable Selection with Structures and Nonlinearities»

La Faculté des sciences, sur le préavis de Monsieur S. MARCHAND-MAILLET, professeur associé et directeur de thèse (Département d'informatique), Monsieur A. KALOUSIS, docteur et codirecteur de thèse (Département d'informatique), Monsieur J. MAIRAL, docteur (Institut national de recherche en sciences du numérique Grenoble Rhône-Alpes, Montbonnot, France), Monsieur L. ROSASCO, professeur (Department of informatics, Bioengineering, Robotics and Systems Engineering, University of Genova, Italy), autorise l'impression de la présente thèse, sans exprimer d'opinion sur les propositions qui y sont énoncées.

Genève, le 5 mars 2019

Thèse - 5318 -

**Le Doyen**

# Abstract

In this thesis we discuss machine learning methods performing automated variable selection for learning sparse predictive models. There are multiple reasons for promoting sparsity in the predictive models. By relying on a limited set of input variables the models naturally counteract the overfitting problem ubiquitous in learning from finite sets of training points. Sparse models are cheaper to use for predictions, they usually require lower computational resources and by relying on smaller sets of inputs can possibly reduce costs for data collection and storage. Sparse models can also contribute to better understanding of the investigated phenomenons as they are easier to interpret than full models.

We are specifically interested in problems with non-trivial sparse relationships amongst the data. In particular, problems where the dependencies exhibit some sparse patterns that can be exploited in the modelling but for which the prior understanding is not sufficient to formulate explicit constraints to be hard-wired into the model. We build on the ideas of learning with structured sparsity to factor such patterns into the models.

Furthermore, as the relationships may be too complex to be satisfactorily captured by simple linear functions we allow the methods to operate over a broader space of nonlinear functions. For this we rely on the theory of regularised learning in the reproducing kernel Hilbert spaces (RKHSs) and extend it in the direction of sparse learning in nonlinear nonadditive models.

Throughout the thesis we propose multiple new methods for sparse learning over reduced set of input variables. We initially concentrate on the problem of multivariate time series forecasting and develop methods that learn forecasting models together with discovering the Granger causality dependencies amongst the series.

We first consider dependencies that are organised around a limited set of series functioning as leading indicators for the whole system or its parts. Our method discovers these leading indicators as well as the groups of series depending on them together with learning the predictive model.

We next allow for nonlinear relationships amongst the series. Calling upon the theory of learning vector-valued functions in the RKHS and the ideas of multiple kernel learning we provide the model with enough flexibility to search for the predictive function while still uncovering sparse Granger causal dependencies.

In the second half of the manuscript we focus on the more general problem of learning sparse nonlinear regression functions. Making parallels to linear modelling, we formulate new regularisers based on partial derivatives of the function to promote structured sparsity in the nonlinear model. We show how these can be incorporated into the kernel regression problem and reformulated into a problem solvable in practice by an iterative algorithm derived from the alternating direction method of multipliers (ADMM).

Finally, we address the scalability issues of sparse learning with kernel methods. We use the random Fourier features to approximate the kernel function and shift the sparsity search from the original function space into the space of the random features. We thus significantly reduce the dimensionality of the search space and therefore the computational complexity even when working over large datasets with thousands of data instances.

# Résumé

Dans cette thèse, nous discutons des méthodes d'apprentissage automatique pour la sélection de variables et l'apprentissage de modèles parcimonieux. Nous favorisons les modèles prédictifs parcimonieux pour diverses raisons. En s'appuyant sur un ensemble limité de variables d'entrée, les modèles surmontent naturellement le problème de surapprentissage omniprésent dans l'apprentissage à partir des ensembles finis des instance d'apprentissage. Les modèles parcimonieux sont moins coûteux à utiliser pour la prédiction, ils nécessitent généralement moins de ressources de calcul et, en reposent sur des ensembles de caractéristiques plus petits, ils peuvent éventuellement réduire les coûts de collecte et de stockage des données. Les modèles parcimonieux peuvent également contribuer à une meilleure compréhension des phénomènes étudiés car ils sont plus faciles à interpréter que les modèles complets.

Nous nous intéressons plus particulièrement aux problèmes traitant de relations parcimonieues non triviales entre les données. En particulier, les problèmes où les dépendances présentent des motifs parcimonieux pouvant être exploités dans la modélisation mais pour lesquels la compréhension préalable n'est pas suffisante pour formuler des contraintes explicites à intégrer au modèle. Nous nous appuyons sur les idées d'apprentissage avec une parcimonie structurée pour intégrer ces motifs dans les modèles.

De plus, comme les relations peuvent être trop complexes pour être capturées de manière satisfaisante par de simples fonctions linéaires, nous permettons aux méthodes de fonctionner sur un espace plus large de fonctions non linéaires. Pour cela, nous nous appuyons sur la théorie de l'apprentissage régularisé dans les espaces de Hilbert du noyau reproducteur (RKHS) et nous l'étendons dans le sens d'un apprentissage clairsemé dans des modèles non linéaires non additifs.

Tout au long de la thèse, nous proposons plusieurs nouvelles méthodes d'apprentissage parcimonieuse sur un ensemble réduit de variables d'entrée. Nous nous sommes d'abord concentrés sur le problème de la prévision multivariée de séries temporelles et avons développé des méthodes permettant d'apprendre les modèles de prédiction et de découvrir les dépendances de causalité de Granger parmi les séries.

Nous considérons d'abord les dépendances organisées autour d'un ensemble limité de séries servant d'indicateurs avancés pour tout le système ou ses composants. Notre méthode découvre ces indicateurs avancés ainsi que les groupes de séries qui en dépendent et apprend le modèle prédictif. Nous autorisons ensuite les relations non linéaires entre les séries. En faisant appel à la théorie de l'apprentissage des fonctions vectorielles dans un RKHS et aux idées de l'apprentissage par noyaux multiples, nous donnons au modèle suffisamment de flexibilité pour rechercher la fonction prédictive tout en mettant à jour les dépendances causales de Granger.

Dans la deuxième moitié du manuscrit, nous nous concentrons sur le problème plus général de l'apprentissage des fonctions de régression non linéaires parcimonieuses. En faisant des parallèles avec la modélisation linéaire, nous formulons de nouveaux régulariseurs basés sur des dérivées partielles de la fonction afin de promouvoir la clarté structurée dans le modèle non linéaire. Nous montrons comment ceux-ci peuvent être incorporés au problème de régression du noyau, que nous reformulons en un problème pouvant être résolu en pratique par un algorithme itératif dérivé de la méthode des multiplicateurs de direction alternative (ADMM).

Finalement, nous abordons les problèmes de scalabilité de l'apprentissage parcimonieu avec les méthodes à noyau. Nous utilisons les caractéristiques aléatoires de Fourier pour approximer la fonction du noyau et déplacer la recherche de parcimonie de l'espace de fonctions d'origine vers l'espace des caractéristiques aléatoires. Nous réduisons ainsi de manière significative la dimensionnalité de l'espace de recherche et donc la complexité de calcul même avec de grands ensembles de données avec des milliers d'instances.

# Contents

# Notation

| | |
|---|---|
| $a \; \alpha \; b \; \beta$ | scalars (low-case letters) |
| $\mathbf{a} \; \boldsymbol{\alpha} \; \mathbf{b} \; \boldsymbol{\beta}$ | vectors (bold low-case letters) |
| $\mathbf{A} \; \mathbf{X} \; \mathbf{Y} \; \mathbf{Z}$ | matrices (bold capital letters) |
| $\mathbf{a}_{i.} \; \mathbf{a}_{.j} \; a_{i,j}$ | $i$th row, $j$th column and $(i,j)$ element of matrix $\mathbf{A}$ |
| $\mathbf{A}^T$ | transpose of matrix $A$ |
| $\text{vec}(\mathbf{A})$ | vectorization operator |
| $\text{diag}(\mathbf{A})$ | matrix constructed from the diagonal elements of $\mathbf{A}$ |
| $\text{Tr}(\mathbf{A})$ | trace of the square matrix $\mathbf{A}$ |
| $\odot \; \otimes$ | Hadamard and Kronecker product |
| $\mathbf{x} = (x_1, \ldots, x_n)^T$ | vector (by convention column) and its elements |
| $\mathbb{R} \; \mathbb{R}^m$ | real numbers and $m$-dimensional real vectors |
| $\mathbb{R}_{\geq 0}$ | non-negative real numbers |
| $\mathbb{R}^{m \times n} \; \mathbb{R}_+^{m \times n}$ | $m \times n$ real matrices and non-negative matrices |
| $\mathbb{D}_+^m \; \mathbb{S}_+^m$ | non-negative diagonal and symmetric positive definite matrices |
| $\mathbb{N} \; \mathbb{N}_k$ | positive integers and list of integers $1, \ldots, k$ |
| $\mathcal{H}$ | Hilbert space |
| $\mathcal{F} \; \mathcal{H}_K$ | general function space and reproducing kernel Hilbert space |
| $\mathcal{X} \; \mathcal{Y}$ | input and output space |
| $\mathcal{L}_{\mathcal{Y}}$ | space of bounded linear operators from $\mathcal{Y}$ into itself |
| $\langle \mathbf{a}, \mathbf{x} \rangle$ | inner product of vectors $\mathbf{a}$ and $\mathbf{x}$ |
| $\|\mathbf{a}\|_p$ | $\ell_p$-norm of vector $\mathbf{a}$ |
| $\langle \mathbf{A}, \mathbf{X} \rangle_F := \text{Tr}(\mathbf{A}^T \mathbf{B})$ | Frobenius inner product of matrices $\mathbf{A}$ and $\mathbf{X}$ |
| $\|\mathbf{A}\|_F := \sqrt{\langle \mathbf{A}, \mathbf{A} \rangle_F}$ | Frobenius norm of matrix $\mathbf{A}$ |
| $\langle ., . \rangle_{\mathcal{H}}$ | inner product in the Hilbert space $\mathcal{H}$ |
| $\|\cdot\|_{\mathcal{H}}$ | norm in the Hilbert space $\mathcal{H}$ |
| $\mathbf{1}_K$ | $K$-dimensional vector of ones. |
| $\text{E}(.)$ | expectation of a random variable |

i

# Acronyms

**ADMM** alternating direction method of multipliers

**AR** autoregressive model

**DGP** data generating process

**ERM** empirical risk minimisation

**IID** independently identically distributed

**MKL** multiple kernel learning

**ML** machine learning

**MSE** mean squared error

**OLS** ordinary least squares

**RKHS** reproducing kernel Hilbert space

**RMSE** root mean squared error

**RRM** regularised risk minimisation

**SVM** support vector machine

**VAR** vector autoregressive model

# Chapter 1

# Introduction

This thesis is about learning sparse predictive models for variable selection. What do we understand by learning predictive models? What by sparse models and variable selection? And how do we bring the "structures and nonlinearities" we mention in the title into the modelling?

Learning predictive models is one of the main goals of machine learning (ML). More specifically of *supervised* learning in which we use observed data of some inputs and related outputs to build models that can predict the output values given new inputs. *Sparse models* are in some respect simple, they use smaller sets of parameters or inputs to predict the outputs. We are particularly interested in *variable selection*, therefore the models we build rely on reduced sets of the input variables.

Often times, the dependencies between (and amongst) the input and output data may form some *structures* that may be exploited in the modelling. The outputs may depend on some parts of the inputs, some groups of the input variables, the outputs may themselves form clusters independent of the rest. We build on the ideas of learning with *structured sparsity* to factor such patterns into the models.

Furthermore, the dependencies in the data may often be rather complex. So much so that it may be impossible to capture them via simple linear functions. We therefore extend the class of model hypotheses we use for the sparse learning to *nonlinear functions* through the techniques of learning in reproducing kernel Hilbert spaces (RKHSs).

We elaborate on the rather simplistic answers above in this introductory chapter before investigating some specific problems and proposing new methods to address them in the subsequent text. We introduce the problem of supervised learning and the principles of regularised risk minimisation (RRM) in section 1.1. In section 1.2 we discuss the major ideas of sparse learning and recap some well-known linear methods for learning with structured sparsity. Section 1.3 reviews the theory of learning in RKHSs of possibly nonlinear functions. Finally, we devote section 1.4 to the particularities of multivariate time series forecasting.

The next four chapters contain our main research contributions. In chapters 2 and 3 we focus on the time series forecasting task. We develop new methods that explore and exploit the hidden structures in the system dynamics in order to improve the model forecast performance and to provide the users with better insights into the patterns underlying the system developments. In chapter 2 the method discovers leading indicators for the future evolution and clusters the individual predictive tasks of a linear vector autoregressive model (VAR) around these. The method in chapter 3 allows for nonlinearities when searching for the subset of relevant dynamical dependencies amongst the series in the system using the theory of operator-valued kernels (section 1.3.2).

Chapters 4 and 5 leave aside the time series forecasting question and instead dive deeper into the problem of variable selection when learning general *nonlinear regression* functions. In chapter 4 we propose new derivative-based regularisers to learn models with structured sparsity within the RKHS of nonlinear functions. We address the scalability issues of sparse learning with kernel methods in chapter 5 through the approximating sparse random Fourier features. To conclude we look back at our work from a somewhat broader perspective and suggest possible avenues for future exploration in chapter 6.

This thesis is largely based on the following peer-reviewed conference articles:

- Magda Gregorová, Alexandros Kalousis, Stéphane Marchand-Maillet. (2017) *"Learning Predictive Leading Indicators for Forecasting Time Series Systems with Unknown Clusters of Forecast Tasks."* Asian Conference on Machine Learning (ACML)

- Magda Gregorová, Alexandros Kalousis, Stéphane Marchand-Maillet. (2017)

> *"Forecasting and Granger Modelling with Non-linear Dynamical Dependencies."* European Conference on Machine Learning and Principles and Practice of Knowledge Discovery in Databases (ECML/PKDD)

- Magda Gregorová, Alexandros Kalousis, Stéphane Marchand-Maillet. (2018) *"Structured nonlinear variable selection."* Conference on Uncertainty in Artificial Intelligence (UAI)

- Magda Gregorová, Jason Ramapuram, Alexandros Kalousis, Stéphane Marchand-Maillet. (2018) *"Large-scale Nonlinear Variable Selection via Kernel Random Features."* European Conference on Machine Learning and Principles and Practice of Knowledge Discovery in Databases (ECML/PKDD)

and these shorter workshop papers:

- Magda Gregorová, Alexandros Kalousis. (2015) *"Learning coherent Granger-causality in panel vector autoregressive models."* International Conference on Machine Learning (ICML) Workshop on Demand Forecasting

- Magda Gregorová, Francesco Dinuzzo, Alexandros Kalousis. (2015) *"Functional Learning of Time Series Models Preserving Granger Causality Structures."* Advances in Neural Information Processing Systems (NIPS) Times Series Workshop

## 1.1 Supervised learning

In supervised learning we consider the problem of relating some type of input values to some outputs (responses). The aim of the learning is to find a function (mapping) from the inputs to the outputs that captures well their dependence and that will allow us to automatically assign responses to input values we have not yet seen and for which we do not yet have the output values.

We speak about *learning* because coming up with such a function by handcrafting may not be possible or may be too difficult to be practical. The relationships between the inputs and outputs may not be sufficiently well understood by the domain experts or they may be so complex that when captured fully in an automated system the outputs cannot be produced in a timely manner.

In ML we follow an alternative strategy. We develop algorithms that *learn* the assumed functional relationship from previously observed input and output data.

Formally, we assume that any input $\mathbf{x}$ is contained in some space $\mathcal{X}$ and any output $\mathbf{y}$ in some space $\mathcal{Y}$. Our aim is to learn a predictive function $\mathbf{f} : \mathcal{X} \to \mathcal{Y}$ whose output $\mathbf{f}(\mathbf{x})$ approximates well the true output $\mathbf{y}$ for an arbitrary input $\mathbf{x} \in \mathcal{X}$. For this we use a set of observed input-output data pairs $\mathcal{S}_n = \{(\mathbf{x}_i, \mathbf{y}_i) \in (\mathcal{X} \times \mathcal{Y}) : i \in \mathbb{N}_n\}$. The set $\mathcal{S}_n$ is usually called a *training set* and the subscript $n$ indicates its size.

To be able to learn from the training set a function that predicts well over the yet unseen data, the set has to have something in common with the new observations we may get in the future. This idea, ubiquitous in statistical learning, is captured in the assumptions on the generating process of the data. These are formulated using basic notions of probability theory which we now briefly summarise.

Let $(\Omega, \mathcal{A}, \mathrm{P})$ be a probability space where $\Omega$ is the event sample space, $\mathcal{A}$ is its associated sigma algebra, and $\mathrm{P}$ is the probability measure $\mathrm{P} : \mathcal{A} \to [0, 1]$. A $\mathcal{Y}$-valued *random variable* is the function $y : \Omega \to \mathcal{Y}$ with the probability measure (distribution) $P(y \in a) = \mathrm{P}(\{\omega \in \Omega : y(\omega) \in a\})$ where $a \subseteq \mathcal{Y}$. In most cases the space $\mathcal{Y}$ is a subset of the multidimensional real space $\mathbb{R}^m$. A *multivariate real random variable* (random vector) is the vector-valued function $\mathbf{y} : \Omega \to \mathcal{Y} \subseteq \mathbb{R}^m$ with components the scalar-valued random variables $y_i : \Omega \to \mathbb{R}$, $i \in \mathbb{N}_m$ defined over the same probability space. The measure $P(\mathbf{y} \in \mathbf{a}) = \mathrm{P}(\{\omega \in \Omega : y_i(\omega) \in a_i, i \in \mathbb{N}_m\})$, where $\mathbf{a} = (a_1, \ldots a_m)^T \subseteq \mathcal{Y}$, is the joint probability distribution of $\mathbf{y}$. Clearly, the univariate real-valued random variable $y$ is a special case of the multivariate $\mathbf{y}$ with $m = 1$. We indicate both the random variables (the functions $\mathbf{y}$) and their realizations (their values at a specific point $\mathbf{y}(\omega)$) simply by the lower case $\mathbf{y}$ and clarify in the text which of these we mean when this could cause confusion.

The assumption for the input-output pairs $(\mathbf{x}, \mathbf{y})$ observed in the past as well as in the future is that these are realizations of random variables generated *independently* from the same, though unknown, joint probability distribution $P_{x,y}$ over the space $\mathcal{X} \times \mathcal{Y}$. In short, they are independently identically distributed (IID). First, the input $\mathbf{x}$ is generated from the marginal distribution $P_x$ on $\mathcal{X}$.

Second, the output $\mathbf{y}$ is generated from the conditional distribution $P_{y|x}$ on $\mathcal{Y}$ for the fixed value $\mathbf{x}$ of the input.

The randomness in generating the inputs according to the unknown $P_x$ reflects our inability to control what input values we have in the observed training set and will get in the future. Specific methods have been developed for cases when we either can control which inputs to query during the training (*active learning*), e.g. Settles (2012), or we know beforehand for which inputs we need to produce the responses (*transductive learning*), e.g. Gammerman et al. (1998). These are sometimes categorised as *semi-supervised* learning methods and we will not discuss them here any further.

The conditional distribution $P_{y|x}$ accommodates the fact that the inputs may not be associated with unique outputs in a deterministic manner. In general, this distribution is unknown to us as well representing our ignorance about the true relationships between the inputs and the outputs or our inability to capture them formally.

In this thesis we primarily discuss the supervised *regression* problem in which the inputs and outputs are real-valued $\mathbf{x} \in \mathcal{X} \subseteq \mathbb{R}^d$, $\mathbf{y} \in \mathcal{Y} \subseteq \mathbb{R}^m$. When $m > 1$ this is usually referred to as *multi-output regression*.

### 1.1.1 Regularised risk minimisation

In the supervised learning problem our objective is to find a function $\mathbf{f} : \mathcal{X} \to \mathcal{Y}$ that can approximate (*predict*) well the true outputs $\mathbf{y}$ given the inputs $\mathbf{x}$. In principle, any function that produces predictions $\mathbf{f(x)} \in \mathcal{Y}$ for all values of the inputs $\mathbf{x} \in \mathcal{X}$ is valid. To be able to assess whether such a function is any good we need some measure of its quality. From a broader perspective, we are also interested in measuring the quality of the algorithm that produced the specific predictive function. This is what *statistical learning theory* aims to analyse.

The starting point of the analysis is a measure of quality of an individual prediction when compared to the true output value, the *loss function* $L : \mathcal{Y} \times \mathcal{Y} \to \mathbb{R}_{\geq 0}$. For a given input-output pair the value of the loss function $L(\mathbf{y}, \mathbf{f(x)})$ shall be smaller if the predicted output $\mathbf{f(x)}$ is better. There are many possible forms of loss functions defining what this "better" means that may be more or less appropriate for particular problems. For the classical regression problem, unless there are some domain-specific reasons to formulate some specialised loss (e.g. asymmetric), the habitual choice is the *squared error loss*

$$\mathrm{L}(\mathbf{y}, \mathbf{f}(\mathbf{x})) = \|\mathbf{y} - \mathbf{f}(\mathbf{x})\|_2^2 \ . \tag{1.1}$$

Obviously, it is not enough to know how the learned function $\mathbf{f}$ performs on a single data point $(\mathbf{x}, \mathbf{y})$. What we are interested in is how good the function is for an arbitrary input-output pair $(\mathbf{x}, \mathbf{y}) \in \mathcal{X} \times \mathcal{Y}$, what is its *expected loss* (usually called the *expected risk*) over the generative distribution $P_{x,y}$

$$\mathcal{R}_P(\mathbf{f}) = \mathrm{E}\left[\mathrm{L}(\mathbf{y}, \mathbf{f}(\mathbf{x}))\right] = \int \mathrm{L}(\mathbf{y}, \mathbf{f}(\mathbf{x})) \, \mathrm{d}P_{x,y} \ . \tag{1.2}$$

A natural objective for learning a good predictive function is now the minimisation of the expected risk

$$\mathbf{f}^* = \underset{\mathbf{f}}{\arg\min} \, \mathcal{R}_P(\mathbf{f}) \ . \tag{1.3}$$

The minimal possible risk $\mathcal{R}_P^* = \mathcal{R}_P(\mathbf{f}^*)$ over all measurable functions is called the *Bayes risk* and the associated minimiser $\mathbf{f}^*$ the *Bayes predictive function*. It is rather easy to show (proof in the appendix) that in the regression case with the squared error loss (1.1) the Bayes predictive function is the conditional expectation $\mathbf{f}^*(\mathbf{x}) = \mathrm{E}[\mathbf{y}|\mathbf{x}] = \int \mathbf{y} \, \mathrm{d}P_{y|x}$.

Unfortunately, as explained in section 1.1 the joint as well as the conditional distributions above are both unknown and therefore we cannot evaluate the expectation in equation (1.2) nor the conditional $\mathrm{E}[\mathbf{y}|\mathbf{x}]$. Instead of the expected risk (1.2) we therefore may want to resort to using its sample estimate, the *empirical risk*

$$\mathcal{R}_n(\mathbf{f}) = \frac{1}{n} \sum_{i=1}^{n} \mathrm{L}(\mathbf{y}_i, \mathbf{f}(\mathbf{x}_i)) \tag{1.4}$$

calculated over the training data set $\mathcal{S}_n$. While by application of the *law of large numbers* the empirical risk for a fixed $\mathbf{f}$ converges as $n \to \infty$ to the expected risk defined in equation (1.2), the risk of the predictive function chosen by minimising the empirical risk does not in general converge to the Bayes risk.

For finite training samples $n \ll \infty$ the predictive function chosen by minimising the empirical risk will in general have larger expected risk than the Bayes risk of the Bayes predictive function $\mathbf{f}^*$ from equation (1.3). This is due to it focusing too much on fitting the training data including the specific sample random noise, a phenomenon known in ML as *overfitting*. In the extreme, if the learning algorithm were to have absolute freedom in choosing the predictive function, it could fit any training data perfectly reaching zero empirical risk. This would, however, say very little about how the learned predictive function performs on data outside the training set, how it *generalizes*.

A standard ML strategy, e.g. Mohri et al. (2012), for preventing overfitting and hence improving generalization, the so called *empirical risk minimisation (ERM)*, is in suppressing the flexibility of the learning algorithm by restricting the *hypothesis space* of functions within which the algorithm searches to a limited set or class of functions $\mathbf{f} \in \mathcal{F}$

$$\widehat{\mathbf{f}} = \underset{\mathbf{f} \in \mathcal{F}}{\arg\min} \, \mathcal{R}_n(\mathbf{f}) \ . \tag{1.5}$$

For example, the classical linear regression model narrows the functional class $\mathcal{F}$ to linear functions.

Choosing a priori the class of functions so that it has sufficient (not to *underfit*) yet not too large (not to *overfit*) capacity is a difficult if not an impossible task. An alternative approach, *regularised risk minimisation (RRM)*, gives the flexibility to the algorithm by choosing a relatively large hypothesis space $\mathcal{F}$ but combining the empirical loss with a regularisation term $R(\mathbf{f})$ that penalizes more complex hypotheses

$$\widehat{\mathbf{f}} = \underset{\mathbf{f} \in \mathcal{F}}{\arg\min} \, \frac{1}{n} \sum_{i=1}^{n} L(\mathbf{y}_i, \mathbf{f}(\mathbf{x}_i)) + \lambda R(\mathbf{f}) \ . \tag{1.6}$$

In the above, $\lambda$ controls the weight given to the regularisation term $R$ versus the empirical loss term $L$ in searching for the optimal $\mathbf{f}$.

The RRM is a fundamental concept providing the major theoretical background and the starting point for all methods developed in this thesis. We explore various forms of the regularisation term $R$ to achieve better predictive perfor-mance in combination with learning models with (structured) sparsity.

## 1.2 Sparse learning for variable selection

One of the reasons for the raise in popularity of machine learning in comparison to traditional statistical approaches is the focus on developing methods that can process large amounts of data. Lately the availability of data does not seem to be so much of a problem. Data are everywhere, they are collected, processed and stored in amounts and speeds that were hard to imagine when traditional statistical methods have been developed years or even decades ago.

However, everything is not as bright as it may seem. Yes, there are huge amounts of data out there and after overcoming technical, proprietary or other similar obstructions, they can be used for analysis and developing models. Yet, the paramount question which remains to be answered is: Are these data *relevant* for the problem we want to investigate?

We discussed in the previous section the motivation for learning a model from the observed data as opposed to handcrafting it based on some expert knowledge. Now that we have decided to use the techniques of supervised learning, we need to present the algorithm with suitable training examples. What shall these be? What inputs shall the model use to predict the outputs? What inputs does it really need and which can be discarded as irrelevant? Once again the specific domain understanding is often not sufficient to answer these question satisfactorily before commencing the learning.

A natural solution is to provide the learning algorithm with all the data that might be relevant and hope for the best. "The more the better", we may think naively. However, doing so we make the problem for the learning algorithm more difficult. Without knowing where to focus the algorithm has too much space to search through for a reasonable model. In the high-dimensional setting, where the number of input dimensions may be much bigger than the total number of examples in the training ($d \gg n$), it is easy to come up with pathological solutions that seem perfect for the training set yet are useless for predicting over new data.

As explained in section 1.1.1, a reasonable strategy to prevent overfitting the training data is in reducing the hypothesis search space. This leads onto the idea of model *parsimony*, model *sparsity* in terms of using the lowest possible number of input variables to achieve good predictions. In this context, sparse learning methods capable of automated *variable selection* (sometimes referred to as *feature selection* though, as explained below, somewhat ambiguously) constitute an indispensable part of the modern machine learning toolkit.

Overfitting is not the only reason for targeting model sparsity (and in fact many other notions of model simplicity can be used to guide the hypothesis space reduction). Sparse models are also more interpretable and, with some caution, can contribute to better understanding of the investigated phenomenons. They are cheaper to use for predictions, they usually require lower computational resources, and they rely on smaller sets of inputs possibly reducing the costs for data collection and storage.

The term *model sparsity* is somewhat ambiguous in the machine learning literature. In addition to sparsity in the input variables, models can be sparse in the number of input examples they use (such as in the support vector machines in which the predictive model is a function of a limited number of support vectors, e.g. Cortes and Vapnik (1995)) or in the number of hidden features they construct and select (such as in the sparse multiple kernel learning, e.g. Lanckriet et al. (2004), or in sparse autoencoders, e.g. Ranzato et al. (2008)). For us in this thesis, model sparsity shall always be understood as sparsity in the input variables as we introduced it above.

### 1.2.1 Variable selection approaches

"The problem of determining the best subset of variables has long been of interest to applied statisticians and, primarily because of the current availability of high-speed computations, this problem has received considerable attention in the recent statistical literature". The previous is a quote from Hocking's 1976 review of the state of the art on variable selection methods in linear regression (Hocking 1976). Clearly, the problem of variable selection for regression modelling has been stirring the minds of data scientists for many years now. Many of the ideas developed in the early years of research on this topic prevailed and we can find them in various flavours in nowadays practice (such as stepwise selection heuristics).

Modern automated approaches for variable selection address (more or less successfully) the shortcomings of these initial attempts going beyond the simple linear modelling and considering more complex problems (e.g. larger dimensionality, structure). Based on how the selection methods interact with the model learning algorithm, they can principally be categorised into three broad classes (Blum and Langley 1997): *filters*, *wrappers* and *embedded* methods.

The *filter* methods view the variable selection and model learning as two unrelated tasks. The filtering of the variables is conducted as a preprocessing step the results of which can be subsequently passed onto an arbitrary model learning algorithm. The criteria for the variable filtering are typically based on various statistical or information theoretic measures of dependency such as the Fisher score (Koller and Sahami 1996) or the mutual information (Vergara and Estévez 2014). Despite their name, usually the methods actually do not perform any filtering. Rather, they rank the variables by the adopted criteria and the relevant subset is then selected based on some threshold. The advantage of the filter methods is that they can be relatively fast even when operating over large sets of input dimensions. The disadvantage is that the variable search is disconnected from the final objective of learning a predictive function with high predictive accuracy.

The *wrappers* treat the learning algorithm as a black box used within the process of the variable selection to obtain a measure of the selection accuracy. Most often, this measure is the predictive performance of the learning algorithm over the training or an independent validation set. The methods typically define a heuristic for the search within the $2^d$ space of all possible variable subsets. The classical examples mentioned already in Hocking (1976) are the stepwise forward selection or backward elimination. A more recent example of this approach is the recursive elimination heuristic applied over the support vector machine (SVM) algorithm for gene selection in cancer prediction (Guyon et al. 2002). Unlike filters, the wrappers link the variable selection to the predictive accuracy of the underlying learning algorithm. They typically also produce directly a variable subset instead of just variable ranking. On the other hand, depending on the complexity of the underlying learning algorithm they can be rather expensive, especially for large dimensions, as they need to execute the learning step for every candidate variable subset.

The last group, the *embedded* methods, interweave the variable selection with the predictive model learning regarding them together as one problem consisting of two related objectives. The prime example of these are the regularisation methods for learning sparse linear models (Hastie et al. 2015). These methods build directly on the ideas of RRM discussed in section 1.1.1. They formulate regularisers that promote sparsity with respect to the model input variables and combine them with the empirical loss minimisation for learning the predictive function. All the methods we develop in the subsequent chapters of this thesis fall into this category.

### 1.2.2 Regularisation methods for sparsity in linear models

Linear regression models restrict the hypothesis class $\mathcal{F}$ to linear functions of the inputs $\mathbf{x} \in \mathbb{R}^d$. In case of the usual single-dimensional output space $\mathcal{Y} \subseteq \mathbb{R}$ the function is of the form

$$\mathbf{f}(\mathbf{x}) = \langle \mathbf{x}, \mathbf{w} \rangle = \sum_{a=1}^{d} x_a \, w_a \ , \tag{1.7}$$

where $\mathbf{w}$ is the $d$-dimensional vector of the model parameters that the algorithm needs to learn. Variable selection in this case corresponds to zeroing some of the parameters $w_a = 0$, $a \in \mathbb{N}_d$.

The classical ordinary least squares (OLS) approach to learning linear regression models minimises the nonregularised empirical least squares loss

$$\widehat{\mathbf{w}}_O = \arg\min_{\mathbf{w}} \frac{1}{n} \sum_{i=1}^{n} \left( y_i - \sum_{a=1}^{d} x_{ia} \, w_a \right)^2 \ . \tag{1.8}$$

The estimated vector $\widehat{\mathbf{w}}_O$ is typically nonsparse with all the elements nonzero $\widehat{w}_a \neq 0$. In a high-dimensional setting the OLS approach is prone to overfit, it is very sensitive to small changes in the training data, and in the overparmatrised case $d > n$ it does not even have a unique solution (we need a heuristic to pick one).

To abate the above problems the standard learning theory (see section 1.1.1) suggests to restrict the learning to even smaller class of hypotheses. A possible strategy is to learn the function over only a subset of the input variables so that the dimensionality of the reduced input space $\widetilde{\mathcal{X}} \subseteq \mathbb{R}^l$ is much smaller, $l \ll d$, than of the original space $\mathcal{X} \subseteq \mathbb{R}^d$. This idea leads onto the filter and wrapper approaches to variable selection discussed in the previous section.

Another strategy is to constrain the model parameters onto an $\ell_2$-ball (Hoerl and Kennard 1970)

$$\widehat{\mathbf{w}}_R = \arg\min_{\mathbf{w}} \frac{1}{n} \sum_{i=1}^{n} \left( y_i - \sum_{a=1}^{d} x_{ia} w_a \right)^2, \quad \text{s.t. } \|\mathbf{w}\|_2^2 \leq c \ . \tag{1.9}$$

In the Lagrange form this is equivalent to

$$\widehat{\mathbf{w}}_R = \arg\min_{\mathbf{w}} \frac{1}{n} \sum_{i=1}^{n} \left( y_i - \sum_{a=1}^{d} x_{ia} w_a \right)^2 + \lambda \|\mathbf{w}\|_2^2 \ . \tag{1.10}$$

Equation (1.10) is probably the most famous regularised method in machine learning known as the *ridge regression* (Hoerl and Kennard 1970). It reduces the hypotheses search space by shrinking the learned parameters $\mathbf{w}$. It does not, however, set any of the dimensions of the parameter vector to zero and therefore does not perform any variable selection.

A series of papers in the 90's recognised the advantage of combining the two strategies above into one. The soft-thresholding of Donoho and Johnstone 1994, the nonnegative garotte of Breiman (1995), the lasso of Tibshirani (1996) and the basis pursuit of Chen et al. (1998), all used parameter shrinking to achieve variable selection.

The success of these methods triggered further research into the regularisation approach for variable selection. A multitude of methods have been developed since, often introducing some structure into the subset selection. The group lasso of Yuan and Lin (2006), group lasso with overlap, graph lasso and hierarchical lasso of Zhao et al. (2009) and Jacob et al. (2009) operate over some predefined groups of input variables rather than treating them individually. The elastic net of Zou and Hastie (2005) and fused lasso of Tibshirani et al.

(2005) tend to select correlated or similar variables together. Argyriou et al. (2008) and Obozinski et al. (2009) suggest to select common variables across multiple prediction tasks. And there are many more, see e.g. Bach et al. (2012) or Hastie et al. (2015) for a review of the most important ones. We briefly summarise here below the problem formulations most relevant to our work.

**Lasso**

The lasso method of Tibshirani (1996) is supposedly the best known of all sparse machine learning techniques. The problem formulation is very similar to the ridge regression in equations (1.9) and (1.10). It promotes sparsity in the solution by replacing the squared $\ell_2$-norm $\|\mathbf{w}\|_2^2 = \sum_a^d |w_a|^2$ over the parameters with the non-smooth $\ell_1$-norm $\|\mathbf{w}\|_1 = \sum_a^d |w_a|$. In the Lagrange form of the RRM the lasso problem is

$$\widehat{\mathbf{w}}_L = \arg\min_{\mathbf{w}} \frac{1}{n} \sum_{i=1}^n \left( y_i - \sum_{a=1}^d x_{ia} w_a \right)^2 + \lambda \|\mathbf{w}\|_1 \quad . \tag{1.11}$$

The sparsity in the lasso solution emerges due to the singularities of the non-smooth $\ell_1$ norm as opposed to the smooth shrinking of the ridge squared $\ell_2$ penalty (constraint). The original paper provides some graphics to help the intuition for this effect.

**Group lasso**

Lasso has the ability to select individual variables. The aim of group lasso (Yuan and Lin 2006) is to select (or reject) the variables in a priori defined groups. For $J$ groups of input variables each of cardinality $p_j$, $j \in \mathbb{N}_J$ we indicate by $\mathbf{x}^{(j)}$ the $j$th group of inputs and by $\mathbf{w}^{(j)}$ the corresponding vector of parameters. The group lasso learns the model parameters by solving the RRM

problem[1]

$$\widehat{\mathbf{w}}_{GL} = \arg\min_{\mathbf{w}} \frac{1}{n} \sum_{i=1}^{n} \left( y_i - \sum_{j=1}^{J} \left\langle \mathbf{x}_i^{(j)}, \mathbf{w}^{(j)} \right\rangle \right)^2 + \lambda \sum_{j=1}^{J} \sqrt{p_j} \left\| \mathbf{w}^{(j)} \right\|_2 . \quad (1.12)$$

Note the lack of the square on the $\ell_2$-norms of the group parameters in the penalty as compared to the ridge in equation (1.10). The group lasso penalty is sometimes indicated as $\ell_1/\ell_2$ as it combines the $J$-long vector of the $\ell_2$-norms calculated over the group parameters through an $\ell_1$-norm using the same singularity effects of the non-smooth $\ell_1$ as in the lasso applying it onto the parameter groupings.

**Elastic net**

The elastic net (Zou and Hastie 2005) addresses one of the major drawbacks of lasso: in the presence of highly correlated input variables it tends to select fairly arbitrarily just one of these and drop the other ones from the model. This is undesirable for multiple reasons. The selection is highly unstable, lasso will pick different variables with small changes in the training data. The interpretability of the results is questionable, all of the correlated variables may be of interest. Empirically, the prediction accuracy of lasso in these cases is often less than that of ridge regression. In the elastic net, the authors propose to combine the lasso with the ridge penalty to benefit from the desirable properties of both

$$\widehat{\mathbf{w}}_{EN} = \arg\min_{\mathbf{w}} \frac{1}{n} \sum_{i=1}^{n} \left( y_i - \sum_{a=1}^{d} x_{ia} w_a \right)^2 + \lambda_1 \|\mathbf{w}\|_1 + \lambda_2 \|\mathbf{w}\|_2^2 . \quad (1.13)$$

The elastic net maintains the selection ability through the non-smooth $\ell_1$-norm. At the same time, due to the $\ell_2$-norm it exhibits a grouping effect in selecting strongly correlated variables together and in keeping the parameter values of highly correlated variables close to each other. On the other hand, the authors argue for the need of a final de-biasing step to remove the effect of the double shrinkage of the parameter estimates resulting from the double regularisation. Similar de-biasing procedures can often improve the predictive performance of sparse shrinkage methods, e.g. Rosasco et al. (2013).

---

[1]The authors in the original paper proposed a more general formulation of the group lasso penalty. Our formulation here is the one they finally used in their implementation and which has since become the canonical form when referring to group lasso.

**Multi-task lasso**

The multi-task lasso (Obozinski et al. 2009) focuses on the problem of learning models for multiple related predictive tasks. Each of the $m$ tasks has a different output (response) variable $y^{(j)} \in \mathcal{Y}^{(j)} \subseteq \mathbb{R}$, $j \in \mathbb{N}_m$, however, they all share the same space of the input variables $\mathbf{x}^{(j)} \in \mathcal{X} \subseteq \mathbb{R}^d$, $j \in \mathbb{N}_m$. In principle, each of the $m$ tasks can come with its own set of input-output pairs $\mathcal{S}^{(j)} = \{(\mathbf{x}_i^{(j)}, y_i^{(j)}) \in (\mathcal{X} \times \mathcal{Y}^{(j)}) : i \in \mathbb{N}_{n^{(j)}}\}$. The assumption of the method is that while the tasks may have very different models in terms of the parameter values $\mathbf{w}^{(j)}$, they all share the same set of relevant input variables. The multi-task lasso solves the optimisation problem

$$\widehat{\mathbf{W}}_{ML} = \underset{\mathbf{W}}{\arg\min} \sum_{j=1}^{m} \frac{1}{n^{(j)}} \sum_{i=1}^{n^{(j)}} \left( y_i^{(j)} - \sum_{a=1}^{d} x_{ia}^{(j)} w_a^{(j)} \right)^2 + \lambda \sum_{a=1}^{d} \|\mathbf{w}_a\|_2 \ , \qquad (1.14)$$

where $\mathbf{W}$ is the $d \times m$ matrix constructed by concatenating the parameters of all the $m$ models with the elements $W_{aj} = w_a^{(j)}$. The first term in the above is the empirical squared error loss of the $m$ tasks evaluated over their respective training samples. The second term is the multi-task lasso penalty which is reminiscent of the group lasso above. The parameter groupings are now defined for each input dimension $a \in \mathbb{N}_d$ across all the $m$ tasks $\mathbf{w}_a = (w_a^{(1)}, \ldots, w_a^{(m)})^T$, a row in matrix $\mathbf{W}$, so that the respective parameters are either all zero or nonzero and the corresponding input dimension is either discarded or selected for all the learning tasks.

## 1.3 Learning in reproducing kernel Hilbert spaces

We now return to the discussion of regularised risk minimisation (RRM) started in section 1.1.1. In the previous section we focused on the rather narrow hy-pothesis class $\mathcal{F}$ of linear functions $\mathbf{f}(\mathbf{x}) = \langle \mathbf{x}, \mathbf{w} \rangle$. Clearly, such a simple linear assumption may be too restrictive for some problems and we may want to drop it and move to using nonlinear functions instead. To achieve this, we extend the hypotheses class $\mathcal{F}$ in the RRM to functions belonging to reproducing kernel Hilbert spaces (RKHSs) of possibly nonlinear functions.

Without an ambition to provide an in-depth recount of the RKHS theory (available for example in Schölkopf et al. (2001), Berlinet and Thomas-Agnan (2004), and Steinwart and Christmann (2008)) we briefly review the main properties of the RKHS that make them particularly suitable for the function learning within the RRM framework.

**Definition 1** (RKHS). *Let $\mathcal{H}_K$ be a Hilbert space of functions $\mathbf{f} : \mathcal{X} \to \mathbb{R}$ over $\mathcal{X} \neq \emptyset$ with the inner product between two functions $\mathbf{f}, \mathbf{g} \in \mathcal{H}_K$ indicated as $\langle \mathbf{f}, \mathbf{g} \rangle_{\mathcal{H}_K}$ and the induced norm $\|\mathbf{f}\|_{\mathcal{H}_K}$. $\mathcal{H}_K$ is a reproducing kernel Hilbert space if for all $\mathbf{x} \in \mathcal{X}$ the Dirac evaluation functional $\delta_{\mathbf{x}} : \mathcal{H}_K \to \mathbb{R}$ defined as $\delta_{\mathbf{x}}(\mathbf{f}) := \mathbf{f}(\mathbf{x})$, $\forall \mathbf{f} \in \mathcal{H}_K$ is continuous (bounded).*

First note that the evaluation for every $\mathbf{x} \in \mathcal{X}$ is a linear operator with the norm

$$\|\delta_{\mathbf{x}}\| = \sup_{\mathbf{f} \in \mathcal{H}_K} \frac{\|\delta_{\mathbf{x}}(\mathbf{f})\|}{\|\mathbf{f}\|_{\mathcal{H}_K}} . \tag{1.15}$$

which after rewriting gives the additional necessary and sufficient property for all functions $\mathbf{f} \in \mathcal{H}_K$ and $\mathbf{x} \in \mathcal{X}$

$$|\delta_{\mathbf{x}}(\mathbf{f})| = |\mathbf{f}(\mathbf{x})| \leq \|\delta_{\mathbf{x}}\| \, \|\mathbf{f}\|_{\mathcal{H}_K} \leq C_{\mathbf{x}} \, \|\mathbf{f}\|_{\mathcal{H}_K} \text{ for some constatnt } C_{\mathbf{x}} < \infty . \tag{1.16}$$

The RKHS is thus a particularly well behaved function space. In the RKHS if two functions are near (in the norm) their evaluations at every point $\mathbf{x} \in \mathcal{X}$ are close as well

$$|\mathbf{f}(\mathbf{x}) - \mathbf{g}(\mathbf{x})| = |\delta_x(\mathbf{f}) - \delta_x(\mathbf{f})| \leq \|\delta_{\mathbf{x}}\| \, \|\mathbf{f} - \mathbf{g}\|_{\mathcal{H}_K} . \tag{1.17}$$

In consequence, norm convergence implies point-wise convergence, that is if $\lim_{n \to \infty} \|\mathbf{f}_n - \mathbf{f}\|_{\mathcal{H}_K} = 0$ then $\lim_{n \to \infty} |\mathbf{f}_n(\mathbf{x}) - \mathbf{f}(\mathbf{x})| = 0$ for all $\mathbf{x} \in \mathcal{X}$.

In the above definition we hint to an existence of some *reproducing kernel*. Let us define it next.

**Definition 2** (Kernel). *A function $k : \mathcal{X} \times \mathcal{X} \to \mathbb{R}$, $k : (\mathbf{x}, \mathbf{x}') \to k(\mathbf{x}, \mathbf{x}')$ is a kernel if there exists a Hilbert space $\mathcal{H}$ (feature space) and a feature map $\phi : \mathcal{X} \to \mathcal{H}$ such that for all $\mathbf{x}, \mathbf{x}' \in \mathcal{X}$ we have $k(\mathbf{x}, \mathbf{x}') = \langle \phi(\mathbf{x}), \phi(\mathbf{x}') \rangle_{\mathcal{H}}$.*

For a given kernel, neither the feature map $\phi$ nor the feature space $\mathcal{H}$ are unique. The kernel is clearly symmetric (by the symmetry of inner product) and it is also positive semidefinite since we have

$$\sum_{i=1}^{n} \sum_{j=1}^{n} a_i a_j k(\mathbf{x}_i, \mathbf{x}_j) = \left\langle \sum_{i=1}^{n} a_i \phi(\mathbf{x}_i), \sum_{j=1}^{n} a_j \phi(\mathbf{x}_j) \right\rangle_{\mathcal{H}} = \left\| \sum_{i=1}^{n} a_i \phi(\mathbf{x}_i) \right\|_{\mathcal{H}}^2 \geq 0 \quad (1.18)$$

Conversely, every symmetric positive definite function is a kernel (see e.g. Steinwart and Christmann (2008)).

**Definition 3** (Reproducing kernel). *A function $k : \mathcal{X} \times \mathcal{X} \to \mathbb{R}$, $k : (\mathbf{x}, \mathbf{x}') \to k(\mathbf{x}, \mathbf{x}')$ is a reproducing kernel of the Hilbert space $\mathcal{H}_K$ if and only if*

a) *the kernel sections $k_{\mathbf{x}} : \mathcal{X} \to \mathbb{R}$ centred at $\mathbf{x}$ and defined as $k_{\mathbf{x}}(\mathbf{x}') := k(\mathbf{x}, \mathbf{x}')$ belong to the $\mathcal{H}_K$ for all $\mathbf{x} \in \mathcal{X}$, and*

b) *the kernel $k$ has the reproducing property $\langle \mathbf{f}, k_{\mathbf{x}} \rangle_{\mathcal{H}_K} = \mathbf{f}(\mathbf{x})$ for all $\mathbf{x} \in \mathcal{X}$ and $\mathbf{f} \in \mathcal{H}_K$.*

c) *In result of a) and b) we have $\langle k_{\mathbf{x}}, k_{\mathbf{x}'} \rangle_{\mathcal{H}_K} = k(\mathbf{x}, \mathbf{x}')$ for all $\mathbf{x}, \mathbf{x}' \in \mathcal{X}$.*

From c) above we see that every reproducing kernel is a kernel as per definition 2 simply by choosing $\mathcal{H}_K$ as the feature space with the canonical feature map $\phi(\mathbf{x}) = k_{\mathbf{x}}$ (hence, it is also symmetric positive definite). Importantly, every Hilbert space $\mathcal{H}$ with a reproducing kernel $k$ is a RKHS (proof in appendix), a fact that is sometimes used as an alternative definition of the space. Finally, every RKHS has a unique reproducing kernel (from the Riesz representation theorem, see appendix) and every reproducing kernel has a unique RKHS (see e.g. Berlinet and Thomas-Agnan (2004) that for every reproducing kernel we can construct a unique RKHS).

### 1.3.1 Learning with kernels

After setting up the theoretical basis from the functional analysis perspective we now move onto more practical questions of learning with kernels. Kernel methods can be seen as methods based on comparing pairs of input instances $\mathbf{x}, \mathbf{x}' \in \mathcal{X}$ via the kernel function $k(\mathbf{x}, \mathbf{x}')$. The comparisons are, however, not performed in the original input space $\mathcal{X}$. Instead, the instances are first mapped from the original input space $\mathcal{X}$ into some feature space $\mathcal{H}$ deemed more appropriate for the given problem. The kernel is then defined as an inner product $k(\mathbf{x}, \mathbf{x}') = \langle \phi(\mathbf{x}), \phi(\mathbf{x}') \rangle_{\mathcal{H}}$ over the feature mappings $\phi : \mathcal{X} \to \mathcal{H}$.

The biggest advantage of kernel methods stems from results known as *representer theorems*. These show that the solutions of many RRM problems depend on the input points only through their inner products. As a result, the feature maps $\phi$ can be arbitrarily complex or even infinite dimensional (as long as they map to an inner product space) for they actually never need to be evaluated. Instead, their inner products can be directly replaced by the kernel function which is typically much easier to compute (this replacement is often referred to as the *kernel trick*).

Before providing more general results we illustrate the ideas on the example of the $\ell_2$ regularised least squares problem defined in equation (1.10) under the alternative name ridge regression.

**Regularised least squares**

Problem (1.10) is initially formulated to learn a linear function $\mathbf{f}(\mathbf{x}) = \mathbf{w}^T \mathbf{x}$. We can introduce non-linearity into the learning through some map $\phi : \mathcal{X} \to \mathcal{H}$ of the input instances $\mathbf{x} \in \mathcal{X}$ into an inner-product feature space $\mathcal{H} \subseteq \mathbb{R}^q$. For example, the map $\phi$ can build powers, interaction terms or other non-linear transformations of the input variables (dimensions of the input vectors $\mathbf{x}$). The function $\mathbf{f}(\mathbf{x}) = \mathbf{w}^T \phi(\mathbf{x})$ now operates over the non-linear features $\phi(\mathbf{x})$ and therefore $\mathbf{f}(\mathbf{x})$ is a non-linear function of the original inputs $\mathbf{x}$.

We indicate by $\Phi$ the $n \times q$ matrix of input mappings into the $q$-dimensional space $\mathcal{H}$ ($q < \infty$), by $\mathbf{y}$ the $n$-long vector of outputs (we consider here the single output case with $y \in \mathcal{Y} \subseteq \mathbb{R}$) and restate for reference the $\ell_2$ regularised least squares problem in a matrix form

$$\min_{\mathbf{w}} \frac{1}{n} \|\mathbf{y} - \Phi \mathbf{w}\|_2^2 + \lambda \|\mathbf{w}\|_2^2 \quad . \tag{1.19}$$

The problem is strictly convex and from the necessary and sufficient optimality condition we get the minimising solution

$$\widehat{\mathbf{w}} = (\Phi^T \Phi + \lambda n \mathbf{I}_q)^{-1} \Phi^T \mathbf{y} \quad . \tag{1.20}$$

By the matrix inversion lemma this is equivalent to

$$\widehat{\mathbf{w}} = \Phi^T (\Phi \Phi^T + \lambda n \mathbf{I}_n)^{-1} \mathbf{y} \tag{1.21}$$

and we can pick either of the equations (1.20) or (1.21) to obtain the function parameters $\mathbf{w}$ depending on the relative size of $n$ vs $q$ and hence, the respective computational complexity.

**Kernel regularised least squares**

To get to a kernel-based solution of the regularised least squares we observe from equation (1.21) that

1. we can represent the function parameters as

$$\widehat{\mathbf{w}} = \Phi^T \mathbf{c} = \sum_i^n \phi(\mathbf{x}_i) c_i \quad , \tag{1.22}$$

   where $\mathbf{c} = (\Phi \Phi^T + \lambda n \mathbf{I}_n)^{-1} \mathbf{y}$,

2. we can represent the function $\mathbf{f}(\mathbf{x}') = \mathbf{w}^T \phi(\mathbf{x}')$ in terms of $\mathbf{c}$ as

$$\mathbf{f}(\mathbf{x}') = \mathbf{w}^T \phi(\mathbf{x}') = \sum_i^n \phi(\mathbf{x}_i)^T \phi(\mathbf{x}') c_i = \sum_i^n c_i k(\mathbf{x}_i, \mathbf{x}') = \sum_i^n c_i k_{\mathbf{x}_i}(\mathbf{x}') \quad , \tag{1.23}$$

   where $k(\mathbf{x}, \mathbf{x}') = \langle \phi(\mathbf{x}), \phi(\mathbf{x}') \rangle$ is the kernel, and

3. the parameters $\mathbf{c}$ can be obtained as

$$\mathbf{c} = (\mathbf{K} + \lambda n \mathbf{I}_n)^{-1} \mathbf{y} \ , \qquad (1.24)$$

where $\mathbf{K} = \Phi\Phi^T$ is the kernel matrix with the elements $K_{ij} = k(\mathbf{x}_i, \mathbf{x}_j) = \langle \phi(\mathbf{x}_i), \phi(\mathbf{x}_j) \rangle$.

To summarise, instead of representing the function as $\mathbf{f}(\mathbf{x}) = \mathbf{w}^T \phi(\mathbf{x})$ and obtaining the parameters $\mathbf{w}$ from equations 1.20 or 1.21, we can represent the function as $\mathbf{f}(\mathbf{x}) = \sum_i^n c_i k_{\mathbf{x}_i}(\mathbf{x})$ and obtain the parameters $\mathbf{c}$ from equation (1.24). Whichever we decide to do, the values of the learned function $\mathbf{f}(\mathbf{x})$ will be the same for all points $\mathbf{x} \in \mathcal{X}$ (up to computational precision) because the two problems are exactly equivalent.

Nevertheless, there are practical reasons to use one or the other. The classical approach may seem more intuitive, may be less expensive than the kernel approach in the cases where the feature map $\phi$ is simple to evaluate and will involve fewer parameters to learn in the cases where $q < n$. The kernel approach is more general, applicable even in cases where the feature space is infinite dimensional (so clearly $q \gg n$) and can be cheaper if the kernel function is simpler to compute than the inner product over the features.

This last point means that rather than deciding on a suitable feature map and then deriving the corresponding kernel, it is habitual in kernel methods to start by choosing a suitable existing kernel (such as polynomial, Gaussian or Laplace, see e.g. Souza (2010) or Gärtner (2003) for more complete lists of kernel functions) or by constructing one with the desired properties. Recall from section 1.3 that for a given kernel neither the feature map $\phi$ nor the feature space $\mathcal{H}$ are unique and that one of the possible feature maps is the kernel section $\phi(\mathbf{x}) = k_{\mathbf{x}}$ living in the unique RKHS associated with the kernel.

To complete our kernel regularised least squares example we reformulate the optimisation problem (1.19) in terms of the kernels and the parameters $\mathbf{c}$. Using the representations (1.22) and (1.23) we have

$$\|\mathbf{y} - \Phi\mathbf{w}\|_2^2 = \sum_i^n (y_i - \mathbf{w}^T \phi(\mathbf{x}_i))^2 = \sum_i^n \left(y_i - \sum_j^n k(\mathbf{x}_i, \mathbf{x}_j')c_j\right)^2 = \|\mathbf{y} - \mathbf{K}\mathbf{c}\|_2^2 \quad (1.25)$$

and

$$\|\mathbf{w}\|_2^2 = \left\| \sum_i^n \phi(\mathbf{x}_i) c_i \right\|_2^2 = \left\langle \sum_i^n \phi(\mathbf{x}_i) c_i, \sum_j^n \phi(\mathbf{x}_j) c_j \right\rangle = \sum_i^n \sum_j^n c_i c_j k(\mathbf{x}_i, \mathbf{x}_j) = \mathbf{c}^T \mathbf{K} \mathbf{c} \tag{1.26}$$

The equivalent kernel formulation of the optimisation problem (1.19) is thus

$$\min_{\mathbf{c}} \frac{1}{n} \|\mathbf{y} - \mathbf{K}\mathbf{c}\|_2^2 + \lambda \mathbf{c}^T \mathbf{K} \mathbf{c} \quad, \tag{1.27}$$

with the elements of the kernel matrix $K_{ij} = k(\mathbf{x}_i, \mathbf{x}_j) = \langle \phi(\mathbf{x}_i), \phi(\mathbf{x}_j) \rangle$. Note that if the kernel functions $k(\mathbf{x}_i, \mathbf{x}_j)$ can be computed directly (without first obtaining the feature mappings and evaluating the inner products) (1.27) is a valid optimisation problem even for infinite dimensional feature spaces $\mathcal{H}^2$.

Finally, we see from (1.26), (1.23) and point 3 in definition 3 that

$$\mathbf{c}^T \mathbf{K} \mathbf{c} = \sum_i^n \sum_j^n c_i c_j k(\mathbf{x}_i, \mathbf{x}_j) = \sum_i^n \sum_j^n c_i c_j \left\langle k_{\mathbf{x}_i}, k_{\mathbf{x}_j} \right\rangle_{\mathcal{H}_K} = \left\| \sum_i^n k_{\mathbf{x}_i} c_i \right\|_{\mathcal{H}_K}^2 = \|\mathbf{f}\|_{\mathcal{H}_K}^2 \quad. \tag{1.28}$$

This brings us to yet another equivalent formulation of the regularised least squares problem in the form of the RRM functional introduced in equation (1.6)

$$\min_{\mathbf{f} \in \mathcal{H}_K} \frac{1}{n} \sum_i^n (y_i - \mathbf{f}(\mathbf{x}_i))^2 + \lambda \|\mathbf{f}\|_{\mathcal{H}_K}^2 \quad. \tag{1.29}$$

Here the hypotheses space $\mathcal{H}_K$ is the reproducing kernel Hilbert space uniquely associated with the reproducing kernel $k$ used for constructing the kernel matrices $\mathbf{K}$ in (1.27), the loss function L is the squared error loss, and the regulariser R is the square of the function norm in $\mathcal{H}_K$.

### Beyond regularised least squares

In the above we have shown the equivalence of the variational problem (1.29) learning a predictive function $\mathbf{f} \in \mathcal{H}_K$ with the finite dimensional ($n < \infty$) problem (1.27) learning the parameters $\mathbf{c}$ of the function representations $\mathbf{f} =$ $\sum_i^n c_i k_{\mathbf{x}_i}$ and, assuming there exists a finite dimensional feature map $\phi$ such that $k(\mathbf{x}_i, \mathbf{x}_j) = \langle \phi(\mathbf{x}_i), \phi(\mathbf{x}_j) \rangle$, with the finite dimensional problem (1.19).

---

[2]It does not rely on constructing the infinite dimensional $\Phi$ matrix while (1.19) does.

Important building stones for showing the equivalence were the properties of the reproducing kernel from definition 3 and the function representation (1.23) as a linear combination of the kernel sections centered at the training input instances $\mathbf{x}_i \in \mathcal{S}_n$.

In fact, it can be shown (see the appendix) that for all RRM functionals

$$\min_{\mathbf{f} \in \mathcal{H}_K} \frac{1}{n} \sum_{i=1}^{n} L(y_i, \mathbf{f}(\mathbf{x}_i)) + \lambda Q(\|\mathbf{f}\|_{\mathcal{H}_K})$$

with RKHS hypotheses spaces $\mathcal{H}_K$, arbitrary convex losses L, and regularisers $R(\mathbf{f}) = Q(\|\mathbf{f}\|_{\mathcal{H}_K})$, where Q are strictly increasing functions, the minimisers can be always represented as $\mathbf{f}(\mathbf{x}) = \sum_i^n c_i k_{\mathbf{x}_i}(\mathbf{x})$ (hence, the input instances enter the solution only through some inner products). This crucial result means that such a function learning problem can always[3] be reduced to solving a finite dimensional problem with respect to the $n$-long parameter vector $\mathbf{c}$ of some functional $\Omega(\mathbf{Kc}, \sqrt{\mathbf{c}^T \mathbf{Kc}})$ of which the specific form depends on the particular choices of L and Q[4]. On the other hand, being able to represent the function as a finite linear combination of the basis $k_{\mathbf{x}_i}$ also shows that we cannot learn an arbitrarily complex function from a limited set of input examples $\mathbf{x}_i \in \mathcal{S}_n$.

### 1.3.2 Extension to multivariate outputs

So far we considered scalar output values $y \in \mathcal{Y} \subseteq \mathbb{R}$ and correspondingly the RKHS of scalar-valued functions $\mathbf{f}$. We now extend the concepts to multivariate outputs $\mathbf{y}$ belonging to real Hilbert spaces $\mathcal{Y} \subseteq \mathbb{R}^m$ with well-defined inner products $\langle .,. \rangle_{\mathcal{Y}}$. Following the texts of Micchelli and Pontil (2005) and Carmeli et al. (2006) we define the RKHS of vector-valued functions in analogy with the scalar case.

---

[3]We focus here on the regularised least squares as the most relevant setting for our work. Examples of other well known problems covered by this general result comprise the support vector machines for classification and regression, kernel logistic regression, kernel robust regression, etc.

[4]Though it may not be possible to express the minimising $\mathbf{c}$ in a closed form similar to (1.24) it can usually be found by some iterative procedure such as gradient descent.

**Definition 4** (RKHS of vector-valued functions)**.** *Let $\mathcal{H}_K$ be a Hilbert space of functions* $\mathbf{f} : \mathcal{X} \rightarrow \mathcal{Y}$ *over* $\mathcal{X} \neq \emptyset$ *with the inner product between two functions* $\mathbf{f}, \mathbf{g} \in \mathcal{H}_K$ *indicated as* $\langle \mathbf{f}, \mathbf{g} \rangle_{\mathcal{H}_K}$ *and the induced norm* $\|\mathbf{f}\|_{\mathcal{H}_K}$. $\mathcal{H}_K$ *is a reproducing kernel Hilbert space if for all* $\mathbf{x} \in \mathcal{X}$ *there exists a positive constant* $C_{\mathbf{x}} < \infty$ *such that*

$$\|\mathbf{f}(\mathbf{x})\|_{\mathcal{Y}} \leq C_{\mathbf{x}} \|\mathbf{f}\|_{\mathcal{H}_K} \quad . \tag{1.30}$$

*Equivalenty, for all* $\mathbf{x} \in \mathcal{X}$ *and* $\mathbf{y} \in \mathcal{Y}$ *the linear functional* $T_{xy}$ *that maps* $\mathbf{f}$ *to* $T_{xy}(\mathbf{f}) = \langle \mathbf{y}, \mathbf{f}(\mathbf{x}) \rangle_{\mathcal{Y}}$ *is continuous (bounded).*

As in the scalar case, each RKHS is uniquely associated with a reproducing kernel $\mathbf{H}$.

**Definition 5** (Operator-valued reproducing kernel)**.** *Let* $\mathcal{L}_{\mathcal{Y}}$ *be the set of all bounded linear operators from the Hilbert space* $\mathcal{Y}$ *into itself. A function* $\mathbf{H} : \mathcal{X} \times \mathcal{X} \rightarrow \mathcal{L}_{\mathcal{Y}}$, $\mathbf{H} : (\mathbf{x}, \mathbf{x}') \rightarrow \mathbf{H}(\mathbf{x}, \mathbf{x}')$, *is the unique reproducing kernel of the RKHS* $\mathcal{H}_K$ *of functions* $\mathbf{f} : \mathcal{X} \rightarrow \mathcal{Y}$ *if and only if*

a) $\mathbf{H}(\mathbf{x}, \mathbf{x}') = \mathbf{H}(\mathbf{x}', \mathbf{x})$, $\forall \mathbf{x}, \mathbf{x}' \in \mathcal{X}$ *(symmetry),*

b) $\sum_i^n \sum_j^n \langle \mathbf{y}_i, \mathbf{H}(\mathbf{x}_i, \mathbf{x}_j) \mathbf{y}_j \rangle_{\mathcal{Y}} \geq 0$, $\forall (\mathbf{x}, \mathbf{y}) \in (\mathcal{X} \times \mathcal{Y})$, $n \in \mathbb{N}$ *(positive semidefinitness),*

c) *the functions* $\mathbf{H}_{\mathbf{x}} \mathbf{y} : \mathcal{X} \rightarrow \mathcal{Y}$ *centred at* $\mathbf{x}, \mathbf{y}$ *and defined as* $\mathbf{H}_{\mathbf{x}} \mathbf{y}(\mathbf{x}') := \mathbf{H}(\mathbf{x}', \mathbf{x}) \mathbf{y}$ *belong to the* $\mathcal{H}_K$ *for all* $(\mathbf{x}, \mathbf{y}) \in \mathcal{X} \times \mathcal{Y}$,

d) *the kernel has the reproducing property* $\langle \mathbf{f}, \mathbf{H}_{\mathbf{x}} \mathbf{y} \rangle_{\mathcal{H}_K} = \langle \mathbf{y}, \mathbf{f}(\mathbf{x}) \rangle_{\mathcal{Y}}$ *for all* $\mathbf{x} \in \mathcal{X}$ *and* $\mathbf{f} \in \mathcal{H}_K$.

e) *As a result of c) and d) we have* $\langle \mathbf{H}_{\mathbf{x}} \mathbf{y}, \mathbf{H}_{\mathbf{x}'} \mathbf{y}' \rangle_{\mathcal{H}_K} = \langle \mathbf{y}', \mathbf{H}(\mathbf{x}', \mathbf{x}) \mathbf{y} \rangle_{\mathcal{Y}}$ *for all* $(\mathbf{x}, \mathbf{y}), (\mathbf{x}', \mathbf{y}') \in \mathcal{X} \times \mathcal{Y}$.

From the above definitions we see that for the output space $\mathcal{Y} = \mathbb{R}^m$ the kernel $\mathbf{H}$ is a $m \times m$ matrix of scalar-valued functions $\mathbf{H}_{ij} : \mathcal{X} \times \mathcal{X} \rightarrow \mathbb{R}$ for all $i, j \in \mathbb{N}_m$. To better understand what these are let us first indicate by $\mathbf{e}_i \in \mathcal{Y} = \mathbb{R}^m$ the vectors of the standard coordinate basis of $\mathbb{R}^m$. From e) in definition 5 we then get for the $(ij)$ element of the kernel matrix

$$\mathbf{H}(\mathbf{x}, \mathbf{x}')_{ij} = \left\langle \mathbf{e}_i, \mathbf{H}(\mathbf{x}, \mathbf{x}') \mathbf{e}_j \right\rangle_{\mathcal{Y}} = \left\langle \mathbf{H}_{\mathbf{x}} \mathbf{e}_i, \mathbf{H}_{\mathbf{x}'} \mathbf{e}_j \right\rangle_{\mathcal{H}_K} \quad . \tag{1.31}$$

Some further properties of the operator-valued kernels with their proofs are provided in section 1.A.1 of the appendix. It is also important to understand that the definitions 4 and 5 are generalizations that reduce to the scalar reproducing kernel $k$ and the corresponding RKHS defined in section 1.3 with $\mathcal{Y} = \mathbb{R}^1$ and $\mathcal{L}_{\mathcal{Y}}$ the set of scalars.

The representer theorem introduced in section 1.3.1 for the scalar case can also be extended to the vector-valued functions $\mathbf{f} : \mathcal{X} \to \mathcal{Y} \subseteq \mathbb{R}^m$. The minimisers of the RRM problem

$$\min_{\mathbf{f} \in \mathcal{H}_K} \frac{1}{n} \sum_{i=1}^{n} \mathrm{L}(\mathbf{y}, \mathbf{f}(\mathbf{x}_i)) + \lambda \mathrm{Q}(\|\mathbf{f}\|_{\mathcal{H}_K}) \ ,$$

with RKHS hypotheses spaces $\mathcal{H}_K$, arbitrary convex losses L, and regularisers $R(\mathbf{f}) = Q(\|\mathbf{f}\|_{\mathcal{H}_K})$, where Q are strictly increasing functions, can always be represented as $\mathbf{f}(\mathbf{x}) = \sum_{i}^{n} H(\mathbf{x}_i, \mathbf{x}) \mathbf{c}_i$, where $\mathbf{c}_i \in \mathcal{Y}$, $\forall i \in \mathbb{N}_n$. Combining this result with the reproducing property means that the original infinite dimensional problem can be reduced to solving a finite dimensional problem with respect to the $n$ parameter vectors $\mathbf{c}_i$.

## 1.4 Multivariate time series forecasting

Time series forecasting is a particular supervised learning problem concerned with predicting the future values of some sequentially observed data. The problem itself is certainly not new. The need for such predictions has since long been recognised by planners and managers as invaluable support for their decisions. What has changed over the past several years is the quantity of indicators that are now routinely monitored and for which the forecasts are required, be it for various technological, physical, financial, socio-economical, bio-chemical, or other processes.

There exists a plethora of more-or-less complex methods for forecasting individual time series, ranging from simple linear models, neural networks, similarity and distance based approaches, and more. However, when it comes to forecasting multivariate time-series systems the availability of customised methods and efficient tools is a lot less abundant and in many aspects not satisfactory. Forecasting in large systems of time series brings along several complications but also some opportunities that had not existed or had not been thoroughly considered when tools for predicting smaller sets of series had been developed. Multivariate models need to capture much more complex relationships, yet due to the quickly increasing dimensionality they are often difficult to estimate. Many forecasting methods do not scale favourably with the number of indicators in the system and they quickly suffer from over-parametrization as more and more series are added to the modelled system. On the other hand, multiple indicators developing in parallel are likely to follow some shared patterns and form structures which can be exploited in developing the models.

As De Gooijer and Hyndman (2006) note in their review of 25 years of research into time series forecasting, the multivariate forecasting methods haven't been widely adopted in practice, largely due to "...  lack of empirical research on robust forecasting algorithms for multivariate models ...  ". Similar messages surface from the specialised time-series workshops organised alongside major machine learning conferences such as NIPS and ICML where the problem of forecasting high-dimensional time series has been on multiple occasions flagged as one of the research priorities for the community.

### 1.4.1   Time series as a stochastic process

We first develop some key concepts that will allow us to formulate mathematical (statistical) models necessary for the analysis of the multivariate time series dynamics.

Let $(\Omega, \mathcal{A}, \mathrm{P})$ be a probability space, $\mathbb{T}$ a countable index set (e.g. the set of all positive integers $\mathbb{N}$) and $\mathbf{y} : \mathbb{T} \times \Omega \to \mathcal{Y} \subseteq \mathbb{R}^m$ functions such that for each $t \in \mathbb{T}$, $\mathbf{y}(t,.) : \Omega \to \mathcal{Y} \subseteq \mathbb{R}^m$ are the multivariate[5] random variables over the same probability space $(\Omega, \mathcal{A}, \mathrm{P})$. We indicate the random variable $\mathbf{y}(t,.)$ for a fixed $t$ by $\mathbf{y}_t$.

A *stochastic process* is the collection of the random variables $\{\mathbf{y}_t : t \in \mathbb{T}\}$. Note that for each fixed $t \in \mathbb{T}$, $\mathbf{y}_t = \mathbf{y}(t,.)$ is a random variable (function) defined on $\Omega$. On the other hand, for each $\omega \in \Omega$, $\mathbf{y}(.,\omega)$ is a function of $\mathbb{T}$. A *realization of a stochastic process* (sample path) is the collection of functions $\mathbf{y}(.,\omega) : \mathbb{T} \rightarrow \mathcal{Y}$ for a fixed $\omega \in \Omega$, or equivalently the collection of vectors $\{\mathbf{y}_t(\omega) : t \in \mathbb{T}\}$. The underlying stochastic process is usually referred to as the data generating process (DGP).

---

[5]As explained in section 1.1, the univariate case is subsumed in our discussion by putting $m = 1$ and therefore all concepts described in this section are valid also for univariate time series.

The term *time series* is used for both the stochastic process and its realization when the ordered index set $\mathbb{T}$ represents time. The time series is usually denoted simply by $\{\mathbf{y}_t\}$ and its elements at time points $t$ by $\mathbf{y}_t$ when it is clear from the context whether this refers to the stochastic process or its realization. We consider here discrete time series with equidistant time points $t = 1, 2, 3 \ldots$.

**Stationary time series**

An important concept in time series theory is related to the regularity of the series in time (invariance to time shift). A time series $\{\mathbf{y}_t\}$ is *strictly stationary* if the joint probability distribution of every sub-collection $\{\mathbf{y}_t, \ldots \mathbf{y}_{t+k}\}$ is the same as of the time-shifted sub-collection $\{\mathbf{y}_{t+h}, \ldots \mathbf{y}_{t+k+h}\}$ for all time points $t$ and time shifts $k, h \in \mathbb{N}$.

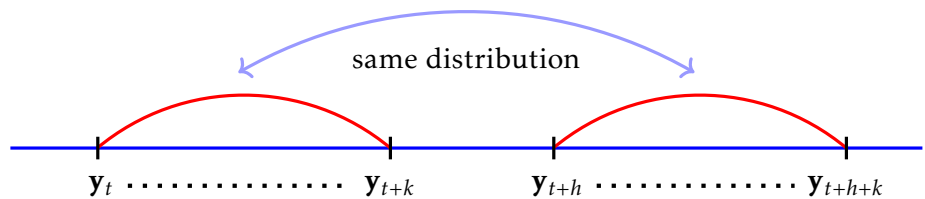

Figure 1.1: In a stationary time series, every subcollection of a fixed number of points has the same probability distribution irrespective of shifts in time.

Often in theory and practice the strict stationarity concept is considered unnecessarily restrictive and too complicated to work with. Instead of examining the full probability distributions, a milder concept based on the time-invariance of the moments of the series is often used. A time series $\{\mathbf{y}_t\}$ is *covariance stationary* (weakly stationary, second-order stationary) if its mean and autocovariance functions do not depend on the specific time point $t$; that is for all $t \in \mathbb{T}$ we have

$$\text{mean:} \qquad \mathrm{E}[\mathbf{y}_t] = \int \mathbf{y}_t \, \mathrm{d}P = \boldsymbol{\mu} \qquad\qquad (1.32)$$

$$\text{auto-covariance:} \qquad \mathrm{Cov}[\mathbf{y}_t, \mathbf{y}_{t+h}] = \mathrm{E}[(\mathbf{y}_t - \boldsymbol{\mu})(\mathbf{y}_{t+h} - \boldsymbol{\mu})] = \boldsymbol{\Gamma}^{(h)} \ . \qquad (1.33)$$

A strictly stationary process with finite first and second moments is covariance stationary. However, the converse is generally not true. An important example of a process for which the covariance stationarity implies strict stationarity is the Gaussian time series, that is a series with the distributions of every collection $\{\mathbf{y}_t, \dots, \mathbf{y}_{t+h} : t \in \mathbb{T},\ h \in \mathbb{N}\}$ being multivariate normal. Since the Gaussian distribution is completely described by its mean and covariance, a weakly stationary Gaussian time series is also strictly stationary.

Stationarity is a critical property in time series analysis and modelling. It replaces the IID assumption on data samples customarily used in supervised ML (section 1.1). The random variables $\mathbf{y}_t$ in a stationary time series are still identically distributed. They are, however, not necessarily independent. In fact, it is the dependence between the variables $\mathbf{y}_t$ across different time points $t$ that is exploited when constructing models for time series forecasting.

In our work we follow the stationarity assumption for all the time series we explore. Many series observed in practice may exhibit multiple signs of nonstationarity. These may be more or less difficult to identify and understand as they take different forms such as gradual, periodic or sudden changes in levels or variability. There exist multiple techniques specifically developed for dealing with various nonstationarities, sometimes making direct use of them for forecasting (such as trend and/or seasonality modelling), other times transforming the series into one that can be considered stationary (e.g. differencing, variance stabilization). We do not discuss these in any detail here (the interested reader may find more in the abundant literature on this topic, e.g. Brockwell and Davis (1991)), we simply assume that the necessary transformations have been applied and that the resulting series can be reasonably viewed and treated as stationary.

### 1.4.2 The forecasting problem

The goal of time series forecasting is to predict the future value $\mathbf{y}_{o+h}$ of some time series $\{\mathbf{y}_t\}$ observed in the past. The last observed time point $o \in \mathbb{T}$ is usually referred to as the *forecast origin* and the number of steps $h$ the forecast shall look ahead as the *forecast horizon*. To produce the forecast $\widehat{\mathbf{y}}_{o+h}$ we can, in principle, use any information $\mathrm{i} \in \mathcal{I}$ available to us at the moment of the forecasting exercise. Following the standard machine learning terminology, the forecasting goal is to learn the predictive function $\mathbf{f}_h : \mathcal{I} \to \mathcal{Y}$ that approximates well the unknown future value $\mathbf{y}_{o+h}$.

In this thesis we focus on the class of *vector autoregressive models (VARs)* which constitute one of the principal tools of *multivariate* time series analysis and forecasting, and in which the input $\mathcal{I}$ to the predictive function is the history of the time series itself.

Without the loss of generality we limit our discussion to *one-step-ahead forecasts* (for which the forecast horizon is fixed to $h = 1$) and for notational simplicity drop in the following text the subscript $h$ from the predictive function $\mathbf{f}$. For longer horizons $h > 1$, the predictions can be obtained recursively from $\mathbf{f}$ by replacing the unknown true data by their forecasts. Alternatively, a specific predictive function can be learned directly for the required horizon $h > 1$, keeping most of the discussion in here unchanged. The advantages and disadvantages of these two approaches are discussed for example in Taieb (2014).

Figure 1.2 illustrates the forecasting approach in a VAR model. The forecast $\widehat{\mathbf{y}}_t$ for any time point $t$ is a vector-valued function of the $p$ previous values of the time series $\{\mathbf{y}_{t-p}, \dots, \mathbf{y}_{t-1}\}$ concatenated into a single input vector $\mathbf{x}_t$

$$\widehat{\mathbf{y}}_t = \mathbf{f}(\mathbf{y}_{t-p}, \dots, \mathbf{y}_{t-1}) = \mathbf{f}(\mathbf{x}_t) \ . \tag{1.34}$$

Note the index alignment between the inputs $\mathbf{x}_t$ and outputs $\mathbf{y}_t$ even though these relate to shifted parts of the same series.

To learn the forecasting function $\mathbf{f}$, similarly to section 1.1.1 we first fix a measure of quality of the prediction, typically in time series forecasting to the usual squared error loss $\mathrm{L}(\mathbf{y}_t, \mathbf{f}(\mathbf{x}_t)) = \|\mathbf{y}_t - \mathbf{f}(\mathbf{x}_t)\|_2^2$ (see also equation (1.1)). As at the learning time the future value $\mathbf{y}_t$ is yet unknown, we can only aim at

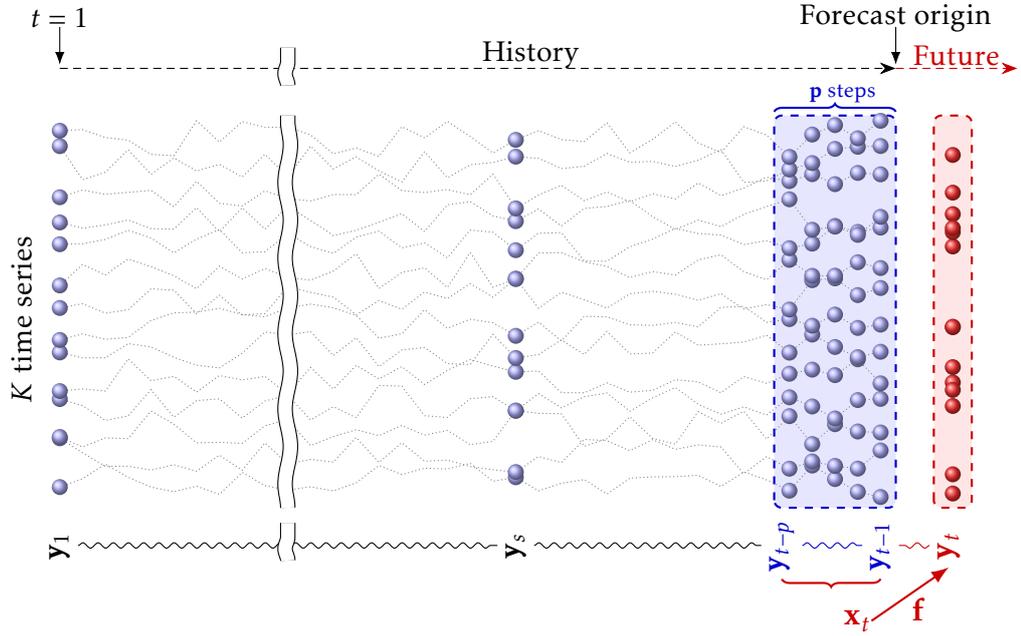

Figure 1.2: In VAR models the future of a time series is predicted using its own past. The point to predict is indicated by $\mathbf{y}_t$, the $p$ previous steps are indicated $\mathbf{x}_t$.

minimising the expected loss (the expected risk), equation (1.2). For we typically do not know the probability distribution generating the time series data, same as in equation (1.5) we replace the minimisation of the expectation by its sample estimate, the empirical risk

$$\widehat{\mathbf{f}} = \arg\min_{\mathbf{f}} \frac{1}{n} \sum_{t=1}^{n} \|\mathbf{y}_t - \mathbf{f}(\mathbf{x}_t)\|_2^2 \qquad (1.35)$$

over the available observations $\{\mathbf{y}_1, \ldots, \mathbf{y}_n\}$ of the time series and the corresponding past value vectors $\{\mathbf{x}_1, \ldots, \mathbf{x}_n\}$, the *training data*.

There are two caveats worth noticing in the above minimisation problem. First, the average in the empirical risk is a *time average* calculated over multiple time points of the single realization path of the time series instead of the *ensemble average*, the average calculated over multiple realizations of the DGP for a fixed time point (corresponding to the expectation in equation (1.32)). Second, as we are dealing with time series, the summands in equation (1.35) are not independent and therefore we cannot apply the standard law of large numbers to argue that the empirical risk is a consistent estimate of the expected risk.

Using the time average instead of the ensemble average is a must in most practical situations as we rarely observe multiple realizations of the same DGP. From the theoretical point of view, see e.g. Hamilton (1994), this replacement is admissible if the time series is *mean ergodic* that is if its time averages converge in probability to the DGP expectation, $\text{plim}_{n\to\infty} \frac{1}{n} \sum_{t=1}^{n} \mathbf{y}_t = \text{E}[\mathbf{y}_t] = \boldsymbol{\mu}$. A covariance stationary multivariate time series is mean ergodic if the autocovariance functions of all the components are absolutely summable, $\sum_{h=1}^{\infty} \boldsymbol{\Gamma}_{ii}^{(h)} < \infty$, and we will assume it throughout this thesis. In this case, using an alternative version of the law of large numbers for serially dependent data it can be shown that (1.35) converges to the expected risk and therefore as $n \to \infty$ it can give us a good idea of the quality of the forecasts.

A prominent place within the class of VAR models is occupied by the linear VARs, e.g. Lütkepohl (2005), in which the predictive function $\mathbf{f}$ is a linear function of the inputs $\mathbf{f}(\mathbf{x}_t) = \mathbf{W}^T \mathbf{x}_t$. Linear VARs have been at the centre of research in multivariate time series analysis and forecasting since the late 70's, when they were first investigated as a more flexible alternative to structural equation models in the works of Sargent and Sims (Qin 2011). For their relative simplicity and often good performance they are also the most common in forecasting practice. More recently, approaches for capturing possible nonlinearities in the time series dynamics have attracted more interest of the research community, e.g. Turkman et al. (2014).

A major problem in all VARs is the number of free parameters that increases rapidly with the number of modelled series (quadratically in the linear case). The traditional estimation methods developed for small VARs fail to recover well-performing models for larger-systems. For finite time series of length $n < \infty$, the predictive function $\widehat{\mathbf{f}}$ chosen by minimising the empirical risk (1.35) suffers from the same overfitting tendency as discussed in section 1.1.1. As in standard supervised learning, the generalization performance can be improved by controlling the hypothesis search space $\mathbf{f} \in \mathcal{F}$ or by applying the RRM strategy, equation (1.6) (with some caution due to the violation of the independence assumptions[6]).

---

[6]In the statistical learning theory, the risk associated with the hypothesis learned by different algorithms and methods is analysed by formulating generalization error bounds based on some measures

Early attempts to address the overfitting problem in linear VARs were based on Bayesian approaches (Doan et al. 1984). Although theoretically sound, these methods are still difficult to use in practice due to the non-trivial decisions related to choosing suitable priors, and due to the complexity of the resulting optimisation problems. Current research, e.g. Koop 2013, focuses on addressing these drawbacks of Bayesian VARs. An alternative approach, e.g. Lozano et al. 2009, derived from the regularization learning theory is typically coupled with discovering graphs of Granger causality (Granger 1969). This is where our research is situated as well and we present some of our proposals in this area in chapters 2 and 3.

### 1.4.3 Granger causality

To study the dynamical relationships in time series processes, Granger (1969) proposed a practical definition of *sequential causality* based on the accuracy of least-squares prediction functions. In brief, for two time series processes $\{\mathbf{y}_t\}$ and $\{\mathbf{z}_t\}$, $\{\mathbf{y}_t\}$ is said to Granger-cause $\{\mathbf{z}_t\}$ if given all other relevant information we can predict the future of $\{\mathbf{z}_t\}$ better (in the mean-square-error sense) using the history of $\{\mathbf{y}_t\}$ than without it. The above definition is based on two fundamental principals (Granger 1988): i) the cause occurs before the effect, ii) $\{\mathbf{y}_t\}$ contains some information about $\{\mathbf{z}_t\}$ that is not available elsewhere.

Though the Granger-causality concept seems rather straightforward, there are (at least) three points worth considering. First, while the definition above uses the term *all other relevant information* in the sense of all information in the universe, in practice the choice of information $i_t \in \mathcal{I}$ included in the analysis lies in the hands of the investigator. A different choice of the information set $j_t \in \mathcal{I}$ may lead to different conclusions about the causal effect of $\{\mathbf{y}_t\}$ on $\{\mathbf{z}_t\}$. In this respect the results of the analysis are an outcome of subjective prior decisions.

Second, the Granger causality definition is purely technical based on the pre-

---

of complexity of the hypothesis spaces. For time series, the classical independence assumption used in these bounds needs to be adapted for the serial dependence between observations by taking some relaxed assumptions (most often *mixing properties* related to diminishing dependence for longer lags). Though this is an interesting active area of research, e.g. McDonald (2012) or Kuznetsov and Mohri (2014), it remains out of the scope of our discussion here.

dictive accuracy of functions with differing input sets. It does not per se seek to understand the underlying forces driving the relationships and leaves the interpretation to the analyst.

Third, taking time series $\{\mathbf{y}_t\}$ into consideration means that the analyst has from the start some degree of belief about its causal effects on $\{\mathbf{z}_t\}$. The results of the Granger causality analysis are likely to change this belief (increase or decrease) but the extent will depend on the quality and quantity of data used to support the analysis, and the coverage and relevance of the other information $i_t$ considered.

These caveats together with more philosophical objections to the above concept even being called *causality*, e.g. Leamer (1985), have led to some criticism. For example, Zellner (1979) questioned its restriction to stochastic processes and its disconnect from economic laws, Holland (1986) strongly argued for the need of experimentation to discuss causality.

Despite such reservations, the notion of Granger causality has now been widely accepted as a useful operational tool for multivariate time series analysis, e.g. Hamilton (1994), Lütkepohl (2005). It has been generalized by Florens and Mouchart (1982) to cater for non-linear relationships by considering conditional independence as a sign of non-causality. Eichler (2012) further extended it to multivariate analysis through graphical models capturing the causal dependencies within the time series system. Lately, graphical Granger methods, e.g. Arnold et al. (2007), follow the principles of sparse learning, section 1.2, to discover the Granger-causal structure of the graphs from observed data. In this spirit, and acknowledging that the concept of Granger causality may not be quite coherent with other more modern formalisms of causality, e.g. Pearl (2009), we build upon it in this thesis.

# Appendix

## 1.A   Proofs

**The minimiser of problem** (1.35) **is the conditional expectation** $\mathbf{f}^*(\mathbf{x}) = \mathrm{E}[\mathbf{y}|\mathbf{x}]$.

*Proof.* Let $\mathbf{f}$ be an arbitrary measurable function over $\mathcal{X}$ that shall serve as a predictor for the unknown random variable $\mathbf{y}$. Its expected squared error loss is

$$\mathrm{E}\left[\|\mathbf{y} - \mathbf{f}(\mathbf{x})\|_2^2\right] = \mathrm{E}\left[\|\mathbf{y} - \mathrm{E}(\mathbf{y}|\mathbf{x}) + \mathrm{E}(\mathbf{y}|\mathbf{x}) - \mathbf{f}(\mathbf{x})\|_2^2\right]$$

$$= \underbrace{\mathrm{E}\left[\|\mathbf{y} - \mathrm{E}(\mathbf{y}|\mathbf{x})\|_2^2\right]}_{A} + \underbrace{\mathrm{E}\left[\|\mathrm{E}(\mathbf{y}|\mathbf{x}) - \mathbf{f}(\mathbf{x})\|_2^2\right]}_{B} - \underbrace{2\mathrm{E}\left[(\mathbf{y} - \mathrm{E}(\mathbf{y}|\mathbf{x}))^T(\mathrm{E}(\mathbf{y}|\mathbf{x}) - \mathbf{f}(\mathbf{x}))\right]}_{C}$$

In the above the term $C$ disappears because

$$\mathrm{E}\left[(\mathbf{y} - \mathrm{E}(\mathbf{y}|\mathbf{x}))^T(\mathrm{E}(\mathbf{y}|\mathbf{x}) - \mathbf{f}(\mathbf{x}))\right] = \mathrm{E}\left[(\mathbf{y} - \mathrm{E}(\mathbf{y}|\mathbf{x}))^T\mathbf{h}(\mathbf{x})\right] \qquad \mathbf{h}(\mathbf{x}) = \mathrm{E}(\mathbf{y}|\mathbf{x}) - \mathbf{f}(\mathbf{x})$$

$$= \mathrm{E}\left[\mathbf{y}^T\mathbf{h}(\mathbf{x})\right] - \mathrm{E}\left[\mathrm{E}(\mathbf{y}|\mathbf{x})^T\mathbf{h}(\mathbf{x})\right]$$

$$= \mathrm{E}\left[\mathbf{y}^T\mathbf{h}(\mathbf{x})\right] - \mathrm{E}\left[\mathrm{E}(\mathbf{y}^T\mathbf{h}(\mathbf{x})|\mathbf{x})\right]$$

$$= \mathrm{E}\left[\mathbf{y}^T\mathbf{h}(\mathbf{x})\right] - \mathrm{E}\left[\mathbf{y}^T\mathbf{h}(\mathbf{x})\right] \qquad \text{iterated expectation}$$

$$= 0 \; .$$

We have no control over the term $A$ (it is constant), and the term $B$ is minimised if $\mathbf{f}(\mathbf{x}) = \mathrm{E}(\mathbf{y}|\mathbf{x})$.  $\square$

**A Hilbert space $\mathcal{H}$ with a reproducing kernel $k$ is a RKHS**

*Proof.* Using the reproducing property of the kernel we have

$$|\delta_{\mathbf{x}}(\mathbf{f})| = |\mathbf{f}(\mathbf{x})| = |\langle \mathbf{f}, k_{\mathbf{x}} \rangle| \le \|k_{\mathbf{x}}\|_{\mathcal{H}} \|\mathbf{f}\|_{\mathcal{H}} \; , \tag{1.36}$$

which after comparison to the inequality (1.16) shows the continuity of the evaluation operators (assuming $\|k_{\mathbf{x}}\|_{\mathcal{H}} < \infty$).  $\square$

**Every RKHS has a unique reproducing kernel.**

*Proof.* From the Riesz representation theorem we have that the Dirac evaluation functional $\delta_{\mathbf{x}}$ for every $\mathbf{x} \in \mathcal{X}$ can be uniquely represented by an inner product $\delta_{\mathbf{x}}(\mathbf{f}) = \langle \mathbf{f}, k_{\mathbf{x}} \rangle_{\mathcal{H}}$. The unique function $k_{\mathbf{x}}$ is the kernel section with the reproducing property $\delta_{\mathbf{x}}(\mathbf{f}) = \mathbf{f}(\mathbf{x}) = \langle \mathbf{f}, k_{\mathbf{x}} \rangle_{\mathcal{H}}$. By applying the evaluation functional to the kernel section we get $\delta_{\mathbf{x}'}(k_{\mathbf{x}}) = k_{\mathbf{x}}(\mathbf{x}') = k(\mathbf{x}, \mathbf{x}')$ which is the unique reproducing kernel of $\mathcal{H}$. $\qquad\square$

**For all RRM functionals with RKHS hypotheses spaces, arbitrary convex losses L, and regularisers $R(\mathbf{f}) = Q(\|\mathbf{f}\|_{\mathcal{H}_K})$, where Q are strictly increasing functions, the minimisers can be always represented as $\mathbf{f}(\mathbf{x}) = \sum_i^n c_i k_{\mathbf{x}_i}(\mathbf{x})$**

*Proof.* We restate the RRM problem for reference

$$\min_{\mathbf{f} \in \mathcal{H}_K} \frac{1}{n} \sum_{i=1}^{n} L(y_i, \mathbf{f}(\mathbf{x}_i)) + \lambda Q(\|\mathbf{f}\|_{\mathcal{H}_K}) \ .$$

We follow the classical space decomposition approach (e.g. Schölkopf et al. (2001)). Any function $\mathbf{f} \in \mathcal{H}_K$ can be decomposed into $\mathbf{f} = \mathbf{f}_{\|} + \mathbf{f}_{\perp}$, where $\mathbf{f}_{\|}$ lies in the span of the kernel sections $k_{\mathbf{x}_i}$ centred at the $n$ training points, and $\mathbf{f}_{\perp}$ lies in its orthogonal complement.

The 1st term depends on the function $\mathbf{f}$ only through its evaluations at the training points $\mathbf{f}(\mathbf{x}_i)$, $\mathbf{x}_i \in \mathcal{S}_n$. For each training point $\mathbf{x}_i$ we have

$$\mathbf{f}(\mathbf{x}_i) = \left\langle \mathbf{f}, k_{\mathbf{x}_i} \right\rangle_{\mathcal{H}_K} = \left\langle \mathbf{f}_{\|} + \mathbf{f}_{\perp}, k_{\mathbf{x}_i} \right\rangle_{\mathcal{H}_K} = \left\langle \mathbf{f}_{\|}, k_{\mathbf{x}_i} \right\rangle_{\mathcal{H}_K} \ ,$$

where the last equality is the result of the orthogonality of the complement $\left\langle \mathbf{f}_{\perp}, k_{\mathbf{x}_i} \right\rangle_{\mathcal{H}_K} = 0$. By this the 1st term only depends on $\mathbf{f}_{\|}$.

For the 2nd term we have $\|\mathbf{f}\|_{\mathcal{H}_K}^2 = \left\| \mathbf{f}_{\|} + \mathbf{f}_{\perp} \right\|_{\mathcal{H}_K} = \sqrt{\left\| \mathbf{f}_{\|} \right\|_{\mathcal{H}_K}^2 + \|\mathbf{f}_{\perp}\|_{\mathcal{H}_K}^2}$ because $\left\langle \mathbf{f}_{\|}, \mathbf{f}_{\perp} \right\rangle_{\mathcal{H}_K} = 0$. Trivially, this is minimised when $\mathbf{f}_{\perp} = 0$ and therefore any strictly increasing $Q(\|\mathbf{f}\|_{\mathcal{H}_K})$ is also minimised when $\mathbf{f}_{\perp} = 0$. $\qquad\square$

### 1.A.1 Some further properties of operator-valued kernels

We indicate by $\mathbf{H_x} : \mathcal{Y} \to \mathcal{H}_K$ the linear operator defined as $\mathbf{H_x(y)} := \mathbf{H_x y}$. The norm of the linear operator $\mathbf{H_x}$ is $\|\mathbf{H_x}\| = \|\mathbf{H(x,x)}\|^{\frac{1}{2}}$.

*Proof.* From the definition of the operator norms

$$\|\mathbf{H_x}\| = \sup_{y \in \mathcal{Y}} \frac{\|\mathbf{H_x y}\|_{\mathcal{H}_K}}{\|\mathbf{y}\|_{\mathcal{Y}}} \quad \text{and} \quad \|\mathbf{H(x,x')}\| = \sup_{y \in \mathcal{Y}} \frac{\|\mathbf{H(x,x')y}\|_{\mathcal{Y}}}{\|\mathbf{y}\|_{\mathcal{Y}}} \tag{1.37}$$

$$\|\mathbf{H_x y}\|_{\mathcal{H}_K} \le \|\mathbf{y}\|_{\mathcal{Y}} \|\mathbf{H_x}\| \quad \text{and} \quad \|\mathbf{H(x,x')y}\|_{\mathcal{Y}} \le \|\mathbf{y}\|_{\mathcal{Y}} \|\mathbf{H(x,x')}\| \tag{1.38}$$

.

Further we have

$$\|\mathbf{H_x y}\|^2_{\mathcal{H}_K} = \langle \mathbf{H_x y}, \mathbf{H_x y} \rangle_{\mathcal{H}_K} \overset{\text{r.p.}}{=} \langle \mathbf{y}, \mathbf{H(x,x)y} \rangle_{\mathcal{Y}} \overset{\text{CS}}{\le} \|\mathbf{y}\|_{\mathcal{Y}} \|\mathbf{H(x,x)y}\|_{\mathcal{Y}} \overset{1.38}{\le} \|\mathbf{y}\|^2_{\mathcal{Y}} \|\mathbf{H(x,x)}\| \tag{1.39}$$

so that

$$\frac{\|\mathbf{H_x y}\|^2_{\mathcal{H}_K}}{\|\mathbf{y}\|^2_{\mathcal{Y}}} \le \|\mathbf{H(x,x)}\| \overset{1.37}{\implies} \|\mathbf{H_x}\| \le \|\mathbf{H(x,x)}\|^{\frac{1}{2}} \tag{1.40}$$

Similarly

$$\|\mathbf{H(x,x)y}\|^2_{\mathcal{Y}} = \langle \mathbf{H(x,x)y}, \mathbf{H(x,x)y} \rangle_{\mathcal{Y}} \overset{\text{r.p.}}{=} \langle \mathbf{H_x y}, \mathbf{H_x H(x,x)y} \rangle_{\mathcal{H}_K} \overset{\text{CS}}{\le} \|\mathbf{H_x y}\|_{\mathcal{H}_K} \|\mathbf{H_x H(x,x)y}\|_{\mathcal{H}_K}$$

From the 1st result in (1.38) we have

$$\|\mathbf{H_x H(x,x)y}\|_{\mathcal{H}_K} \le \|\mathbf{H(x,x)y}\|_{\mathcal{Y}} \|\mathbf{H_x}\|$$

and combining it directly with the 1st part of (1.38) we continue from the above

$$\|\mathbf{H(x,x)y}\|^2_{\mathcal{Y}} \le \|\mathbf{y}\|_{\mathcal{Y}} \|\mathbf{H_x}\|^2 \|\mathbf{H(x,x)y}\|_{\mathcal{Y}}$$

so that

$$\frac{\|\mathbf{H(x,x)y}\|_{\mathcal{Y}}}{\|\mathbf{y}\|_{\mathcal{Y}}} \le \|\mathbf{H_x}\|^2 \overset{1.37}{\implies} \|\mathbf{H(x,x)}\| \le \|\mathbf{H_x}\|^2 \tag{1.41}$$

From 1.40 and 1.41 we conclude $\|\mathbf{H}_x\| = \|\mathbf{H(x,x)}\|^{\frac{1}{2}}$. $\qquad\square$

**The norm of the matrix-valued kernel is $\|\mathbf{H}(\mathbf{x}, \mathbf{x}')\| \leq \|\mathbf{H}(\mathbf{x}', \mathbf{x}')\|^{\frac{1}{2}} \|\mathbf{H}(\mathbf{x}, \mathbf{x})\|^{\frac{1}{2}}$.**

*Proof.*

$$\left\| \mathbf{H}(\mathbf{x}, \mathbf{x}')\mathbf{y} \right\|_{\mathcal{Y}}^2 = \left\langle \mathbf{H}(\mathbf{x}, \mathbf{x}')\mathbf{y}, \mathbf{H}(\mathbf{x}, \mathbf{x}')\mathbf{y} \right\rangle_{\mathcal{Y}} \overset{\text{r.p.}}{=} \left\langle \mathbf{H}_{\mathbf{x}'}\mathbf{y}, \mathbf{H}_{\mathbf{x}}\mathbf{H}(\mathbf{x}, \mathbf{x}')\mathbf{y} \right\rangle_{\mathcal{H}_K} \overset{\text{CS}}{\leq} \|\mathbf{H}_{\mathbf{x}'}\mathbf{y}\|_{\mathcal{H}_K} \left\| \mathbf{H}_{\mathbf{x}}\mathbf{H}(\mathbf{x}, \mathbf{x}')\mathbf{y} \right\|_{\mathcal{H}_K}$$

From the 1st result in (1.38) we have

$$\left\| \mathbf{H}_{\mathbf{x}}\mathbf{H}(\mathbf{x}, \mathbf{x}')\mathbf{y} \right\|_{\mathcal{H}_K} \leq \left\| \mathbf{H}(\mathbf{x}, \mathbf{x}')\mathbf{y} \right\|_{\mathcal{Y}} \|\mathbf{H}_{\mathbf{x}}\|$$

and combining it directly with the 1st part of (1.38) we continue from the above

$$\left\| \mathbf{H}(\mathbf{x}, \mathbf{x}')\mathbf{y} \right\|_{\mathcal{Y}}^2 \leq \|\mathbf{y}\|_{\mathcal{Y}} \|\mathbf{H}_{\mathbf{x}'}\| \|\mathbf{H}_{\mathbf{x}}\| \|\mathbf{H}(\mathbf{x}, \mathbf{x})\mathbf{y}\|_{\mathcal{Y}}$$

so that

$$\left\| \mathbf{H}(\mathbf{x}, \mathbf{x}')\mathbf{y} \right\|_{\mathcal{Y}} \leq \|\mathbf{y}\|_{\mathcal{Y}} \|\mathbf{H}_{\mathbf{x}'}\| \|\mathbf{H}_{\mathbf{x}}\| \overset{1.37}{\implies} \left\| \mathbf{H}(\mathbf{x}, \mathbf{x}') \right\| \leq \|\mathbf{H}_{\mathbf{x}'}\| \|\mathbf{H}_{\mathbf{x}}\| = \left\| \mathbf{H}(\mathbf{x}', \mathbf{x}') \right\|^{\frac{1}{2}} \|\mathbf{H}(\mathbf{x}, \mathbf{x})\|^{\frac{1}{2}}$$

$\square$

**The evaluation functional $\delta_x : \mathcal{H}_K \to \mathbb{R}$ defined as $\delta_x(\mathbf{f}) := \mathbf{f}(\mathbf{x})$, $\forall \mathbf{f} \in \mathcal{H}_K$ is continuous (bounded).**

*Proof.*

$$\|\mathbf{f}(\mathbf{x})\|_{\mathcal{Y}}^2 = \langle \mathbf{f}(\mathbf{x}), \mathbf{f}(\mathbf{x}) \rangle_{\mathcal{Y}} \overset{\text{r.p.}}{=} \langle \mathbf{f}, \mathbf{H}_{\mathbf{x}}\mathbf{f}(\mathbf{x}) \rangle_{\mathcal{H}_K} \overset{\text{CS}}{\leq} \|\mathbf{f}\|_{\mathcal{H}_K} \|\mathbf{H}_{\mathbf{x}}\mathbf{f}(\mathbf{x})\|_{\mathcal{H}_K} \overset{1.38}{\leq} \|\mathbf{f}\|_{\mathcal{H}_K} \|\mathbf{H}_{\mathbf{x}}\| \|\mathbf{f}(\mathbf{x})\|_{\mathcal{Y}}$$

so that

$$\|\delta_x(\mathbf{f})\|_{\mathcal{Y}} = \|\mathbf{f}(\mathbf{x})\|_{\mathcal{Y}} \leq \|\mathbf{H}(\mathbf{x}, \mathbf{x})\|^{\frac{1}{2}} \|\mathbf{f}\|_{\mathcal{H}_K} \ .$$

(We can use $\|\mathbf{H}(\mathbf{x}, \mathbf{x})\|^{\frac{1}{2}} = C_{\mathbf{x}} \in [0, \infty)$ in definition 4.)

$\square$

# Chapter 2

# Learning Predictive Leading Indicators for Forecasting Time Series Systems with Unknown Clusters of Forecast Tasks

**Chapter abstract:** We present a new method for forecasting systems of multiple interrelated time series. The method learns the forecast models together with discovering leading indicators from within the system that serve as good predictors improving the forecast accuracy and a cluster structure of the predictive tasks around these. The method is based on the classical linear VAR and links the discovery of the leading indicators to inferring sparse graphs of Granger causality. We formulate a new constrained optimisation problem to promote the desired sparse structures across the models and the sharing of information amongst the learning tasks in a multi-task manner. We propose an algorithm for solving the problem and document on a battery of synthetic and real-data experiments the advantages of our new method over baseline VAR models as well as the state-of-the-art sparse VAR learning methods.

## 2.1 Introduction

Time series forecasting is vital in a multitude of application areas. With the increasing ability to collect huge amounts of data, users nowadays call for forecasts for large systems of series. On one hand, practitioners typically strive to gather and include into their models as many potentially helpful data as possible. On the other hand, the specific domain knowledge rarely provides sufficient understanding as to the relationships amongst the series and their importance for forecasting the system. This may lead to cluttering the forecast models with irrelevant data of little predictive benefit thus increasing the complexity of the models with possibly detrimental effects on the forecast accuracy (over-parametrisation and over-fitting).

In this chapter we focus on the problem of forecasting such large time series systems from their past evolution. We develop a new forecasting method that learns sparse structured models taking into account the unknown underlying relationships amongst the series. More specifically, the learned models use a limited set of series that the method identifies as useful for improving the predictive performance. We call such series the *leading indicators*.

In reality, there may be external factors from outside the system influencing the system developments. In this work we abstract from such external confounders for two reasons. First, we assume that any piece of information that could be gathered has been gathered and therefore even if an external confounder exists, there is no way we can get any data on it. Second, some of the series in the system may serve as surrogates for such unavailable data and we prefer to use these to the extent possible rather than chase the holy grail of full information availability.

We focus on the class of linear vector autoregressive models (VARs) which are simple yet theoretically well-supported, and well-established in the forecasting practice as well as the state-of-the-art time series literature, e.g. Lütkepohl (2005). The new method we develop falls into the broad category of graphical-Granger methods, e.g. Lozano et al. (2009), Shojaie and Michailidis (2010), and Songsiri (2013). Granger causality (Granger 1969) is a notion used for describing a specific type of dynamic dependency between time series. In brief, a series $\{\mathbf{y}_t\}$ Granger-causes series $\{\mathbf{z}_t\}$ if, given all the other relevant information, we can predict $\{\mathbf{z}_t\}$ more accurately when we use the history of $\{\mathbf{y}_t\}$ as an input in our forecast function. In our case, we call such series $\{\mathbf{y}_t\}$, that contributes to improving the forecast accuracy, the leading indicator.

For our method, we assume little to no prior knowledge about the structure of the time series systems. Yet, we do assume that most of the series in the system bring, in fact, no predictive benefit for the system, and that there are only few leading indicators whose inclusion into the forecast model as inputs improves the accuracy of the forecasts. Technically this assumption of only few leading indicators translates into a sparsity assumption for the forecast model, more precisely, sparsity in the connectivity of the associated Granger-causal graph.

An important subtlety for the model assumptions is that the leading indicators may not be *leading* for the whole system but only for some parts of it (certainly more realistic especially for lager systems). A series $\{\mathbf{y}_t\}$ may not Granger-cause all the other series in the system but only some of them. Nevertheless, if it contributes to improving the forecast accuracy of a group of series, we still consider it a leading indicator for this group. In this sense, we assume the system to be composed of clusters of series organised around their leading indicators. However, neither the identity of the leading indicators nor the composition of the clusters is known a priori.

To develop our method, we built on the paradigms of multi-task, e.g. Caruana (1997) and Evgeniou and Pontil (2004), and sparse structured learning, e.g. Bach et al. (2012). In order to achieve higher forecast accuracy our method encourages the tasks to borrow strength from one another during the model learning. More specifically, it intertwines the individual predictive tasks by shared structural constraints derived from the assumptions above.

To the best of our knowledge this is the first VAR learning method that promotes common sparse structures across the forecasting tasks of the time series system in order to improve the overall predictive performance. We designed a novel type of structured sparsity constraints coherent with the structural assumptions for the system, integrated them into a new formulation of a VAR optimisation problem, and proposed an efficient algorithm for solving it. The new formulation is unique in being able to discover clusters of series based on the structure of their predictive models concentrated around small number of leading indicators.

## 2.2 Preliminaries

In this section we briefly revisit some concepts we introduced in chapter 1 that are necessary for following the discussion in this chapter.

### 2.2.1 Linear vector autoregressive model

For a set of $K$ time series observed at $T$ synchronous equidistant time points we write the linear VAR in the form of a multi-output regression problem as $\mathbf{Y} = \mathbf{X}\mathbf{W} + \mathbf{E}$. Here $\mathbf{Y}$ is the $T \times K$ output matrix for $T$ observations and $K$ time series as individual 1-step-ahead forecasting tasks, $\mathbf{X}$ is the $T \times Kp$ input matrix so that each row $\mathbf{x}_{t,.}$ is a $Kp$ long vector with $p$ lagged values of the K time series as inputs $\mathbf{x}_{t.} = (y_{t-1,1}, y_{t-2,1}, \ldots, y_{t-p,1}, y_{t-1,2}, \ldots, y_{t-p,K})^T$, and $\mathbf{W}$ is the corresponding $Kp \times K$ parameters matrix where each column $\mathbf{w}_{.k}$ is a model for a single time series forecasting task (see figure 2.2.1a). We follow the standard time series assumptions: the $T \times K$ error matrix $\mathbf{E}$ is a random noise matrix with IID rows with zero mean and a diagonal covariance; the time series are second order stationary and centred (so that we can omit the intercept).

In principle, we can estimate the model parameters by minimising the standard squared error loss

$$L(\mathbf{W}) := \sum_{t=1}^{T} \sum_{k=1}^{K} (y_{t,k} - \langle \mathbf{w}_{.k}, \mathbf{x}_{t.} \rangle)^2 \tag{2.1}$$

which corresponds to maximising the likelihood with IID Gaussian errors and spherical covariance. However, since the dimensionality $Kp$ of the regression problem quickly grows with the number of series $K$ (by a multiple of $p$), often even relatively small VARs suffer from over-parametrisation ($Kp \gg T$). Yet, typically not all the past of all the series is indicative of the future developments of the whole system. In this respect the VARs are typically sparse.

In practice, the univariate autoregressive model (AR) which uses as input for each time series forecast model only its own history (and thus is an extreme sparse version of VAR), is often difficult to beat by any VAR model with the complete input sets. A variety of approaches such as Bayesian or regularisation techniques have been successfully used in the past to promote sparsity and condition the model learning. Those most relevant to our work are discussed in section 2.4.

### 2.2.2 Granger-causality graphs

Granger (1969) proposed a practical definition of causality in time series based on the accuracy of least-squares predictor functions. In brief, for two time series $\{\mathbf{y}_t\}$ and $\{\mathbf{z}_t\}$, we say that $\{\mathbf{y}_t\}$ Granger causes $\{\mathbf{z}_t\}$ if, given all the other relevant information, a predictor function using the history of $\{\mathbf{y}_t\}$ as input can forecast $\{\mathbf{z}_t\}$ better (in the mean-square sense) than a function not using it. Similarly, a set of time series $\left\{\{\mathbf{y}_t^{(1)}\}, \ldots, \{\mathbf{y}_t^{(l)}\}\right\}$ G-causes series $\{\mathbf{z}_t\}$ if it can be predicted better using the past values of the set.

The G-causal relationships can be described by a directed graph $\mathcal{G} = \{\mathcal{V}, \mathcal{E}\}$ (Eichler 2012), where each node $v \in \mathcal{V}$ represents a time series in the system, and the directed edges represent the G-causal relationships between the series. In VARs the G-causality is captured within the $\mathbf{W}$ parameters matrix. When any of the parameters of the $k$-th task ($k$-th column of the $\mathbf{W}$) referring to the $p$ past values of the $l$-th input series is non-zero, we say that the $l$-th series G-causes series $k$, and we denote this in the G-causal graph by a directed edge $e_{l,k}$ from $v_l$ to $v_k$.

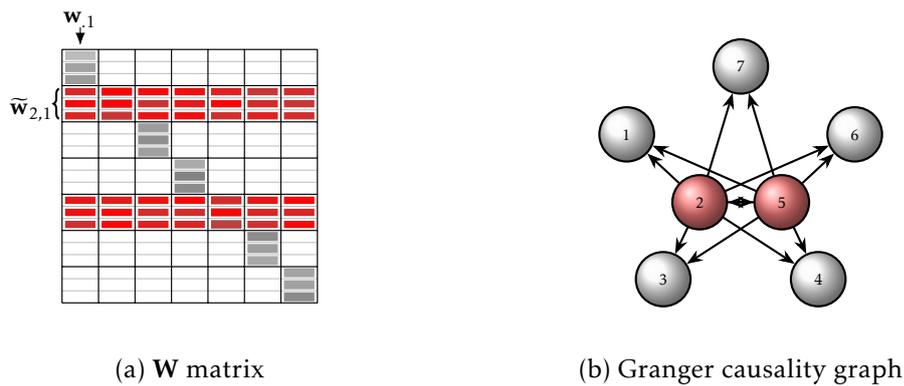

(a) $\mathbf{W}$ matrix     (b) Granger causality graph

Figure 2.2.1: In linear VARs the parameter matrix $\mathbf{W}$ can be seen as an adjacency matrix of the Granger causality graph.

Figure 2.2.1 shows a schema of the VAR parameters matrix $\mathbf{W}$ and the corre-sponding G-causal graph for an example system of $K = 7$ series with the number of lags $p = 3$. In 2.2.1a the gray and red cells are the non-zero elements, in 2.2.1b the circle nodes are the individual time series, the arrow edges are the G-causal links between the series[1]. For example, the arrow from 2 to 1 indicates that series 2 G-causes series 1; correspondingly the cells for the 3 lags in the 2nd block-row and the 1th column are coloured ($\widetilde{\mathbf{w}}_{2,1}$). Series 2 and 5 are the leading indicators for the whole system, their block-rows are coloured in all columns in the $\mathbf{W}$ matrix schema and they have out-edges to all other nodes in the G-graph.

One may question if calling the above notion *causality* is appropriate. Indeed, unlike that other perhaps more philosophical approaches, e.g. Pearl 2009, it does not really seek to understand the underlying forces driving the relationships between the series. Instead, the concept is purely technical based on the series contribution to the predictive accuracy, ignoring also possible confounding effects of unobservables. Nevertheless, the term is well established in the time series community. Moreover, it fits very well our purposes, where the primary objective is to learn models with high forecast accuracy that use as inputs only those time series that contribute to improving the accuracy - the leading indicators. Therefore, acknowledging all the reservations, we stick to it in this chapter always preceding it by Granger or G- to avoid confusion.

## 2.3 Learning VARs with clusters around leading indicators

We present here our new method for learning VAR models with task Clustering around Leading indicators (CLVAR). The method relies on the assumption that the generating process is sparse in the sense of there being only a few leading indicators within the system having an impact on the future developments. The leading indicators may be useful for predicting all or only some of the series in the systems. In this respect the series are clustered around their G-causing leading indicators. However, the method does not need to know the identity of the leading indicators nor the cluster assignments a priori and instead learns these together with the predictive models.

---

[1] The self-loops corresponding to the gray block-diagonal elements in $\mathbf{W}$ are omitted for clarity of display.

In building our method we exploited the multi-task learning ideas Caruana (1997) and let the models benefit from learning multiple tasks together (one task per series). This is in stark contrast to other state-of-the-art VAR and graphical-Granger methods, e.g. Arnold et al. (2007), Lozano et al. (2009), and Liu and Bahadori (2012). Albeit them being initially posed as multi-task (or multi-output) problems, due to their simple additive structure they decompose into a set of single-task problems solvable independently without any interaction and information sharing during the per-task learning. We, on the other hand, encourage the models to share information and borrow strength from one another in order to improve the overall performance by intertwining the model learning via structural constraints on the models derived from the assumptions outlined above.

### 2.3.1 Leading indicators for whole system

For the sake of exposition we first concentrate on a simplified problem of learning a VAR with leading indicators shared by the whole system (without clustering). The structure we assume here is the one illustrated in figure 2.2.1b. We see that the parameters matrix $\mathbf{W}$ is sparse with non-zero elements only in the block-rows corresponding to the lags of the leading indicators for the system (series 2 and 5 in the example in figure 2.2.1a) and on the block diagonal. The block-diagonal elements of $\mathbf{W}$ are associated with the lags of each series serving as inputs for predicting its own 1-step-ahead future. It is a stylised fact that the future of a stationary time series depends first and foremost on its own past developments. Therefore in addition to the leading indicators we want each of the individual series forecast function to use its own past as a relevant input. We bring the above structural assumptions into the method by formulating novel fit-for-purpose constraints for learning VAR models with multi-task structured sparsity.

**Learning problem and algorithm for learning without clusters**

We first introduce some new notation to accommodate for the necessary block structure across the lags of the input series in the input matrix $\mathbf{X}$ and the corresponding elements of the parameters matrix $\mathbf{W}$. For each input vector $\mathbf{x}_{t.}$ (a row of $\mathbf{X}$) we indicate by $\widetilde{\mathbf{x}}_{t,j} = (y_{t-1,j}, y_{t-2,j}, \ldots, y_{t-p,j})^T$ the $p$-long sub-vector of $\mathbf{x}_{t.}$ referring to the history (the $p$ lagged values preceding time $t$) of the series $j$, so that for the whole row we have $\mathbf{x}_{t.} = (x_{t,1}, \ldots, x_{t,Kp})^T = (\widetilde{\mathbf{x}}_{t,1}^T, \ldots, \widetilde{\mathbf{x}}_{t,K}^T)^T$. Correspondingly, in each model vector $\mathbf{w}_{.k}$ (a column of $\mathbf{W}$), we indicate by $\widetilde{\mathbf{w}}_{j,k}$ the $p$-long sub-vector of the $k$th model parameters associated with the input subvector $\widetilde{\mathbf{x}}_{t,j}$. In figure 2.2.1a, $\widetilde{\mathbf{w}}_{2,1}$ is the block of the 3 shaded parameters in column 1 and rows $\{4, 5, 6\}$ - the block of parameters of the model for forecasting the 1st time series associated with the 3 lags of the 2nd time series (a leading indicator) as inputs. Using these blocks of inputs $\widetilde{\mathbf{x}}_{t,j}$ and parameters $\widetilde{\mathbf{w}}_{j,k}$ we can rewrite the inner products in the loss in (2.1) as $\langle \mathbf{w}_{.k}, \mathbf{x}_{t.} \rangle = \sum_{b=1}^{K} \langle \widetilde{\mathbf{w}}_{b,k}, \widetilde{\mathbf{x}}_{t,b} \rangle$.

Next, we associate each of the parameter blocks with a single non-negative scalar $\gamma_{b,k}$ so that $\widetilde{\mathbf{w}}_{b,k} = \gamma_{b,k} \widetilde{\mathbf{v}}_{b,k}$. The $Kp \times K$ matrix $\mathbf{V}$, composed of the blocks $\widetilde{\mathbf{v}}_{b,k}$ in the same way as $\mathbf{W}$ is composed of $\widetilde{\mathbf{w}}_{b,k}$, is therefore just a rescaling of the original $\mathbf{W}$ with the weights $\gamma_{b,k}$ used for each block. With this new re-parametrization the squared-error loss (2.1) is

$$L(\mathbf{W}) = \sum_{t=1}^{T} \sum_{k=1}^{K} (y_{t,k} - \sum_{b=1}^{K} \gamma_{b,k} \langle \widetilde{\mathbf{v}}_{b,k}, \widetilde{\mathbf{x}}_{t,b} \rangle)^2. \tag{2.2}$$

Finally, we use the non-negative $K \times K$ weight matrix $\mathbf{\Gamma} = \{\gamma_{b,k} \mid b, k = 1, \ldots, K\}$ to formulate our multi-task structured sparsity constraints. In $\mathbf{\Gamma}$ each element corresponds to a single series serving as an input to a single predictive model. A zero weight $\gamma_{b,k} = 0$ results in a zero parameter sub-vector $\widetilde{\mathbf{w}}_{b,k} = \mathbf{0}$ and therefore the corresponding input sub-vectors $\widetilde{\mathbf{x}}_{t,b}$ (the past lags of series $b$ for each time point $t$) have no effect in the predictive functions for task $k$.

Our assumption of only small number of leading indicators means that most series shall have no predictive effect for any of the tasks. This can be achieved by $\mathbf{\Gamma}$ having most of its rows equal to zero. On the other hand, the non-zero elements corresponding to the leading indicators shall form full rows of $\mathbf{\Gamma}$. As explained in section 2.3.1, in addition to the leading indicators we also want each series past to serve as an input to its own forecast function. This translates to non-zero diagonal elements $\gamma_{i,i} \neq 0$. To combine these two contradicting structural requirements onto $\mathbf{\Gamma}$ (sparse rows vs. non-zero diagonal) we construct the matrix from two same size matrices $\mathbf{\Gamma} = \mathbf{A} + \mathbf{B}$, one for each of

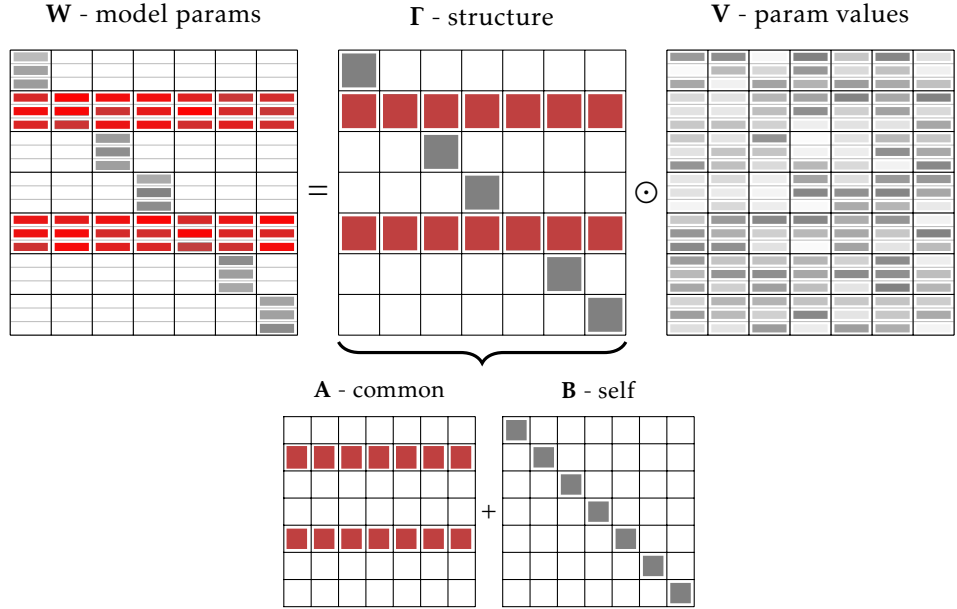

Figure 2.3.1: The parameter matrix $\mathbf{W}$ is decomposed into the structural matrix $\mathbf{\Gamma}$ and the parameter scales $\mathbf{V}$. The structure in $\mathbf{\Gamma}$ is further split into matrix $\mathbf{A}$ capturing the dependency on the leading indicators and $\mathbf{B}$ capturing the dependency of each constituent series on its own history.

the structures: $\mathbf{A}$ for the row-sparse of leading indicators, $\mathbf{B}$ for the diagonal of the own history, figure 2.3.1

We now formulate the optimisation problem for learning VAR with shared leading indicators across the whole system and dependency on own past as the constrained minimisation

$$\arg\min_{\mathbf{A},\mathbf{V}} \sum_{t=1}^{T}\sum_{k=1}^{K}(y_{t,k}-\sum_{b=1}^{K}(\alpha_{b,k}+\beta_{b,k})\langle\widetilde{\mathbf{v}}_{b,k},\widetilde{\mathbf{x}}_{t,j}\rangle)^2 + \lambda\|\mathbf{V}\|_F^2 \quad (2.3)$$
$$\text{s.t.} \quad \mathbf{1}_K^T\,\overline{\boldsymbol{\alpha}}=\kappa;\;\overline{\boldsymbol{\alpha}}\geq\mathbf{0};\;\boldsymbol{\alpha}_{.j}=\overline{\boldsymbol{\alpha}},\;\beta_{j,j}=1-\alpha_{j,j}\;\forall j=1,\dots,K \quad ,$$

where the links between the matrices $\mathbf{A}$, $\mathbf{B}$, $\mathbf{\Gamma}$, $\mathbf{V}$ and the parameter matrix $\mathbf{W}$ of the VAR model are explained in the paragraphs above.

In equation (2.3) we force all the columns of $\mathbf{A}$ to be equal to the same vector $\overline{\boldsymbol{\alpha}}^2$, and we promote the sparsity in this vector by constraining it onto a simplex of size $\kappa$. Here $\kappa$ controls the relative weight of each series own past vs. the past of all the neighbouring series. For identifiability reasons we force the diagonal elements of $\mathbf{\Gamma}$ to equal unity by scaling appropriately the diagonal $\beta_{j,j}$ elements. Lastly, while $\mathbf{\Gamma}$ is constructed and constrained to control for the structure of the learned models (as per our assumptions), the actual value of the final parameters $\mathbf{W}$ is the result of combining it with the other learned matrix $\mathbf{V}$. To confine the overall complexity of the final model $\mathbf{W}$ we impose a standard ridge penalty (Hoerl and Kennard 1970) on the model parameters $\mathbf{V}$.

---

[2]This does not excessively limit the capacity of the models as the final model matrix $\mathbf{W}$ is the result of combining $\mathbf{\Gamma}$ with the learned matrix $\mathbf{V}$.

The optimisation problem (2.3) is jointly non-convex, however, it is convex with respect to each of the optimisation variables with the other variable fixed. Therefore we propose to solve it by an alternating descent for $\mathbf{A}$ and $\mathbf{V}$ as outlined in algorithm 1 below. $\mathbf{B}$ is solved trivially applying directly the equality constraint of (2.3) over the learned matrix $\mathbf{A}$ as $\mathbf{B} = \mathbf{I} - diag(\mathbf{A})$ which implies $\mathbf{\Gamma} = \mathbf{A} + \mathbf{B} = \mathbf{A} - diag(\mathbf{A}) + \mathbf{I}$.

---

**Input** : training data $\mathbf{Y}$, $\mathbf{X}$; hyper-parameters $\lambda$, $\kappa$
**Initialise:** $\overline{\boldsymbol{\alpha}}$ evenly to satisfy constraints in all columns of $\mathbf{A}$; $\mathbf{\Gamma} \leftarrow \mathbf{A} - diag(\mathbf{A}) + \mathbf{I}$

**repeat**  // Alternating descent
    **begin** Step 1: Solve for $\mathbf{V}$
        **foreach** *task k* **do**
            re-weight input blocks $\mathbf{z}_{t,b}^{(k)} \leftarrow \gamma_{b,k}\, \widetilde{\mathbf{x}}_{t,b}$     $\forall$ time point $t$ and input series $b$
            $\mathbf{v}_{.k} \leftarrow \arg\min_{\mathbf{v}} \left\| \mathbf{y}_{.k} - \mathbf{Z}^{(k)}\mathbf{v} \right\|_2^2 + \lambda \left\| \mathbf{v} \right\|_2^2$     // standard ridge regression
        **end**
    **end**
    **begin** Step 2: Solve for $\mathbf{A}$ and $\mathbf{\Gamma}$
        **foreach** *task k* **do**
            input products $h_{t,b}^{(k)} \leftarrow \langle \widetilde{\mathbf{v}}_{b,k}, \widetilde{\mathbf{x}}_{t,b} \rangle$     $\forall$ time point $t$ and input series $b$
            task residuals after using own history $r_{t,k} \leftarrow y_{t,k} - h_{t,k}^{(k)}$     $\forall$ time point $t$
            remove own history from input products $h_{t,k}^{(k)} \leftarrow 0$     $\forall$ time point $t$
        **end**
        concatenate vertically input product matrices $\mathbf{H} = vertcat(\mathbf{H}^{(.)})$
        $\overline{\boldsymbol{\alpha}} \leftarrow \arg\min_{\overline{\boldsymbol{\alpha}}} \| vec(\mathbf{R}) - \mathbf{H}\, \overline{\boldsymbol{\alpha}} \|_2^2$, s.t. $\overline{\boldsymbol{\alpha}}$ on simplex     // projected grad descent
        put $\overline{\boldsymbol{\alpha}}$ to all columns of $\mathbf{A}$; $\mathbf{\Gamma} \leftarrow \mathbf{A} - diag(\mathbf{A}) + \mathbf{I}$
    **end**
**until** *objective convergence*;

**Algorithm 1:** Alternating descent for VAR with system-shared leading indicators

---

To foster the intuition behind our method we provide links to other well-known learning problems and methods. First, we can rewrite the weighted inner product in the loss function (2.2) as $\langle \widetilde{\mathbf{v}}_{b,k}, \gamma_{b,k}\widetilde{\mathbf{x}}_{t,b}\rangle$. In this "feature learning" formulation the weights $\gamma_{b,k}$ act on the original inputs and, hence, generate new task-specific features $\mathbf{z}_{t,b}^{(k)} = \gamma_{b,k}\widetilde{\mathbf{x}}_{t,b}$. These are actually used in Step 1 of our algorithm 1. Alternatively, we can express the ridge penalty on $\mathbf{V}$ used in eq. (2.3) as $\|\mathbf{V}\|_F^2 = \sum_{b,k}\left\|\widetilde{\mathbf{v}}_{b,k}\right\|_2^2 = \sum_{b,k} 1/\gamma_{b,k}^2 \left\|\widetilde{\mathbf{w}}_{b,k}\right\|_2^2$. In this "adaptive ridge" formulation the elements of $\mathbf{\Gamma}$, which in our methods we learn, act as weights for the $\ell_2$ regularization of $\mathbf{W}$. Equivalently, we can see this as the Bayesian maximum-a-posteriori with Guassian priors where the elements of $\mathbf{\Gamma}$ are the learned priors for the variance of the model parameters or (perhaps more interestingly) the random errors.

### 2.3.2 Leading indicators for clusters of predictive tasks

After explaining in section 2.3.1 the simplified case of learning a VAR with leading indicators for the whole system, we now move onto the more complex (and for larger VARs certainly more realistic) setting of the leading indicators being predictive only for parts of the system - clusters of predictive tasks.

To get started we briefly consider the situation in which the cluster structure (not the leading indicators) is known a priori. Here the models could be learned by a simple modification of algorithm 1 where in step 2 we would work with cluster-specific vectors $\overline{\boldsymbol{\alpha}}$ and matrices $\mathbf{H}$ and $\mathbf{R}$ constructed over the known cluster members. In reality the clusters are typically not known and therefore our CLVAR method is designed to learn them together with the leading indicators.

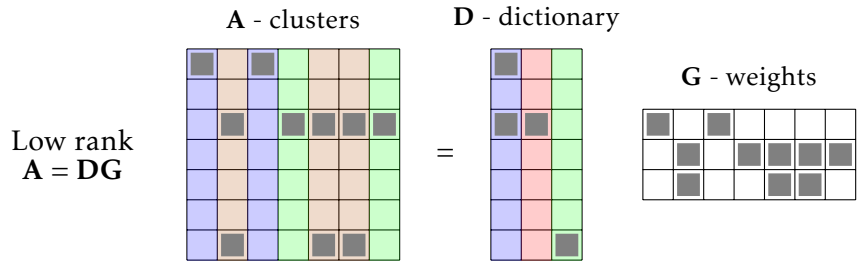

Figure 2.3.2: The soft-cluster dependency structure in matrix $\mathbf{A}$ is constructed from a sparse dictionary matrix $\mathbf{D}$ and a sparse weight matrix $\mathbf{G}$.

We use the same block decompositions of the input and parameter matrices $\mathbf{X}$ and $\mathbf{W}$, and the structural matrices $\boldsymbol{\Gamma} = \mathbf{A} + \mathbf{B} = \mathbf{A} - diag(\mathbf{A}) + \mathbf{I}$ and the rescaled parameter matrix $\mathbf{V}$ defined in section 2.3.1. However, we need to alter the structural assumptions encoded into the matrix $\mathbf{A}$. In the cluster case $\mathbf{A}$ still shall have many rows equal to zero but it shall no longer have all the columns equal (same leading indicators for all the tasks). Instead, we learn it as a low rank matrix by factorizing it into two lower dimensional matrices $\mathbf{A} = \mathbf{DG}$: the $K \times r$ dictionary matrix $\mathbf{D}$ with the dictionary atoms (columns of $\mathbf{D}$) representing the cluster prototypes of the dependency structure; and the $r \times K$ matrix $\mathbf{G}$ with the elements being the per-model dictionary weights, $1 \le r \le K$.

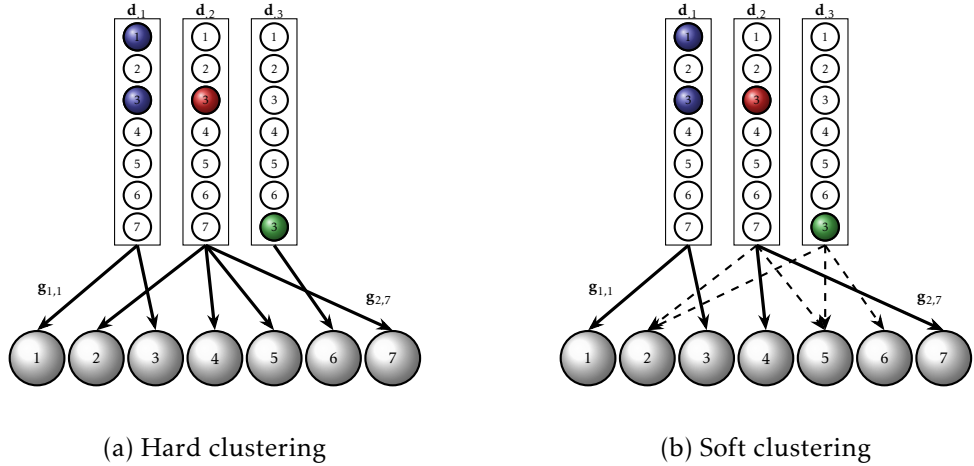

(a) Hard clustering                 (b) Soft clustering

Figure 2.3.3: Each column of matrix $\mathbf{D}$ is a prototipical dictionary atom of the dependency structure. In the hard clustering each task belongs to a cluster specifed by a single prototype, in the soft clustering each task can use multiple prototypes in a sparse convex combination.

To better understand the clustering effect of the low-rank decomposition, figure 2.3.3 illustrates it for an imaginary system of $K = 7$ time series with rank $r = 3$. The $\mathbf{d}_{.j}$ j={1,2,3} columns in the top are the sparse cluster prototypes (the non-zero elements for the leading indicators are shaded). The circles in the bottom are the individual learning tasks and the arrows are the per-model dictionary weights $g_{i,j}$. Solid arrows have weight 1, missing arrows have weight zero, dashed arrows have weight between 0 and 1. So for example, the solid arrow from the 2nd column to the 7th circle in figure 2.3.3b is the $g_{2,7}$ element of matrix $\mathbf{G}$. Since it is a full arrow, it is equal to 1. The arrow from the 3rd column to the 2nd circle in figure 2.3.3b is the $g_{3,2}$ element of $\mathbf{G}$. Since the arrow is dashed, we have $0 < g_{3,2} < 1$.

Figure 2.3.3a uses a binary matrix $\mathbf{G}$ (no dashed arrows) reflecting hard clustering of the tasks consistent with our initial setting of a priori known clusters. Each task (circle at the bottom) is associated with only one cluster prototype (columns of $\mathbf{D}$ in the top). In contrast, figure 2.3.3b uses matrix $\mathbf{G}$ with elements between 0 and 1 to perform soft clustering of the tasks. Each task (circle at the bottom) may be associated with more than one cluster prototype (columns of $\mathbf{D}$ in the top). Our CLVAR is based on this latter approach of soft-clustering of the forecast tasks.

**Learning problem and algorithm for CLVAR**

We now adapt the minimisation problem (2.3) for the multi-cluster setting

$$\arg\min_{\mathbf{D},\mathbf{G},\mathbf{V}} \sum_{t=1}^{T} \sum_{k=1}^{K} \left( y_{t,k} - \sum_{b=1}^{K} (\sum_{j=1}^{K} d_{b,j} g_{j,k} + \beta_{b,k}) \left\langle \widetilde{\mathbf{v}}_{b,k}^{T}, \widetilde{\mathbf{x}}_{t,j}^{T} \right\rangle \right)^2 + \lambda \| \mathbf{V} \|_F^2 \quad (2.4)$$
$$\text{s.t. } \mathbf{1}_K^T \mathbf{d}_{.j} = \kappa; \ \mathbf{d}_{.j} \geq \mathbf{0}; \ \mathbf{1}_r^T \mathbf{g}_{.j} = 1; \ \mathbf{g}_{.j} \geq \mathbf{0}, \ \beta_{j,j} = 1 - \alpha_{j,j} \ \forall j \ .$$

The relations of the optimisation matrices $\mathbf{D}, \mathbf{G}, \mathbf{V}$ to the parameter matrix $\mathbf{W}$ of the VAR model are as explained in the paragraphs above. The principal difference of the formulation (2.4) as compared to problem (2.3) is the low-rank decomposition of matrix $\mathbf{A} = \mathbf{DG}$ using the fact that $a_{b,k} = \sum_{j=1}^{K} d_{b,j} g_{j,k}$. Similarly as for the single column $\overline{\boldsymbol{\alpha}}$ in (2.3) we promote sparsity in the cluster prototypes $\mathbf{d}_{.j}$ by constraining them onto the simplex. And we use the probability simplex constraints to sparsify the per-task weights in the columns of $\mathbf{G}$ so that the task are not based on all the prototypes. Figure 2.3.2 shows an example of such a sparse low-rank decomposition, only the shaded cells are non-zero.

We propose to solve problem (2.4) by alternating descent algorithm 2. While non-convex, the alternating approach for learning the low-rank matrix decomposition is known to perform well in practice and has been recently supported by new theoretical guarantees, e.g. Park et al. (2016). We solve the two subproblems in step 2 by projected gradient descent with FISTA backtracking line

**Input** : training data $\mathbf{Y}, \mathbf{X}$; hyper-parameters $\lambda, \kappa, r$
**Initialise:** $\mathbf{D}, \mathbf{G}$ evenly to satisfy the constraints; $\mathbf{A} \leftarrow \mathbf{DG}$; $\boldsymbol{\Gamma} \leftarrow \mathbf{A} - diag(\mathbf{A}) + \mathbf{I}$

**repeat**                                                  // Alternating descent
  **begin** Step 1: Solve for $\mathbf{V}$
    |   same as in algorithm 1
  **end**
  **begin** Step 2: Solve for $\mathbf{D}, \mathbf{G}$ and $\boldsymbol{\Gamma}$
    **foreach** *task k* **do**
        same as in algorithm 1
        $\mathbf{g_k} \leftarrow \arg\min_{\mathbf{g}} \|\mathbf{r}_{\cdot k} - \mathbf{H}^{(k)}\mathbf{g}\|_2^2$, s.t. $\mathbf{g}$ on simplex     // projected grad desc
    **end**
    concatenate vertically input product matrices $\mathbf{H} = vertcat(\mathbf{H}^{(\cdot)})$
    expand matrices to match dictionary vectorization $\widehat{\mathbf{G}} \leftarrow \mathbf{G}^T \otimes \mathbf{1}_T \mathbf{1}_K^T; \widehat{\mathbf{H}} = \mathbf{1}_r^T \otimes \mathbf{H}$
    $vec(\mathbf{D}) \leftarrow \arg\min_{\mathbf{D}} \|vec(\mathbf{R}) - \widehat{\mathbf{G}} \odot \widehat{\mathbf{H}} vec(\mathbf{D})\|_2^2$     // projected grad desc
                   s.t. $\mathbf{d}_{\cdot j}$ on simplex $\forall j$
    $\mathbf{A} = \mathbf{DG}$; $\boldsymbol{\Gamma} \leftarrow \mathbf{A} - diag(\mathbf{A}) + \mathbf{I}$
  **end**
**until** *objective convergence*;

**Algorithm 2:** CLVAR - VAR with leading indicators for clusters of predictive tasks

search Beck and Teboulle (2009). The algorithm is $\mathcal{O}(T)$ for increasing number of observation and $\mathcal{O}(K^3)$ for increasing number of time series. However, one needs to bear in mind that with each additional series the complexity of the VAR model itself increases by $\mathcal{O}(K)$. Nevertheless, the expensive scaling with $K$ is an important bottleneck of our method and we are investigating options to address it in our future work.

## 2.4 Related Work

We explained in section 2.2.2 how our search for leading indicators links to the Granger causality discovery in VARs. As shows the list of references in the survey of Liu and Bahadori (2012), this has been a rather active research area over the last several years. While the traditional approach for G-discovery was based on pairwise testing of candidate models or the use of model selection criteria such as AIC or BIC, inefficiency of such approaches for builidng predictive models of large time series system has long been recognised[3], e.g. Doan et al. (1984).

---

[3]Due to the lack of domain knowledge to support the model selection and combinatorial complexity of exhaustive search.

As an alternative, variants of so-called graphical Granger methods based on regularization for parameter shrinkage and thresholding along the lines of Lasso (Tibshirani 1996) have been proposed in the literature. We use the two best-established ones, the lasso-Granger (VARL1) of Arnold et al. (2007) and the grouped-lasso-Granger (VARLG) of Lozano et al. (2009), as the state-of-the-art competitors in our experiments. More recent adaptations of the graphical Granger method address the specific problems of determining the order of the models and the G-causality simultaneously (Shojaie and Michailidis 2010; Ren et al. 2013), the G-causality inference in irregular (Bahadori and Liu 2012) and subsampled series (Gong et al. 2015), and in systems with instantaneous effects (Peters et al. 2013). However, neither of the above methods considers or exploits any common structures in the G-causality graphs as we do in our method.

Common structures in the dependency are assumed by Jalali and Sanghavi (2012) and Geiger et al. (2015) though the common interactions are with unobserved variables from outside the system rather then within the system itself. Also, the methods discussed in these have no clustering ability. Songsiri (2015) considers common structures across several datasets (in panel data setting) instead of within the dynamic dependencies of a single dataset. Huang and Schneider (2012) assume sparse bi-clustering of the G-graph nodes (by the in- and out- edges) to learn fully connected sub-graphs in contrast to our shared sparse structures. Most recently, Hong et al. (2017) proposes to learn clusters of series by Laplacian clustering over the sparse model parameters. However, the underlying models are treated independently not encouraging any common structures at learning.

More broadly, our work builds on the multi-task (Caruana 1997) and structured sparsity (Bach et al. 2012) learning techniques developed outside the time-series settings. Similar block-decompositions of the feature and parameter matrices as we use in our methods have been proposed to promote group structures across multiple models (Argyriou et al. 2007; Swirszcz and Lozano 2012). Although the methods developed therein have no clustering capability. Various approaches for learning model clusters are discussed in Bakker and Heskes (2003), Xue et al. (2007), Jacob et al. (2009), Kang et al. (2011), and Kumar and Hal Daume III (2012) of which the latest uses similar low-rank decomposition approach as our method. Nevertheless, neither of these approaches learns sparse models and builds the clustering on similar structural assumptions as our method does.

## 2.5 Experiments

We present here the results of a set of experiments on synthetic and real-world datasets. We compare to relevant baseline methods for VAR learning: univariate auto-regressive model AR (though simple, AR is typically hard to beat by high-dimensional VARs when the domain knowledge cannot help to specify a relevant feature subset for the VAR model), VAR model with standard $\ell_2$ regularisation VARL2 (controls over-parametrisation by shrinkage but does not yield sparse models), VAR model with $\ell_1$ regularisation VARL1 (lasso-Granger of Arnold et al. (2007)), and VAR with group lasso regularisation VARLG (grouped-lasso-Granger of Lozano et al. (2009)). We implemented all the methods in Matlab using standard state-of-the-art approaches: trivial analytical solutions for AR and VARL2, FISTA proximal-gradient (Beck and Teboulle 2009) for VARL1 and VARLG. The full code together with the datasets amenable for full replication of our experiments is available from https://bitbucket.org/dmmlgeneva/var-leading-indicators.

In all our experiments we simulated real-life forecasting exercises. We split the analysed datasets into training and hold-out sets unseen at learning and only used for performance evaluation. The trained models were used to produce one-step ahead forecasts by sliding through all the points in the hold-out. We repeated each experiments over 20 random resamples. The reported performance is the averages over these 20 resamples. The construction of the resamples for the synthetic and real datasets is explained in the respective sections below. We used 3-folds cross-validation with mean squared error (MSE) as the criterion for the hyper-parameter grid search. Unless otherwise stated below, the grids were: $\lambda \in 15$-elements grid $[10^{-4} \dots 10^3]$ (used also for VARL2, VARL1 and VARLG), $\kappa \in \{0.5, 1, 2\}$, rank $\in \{1, 0.1K, 0.2K, K\}$. We preprocessed all the data by zero-centering and unit-standardization based on the training statistics only.

For all the experiments and all the tested methods we fixed the lag of the learned models to $p = 5$. While the search for the best lag $p$ has in the past constituted an important part of time series modelling[4], in high-dimensional settings the exhaustive search through possible sub-set model combinations is clearly impractical. Modern methods therefore focus on using VARs with sufficient number of lags to cater for the underlying time dependency and apply Bayesian or regularization methods to control the model complexity, e.g. Koop (2013). In our case, this is achieved by the ridge shrinkage on the parameter matrix $\mathbf{V}$.

### 2.5.1 Synthetic Experiments

We designed six generating processes for systems varying by number of series and the G-causal structure. The first three are small systems with $K = 10$ series only, the next three increase the size to $K = \{30, 50, 100\}$. Systems 1 and 2 are unfavourable for our method, generated by processes not corresponding to our structural assumptions: in the 1st each series is generated from its own past only and therefore can be best modelled by a simple univariate AR model (the G-causal graph has no links); the 2nd is a fully connected VAR (all series are leading for the whole system). The 3rd system consists of 2 clusters with 5 series each, both depending on 1 leading indicator. Systems 4-6 are composed of $\{3, 5, 10\}$ clusters respectively, each with 10 series concentrated around 2 leading indicators[5].

For each of the 6 system designs we first generated a random matrix of VAR coefficients with the required structure. We ensured the processes are stationary by controlling the roots of the model characteristic polynomials. We then generated 20 random realisation of the VAR processes with uncorrelated standard-normal noise. In each, we separated the last 500 observations into a hold-out set and used the previous $T$ observations for training. Once trained, the same model was used for the 1-step-ahead forecasting of the 500 hold-out points by sliding forward through the dataset.

The predictive performance of the methods in the 6 experimental settings for multiple training sizes $T$ is summarised in figure 2.5.1[6]. We measure the predictive accuracy by the MSE of 1-step-ahead forecasts relative to the forecasts produced by the VAR with the true generative coefficients (RelMSE). Doing so we standardize the MSE by the irreducible error of each of the forecast exercises. The closer to 1 (the gold standard) the better. The plots display the average RelMSE over the twenty replications of the experiments, the error bars are at $\pm 1$ standard deviation.

---

[4]Especially for univariate models within the context of the more general ARMA class (Box et al. 1994).
[5]For the last two, we fixed the rank in CLVAR training to the true number of clusters.

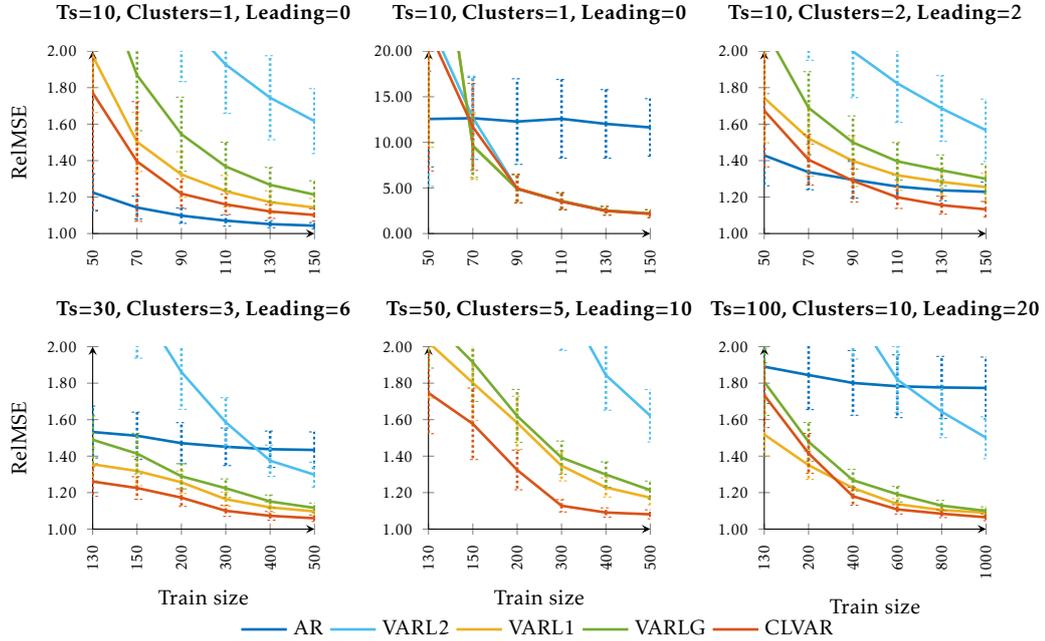

Figure 2.5.1: Synthetic experiments: Average prediction error (MSE relative to oracle predictions) across the 20 experimental replications, the error bars are at 1 standard deviation.

In all the experiments the predictive performance improves with the increasing training size and the differences between the methods diminish. CLVAR outperforms all the other methods in the experiments with sparse structures as per our assumptions (mostly markedly). But CLVAR behaves well even in the unfavourable conditions of the first two systems. It still performs better than the other two sparse methods VARL1 and VARLG and the non-sparse

---

[6]Numerical results behind the plots are listed in the Appendix.

VARL2 in the 1st completely sparse experiment[7], and it is on par with the other methods in the 2nd full VAR experiment.

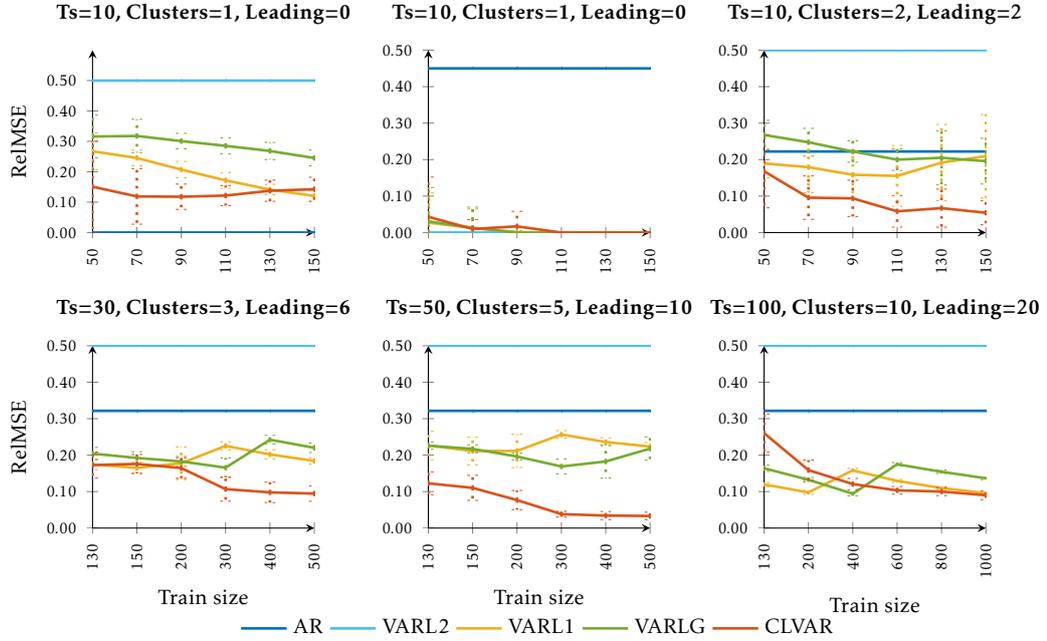

Figure 2.5.2: Synthetic experiments: Average selection error (average of false negative rate and false postive rate) across the 20 experimental replications, the error bars are at 1 standard deviation.

In figure 2.5.2 we show the accuracy of the methods in selecting the true generative G-causal links between the series in the system. The selection error (the lower the better) is measured as the average of the false negative and false positive rates. We plot the averages with ±1 standard deviation over the 20 experimental replications. The CLVAR typically learned models structurally closer to the true generating process than the other tested methods, in most cases with substantial advantage.

To better understand the behaviour of the methods in terms of the structure they learn, we chart in figure 2.5.3 a synthesis of the model matrices **W** learned by the sparse learning methods for the largest training size in the systems with $K = 30$ and $K = 50$ series[8]. The displayed structures correspond to the schema of the **W** matrix presented in figure 2.2.1a. For the figure, the matrices were binarised to simply indicate the existence (1) or non-existence (0) of a G-causal link. The white-to-black shading reflects the number of experimental replications in which this binary indicator is active (equal to 1). So, a black element in the matrix means that this G-causal link was learned in all the 20 resamples of the generating process. White means no G-causality in any of the resamples. Though none of the sparse method was able to clearly and systematically recover the true structures, VARL1 and VARLG clearly suffer from more numerous and more frequent over-selections than CLVAR which matches the true structure more closely and with higher selection stability (fewer light-shaded elements).

---

[7]The AR model is in an advantage here since it has the true-process structure by construction.
[8]Results for the other experiments are deferred to the Supplement.

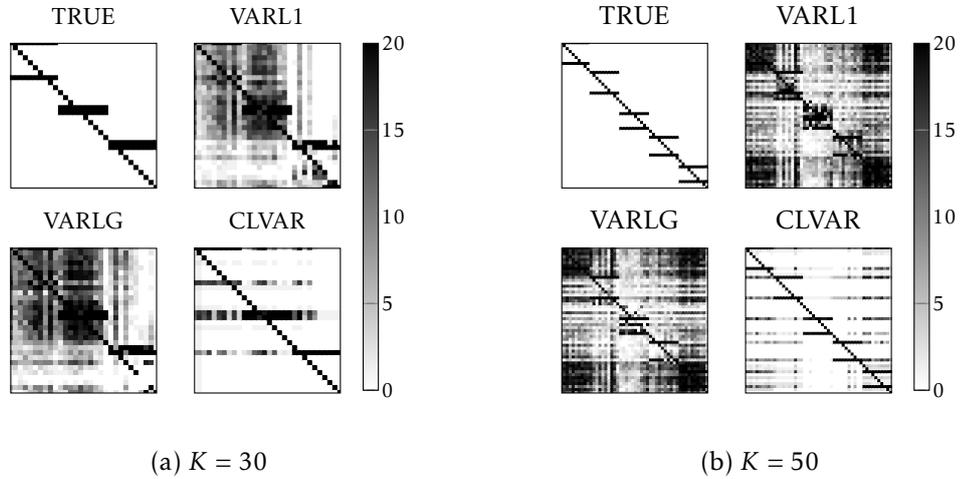

(a) K = 30              (b) K = 50

Figure 2.5.3: Synthetic experiments: Summary of the learned model parameters for the systems with 30 and 50 time series. The shading corresponds to the number of times the element was nonzero in the 20 replications of the experiments with the largest training size.

Finally, we explored how the CLVAR scales with increasing sample size $T$ and the number of time series $K$. The empirical results correspond to the complexity analysis of section 2.3.2: the run-times increased fairly slowly with increasing sample size $T$ but were much longer for systems with higher number of series $K$. Further details are deferred to the Supplement. Overall, the synthetic experiments confirm the desired properties of CLVAR in terms of improved predictive accuracy and structural recovery.

### 2.5.2 Real-data Experiments

We used two real datasets very different in nature, frequency and length of available observations. First, an USGS dataset of daily averages of water physical discharge[9] measured at 17 sites along the Yellowstone (8 sites) and Connecticut (9 sites) river streams (source: Water Services of the US geological survey http://www.usgs.gov/). Second, an economic dataset of quarterly data on 20 major US macro-economic indicators of Stock and Watson (2012) frequently used as a benchmark dataset for VAR learning methods. More details on the datasets can be found in the Supplement.

We preprocessed the data by standard stationary transformations: we followed Stock and Watson (2012) for the economic dataset; by year-on-year log-differences for the USGS. For the short economic dataset, we fixed the hold-out length to 30 and the training sizes from 50 to 130. For the much longer USGS dataset, the hold-out is 300 and the training size increases from 200 to 600. The resamples are constructed by dropping the latest observation from the data and constructing the shifted train and hold-out from this curtailed dataset.

The results of the two sets of experiments are presented in figure 2.5.4. The true parameters of the generative processes are unknown here. Therefore the predictive accuracy is measured in terms of the MSE relative to a random walk model (the lower the better), and the structural recovery is measured in terms of the proportion of active edges in the G-causal graph (the lower the better), always averaged across the 20 resamples with ±1 standard deviation errorbar.

Similarly as in the synthetic experiments, the predictive performance improves with increasing training size and the differences between the methods get smaller. In both experiments, the non-sparse VARL2 has the worst forecasting accuracy (which corresponds to the initial motivation that real large time-series systems tend to be sparse). CLVAR outperformed the other two sparse learning methods VARL1 and VARLG in predictive accuracy as well as sparsity of the learned G-causal graphs. In the economic experiment, the completely (by construction) sparse AR achieved similar predictive accuracy. CLVAR clearly outperforms all the other methods on the USGS dataset.

---

[9]USGS parameter code 00060 - physical discharge in cubic feet per second.

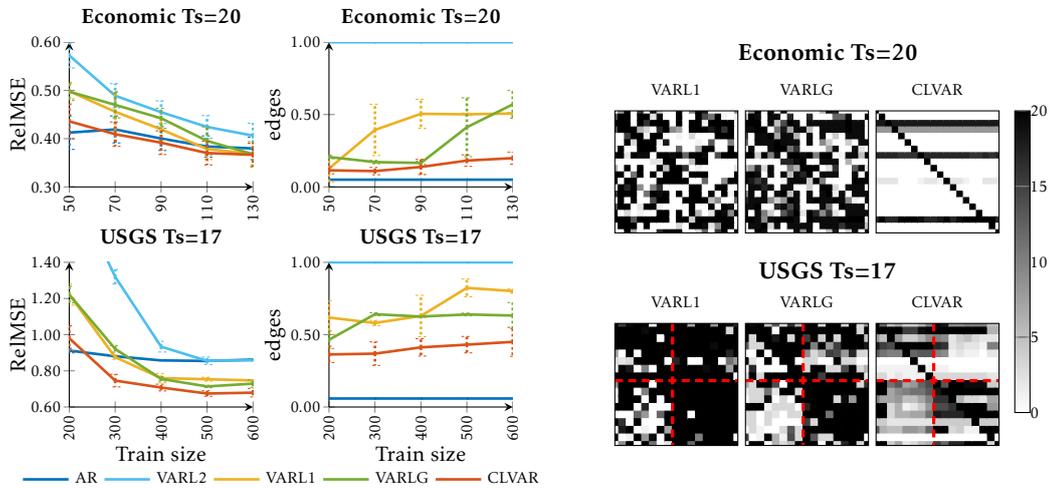

Figure 2.5.4: Real-data experiments: Prediction error in terms of relative MSE as compared to a random walk model, number of edges in the model Granger causality graph, and summary of model parameters. Results are averages, standard deviations and sums across 20 replications of the experiments. The red dashed lines in the USGS parameter matrix separate Yellowstone from Connecticut river sites.

Figure 2.5.4 in the right explores the effect of the structural assumptions on the final shape of the model parameter matrices **W** in the same manner as in figure 2.5.3. The CLVAR matrices are much sparser than the VARL1 and VARLG matrices, organised around a small number of leading indicators. In the economic dataset, the CLVAR method identified three leading indicators for the whole system. In the USGS dataset, the dashed red lines delimit the the Yellowstone (top-left) from the Connecticut (bottom-right) sites. In both these sets of experiments the recovered structure helped improving the forecasts beyond the accuracy achievable by the other tested learning methods.

## 2.6 Conclusions

We presented here a new method for learning sparse VAR models with shared structures in their Granger causality graphs based on the leading indicators of the system, a problem that had not been previously addressed in the time series literature.

The new method has multiple learning objectives: good forecasting perfor-mance of the models, and the discovery of the leading indicators and the clusters of series around them. Meeting these simultaneously is not trivial and we used the techniques of multi-task and structured sparsity learning to achieve it. The method promotes shared patterns in the structure of the individual predictive tasks by forcing them onto a lower-dimensional sub-space spanned by sparse prototypes of the cluster centres. The empirical evaluation confirmed the efficacy of our approach through favourable results of our new method as compared to the state-of-the-art.

# Appendix

## 2.A    Experimental data and transformations

Table 2.A.1 lists the measurement sites of the Water Service of the US Geological Survey (http://www.usgs.gov/) whose data we use in the USGS experiments in section 5.2 of the main text. The original data are the daily averages of the physical discharge in cubic feet per second (parameter code 00060) downloaded from the USGS database on 9/9/2016. We have used the data up to 31/12/2014 and before modelling transformed them by taking the year-on-year log-differences.

Table 2.A.1: Measurement sites for the river-flow data

| Code | Description |
| --- | --- |
| 06191500 | Yellowstone River at Corwin Springs MT |
| 06192500 | Yellowstone River near Livingston MT |
| 06214500 | Yellowstone River at Billings MT |
| 06295000 | Yellowstone River at Forsyth MT |
| 06309000 | Yellowstone River at Miles City MT |
| 06327500 | Yellowstone River at Glendive MT |
| 06329500 | Yellowstone River near Sidney MT |
| 01129200 | CONNECTICUT R BELOW INDIAN STREAM NR PITTSBURG, NH |
| 01129500 | CONNECTICUT RIVER AT NORTH STRATFORD, NH |
| 01131500 | CONNECTICUT RIVER NEAR DALTON, NH |
| 01138500 | CONNECTICUT RIVER AT WELLS RIVER, VT |
| 01144500 | CONNECTICUT RIVER AT WEST LEBANON, NH |
| 01154500 | CONNECTICUT RIVER AT NORTH WALPOLE, NH |
| 01170500 | CONNECTICUT RIVER AT MONTAGUE CITY, MA |
| 01184000 | CONNECTICUT RIVER AT THOMPSONVILLE, CT |

Table 2.A.3 lists the macro-economic indicators of Stock and Watson (2012) used in our economic experiment in section 5.2 in the main text. Before using for modelling we have applied the same pre-processing steps as in Stock and Watson (2012).

We have

- transformed the monthly data to quarterly by taking the quarterly averages (column Q in table 2.A.3);

- applied the stationarizing transformations described in table 2.A.2 (column T in table 2.A.3);
- cleaned the data from outliers by replacing observations with absolute deviations from median larger than 6 times the interquartile range by the median of the 5 preceding values.

Table 2.A.2: Stationarizing transformations

| T | Transformation |
|---|---|
| 1 | $y_t = z_t$ |
| 2 | $y_t = z_t - z_{t-1}$ |
| 3 | $y_t = (z_t - z_{t-1}) - (z_{t-1} - z_{t-2})$ |
| 4 | $y_t = \log(z_t)$ |
| 5 | $y_t = \ln(z_t/z_{t-1})$ |
| 6 | $y_t = \ln(z_t/z_{t-1}) - \ln(z_{t-1}/z_{t-2})$ |

$z_t$ is the original data, $y_t$ is the transformed series

Table 2.A.3: Macro-economic data and transformations

| Code | Q | T | Description |
|---|---|---|---|
| GDP251 | Q | 5 | real gross domestic product, quantity index (2000=100) , saar |
| CPIAUCSL | M | 6 | cpi all items (sa) fred |
| FYFF | M | 2 | interest rate: federal funds (effective) (% per annum,nsa) |
| PSCCOMR | M | 5 | real spot mrkt price idx:bls & crb: all commod(1967=100) |
| FMRNBA | M | 3 | depository inst reserves:nonborrowed,adj res req chgs(mil$ ,sa) |
| FMRRA | M | 6 | depository inst reserves:total,adj for reserve req chgs(mil$ ,sa) |
| FM2 | M | 6 | money stock:m2 (bil$,sa) |
| GDP252 | Q | 5 | real personal consumpt expend, quantity idx (2000=100) , saar |
| IPS10 | M | 5 | industrial production index - total index |
| UTL11 | M | 1 | capacity utilization - manufacturing (sic) |
| LHUR | M | 2 | unemployment rate: all workers, 16 years & over (% ,sa) |
| HSFR | M | 4 | housing starts:nonfarm(1947-58),total farm& nonfarm(1959-) |
| PWFSA | M | 6 | producer price index: finished goods (82=100,sa) |
| GDP273 | Q | 6 | personal consumption expenditures, price idx (2000=100) , saar |
| CES275R | M | 5 | real avg hrly earnings, prod wrkrs, nonfarm - goods-producing |
| FM1 | M | 6 | money stock: m1(bil$ ,sa) |
| FSPIN | M | 5 | s& p's common stock price index: industrials (1941-43=10) |
| FYGT10 | M | 2 | interest rate: u.s.treasury const matur,10-yr.(% per ann,nsa) |
| EXRUS | M | 5 | united states,effective exchange rate(merm)(index no.) |
| CES002 | M | 5 | employees, nonfarm - total private |

## 2.B   Experimental results

This section provides further details on experimental results not included in the main text due to space limitation.

### 2.B.1   Synthetic experiments

Fig. 2.B.1 shows the synthesis of the model parameter matrices **W** for the six synthetic experimental designs. The displayed structures correspond to the schema of the **W** matrix presented in figure 2.2.1a of the main text. For the figure, the matrices were binarised to simply indicate the existence (1) or non-existence (0) of a G-causal link. The white-to-black shading reflects the number of experimental replications in which this binary indicator is active (equal to 1). So, a black element in the matrix means that this G-causal link was learned in all the 20 re-samples of the generating process. White means no G-causality in any of the re-samples. Though none of the sparse method was able to clearly and systematically recover the true structures, VARL1 and VARLG clearly suffer from more numerous and more frequent over-selections than CLVAR which matches the true structures more closely and with higher selection stability (fewer light-shaded elements). The 4th and 5th experimental set-ups are included in the main text as figure 2.5.3.

Fig. 2.B.2 summarises the scaling properties of the CLVAR method with increasing increasing sample size $T$ and the number of time series $K$. In each experiment, we selected a single hyper-parameter combination (near the optimal) and measured the time in seconds (on a single Intel(R) Xeon(R) CPU E5-2680 v2 @ 2.80GHz) and the number of iterations needed till the convergence of the objective (with $10^{-5}$ tolerance) for the 20 data re-samples. We used the $\ell_2$ regularised solution as a warm start. The empirical results correspond to the theoretical complexity analysis of section 3.2 in the main text. For an experimental set-up with fixed number of series $K$ (and G-causal structure), the run-time typically grows fairly slowly with the sample sizes $T$. However, the increases are much more important when moving to larger experiments, with higher $K$ and more complicated structures. Here the growth in run-time is accompanied by higher number of iterations. From our experimental set-

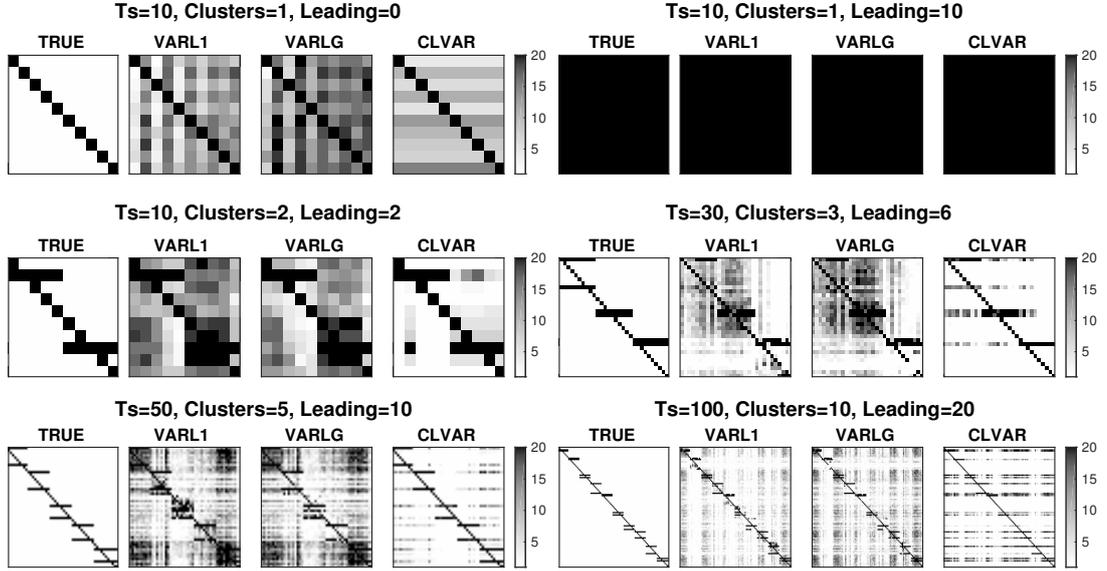

Figure 2.B.1: Synthetic experiments: Summary of the learned model parameters **W**. The shading corresponds to the number of times the element was nonzero in the 20 replications of the experiments with the largest training size.

up it is difficult to separate the effect of enlarging the time-series systems in terms of higher $K$ from the effect of more complicated structures in terms of higher number of clusters and leading indicators. In reality, we expect these to go hand-in-hand so in this sense our empirical analysis complements the theoretical asymptotic complexity analysis of section 3.2 of the main text

Table 2.B.1 provides the numerical data behind the plots of figure 2.5.1 in the main text. The predictive accuracy is measured by MSE of 1-step-ahead forecasts relative to the forecasts produced by the VAR with the true generative coefficients (the irreducible error). The relative MSE is averaged over the 500 hold-out points (the models are fixed and the forecasts are produced by sliding forward over the dataset). The table shows the averages and the standard deviations (in brackets) calculated over the 20 re-samples of the data for each experimental design.

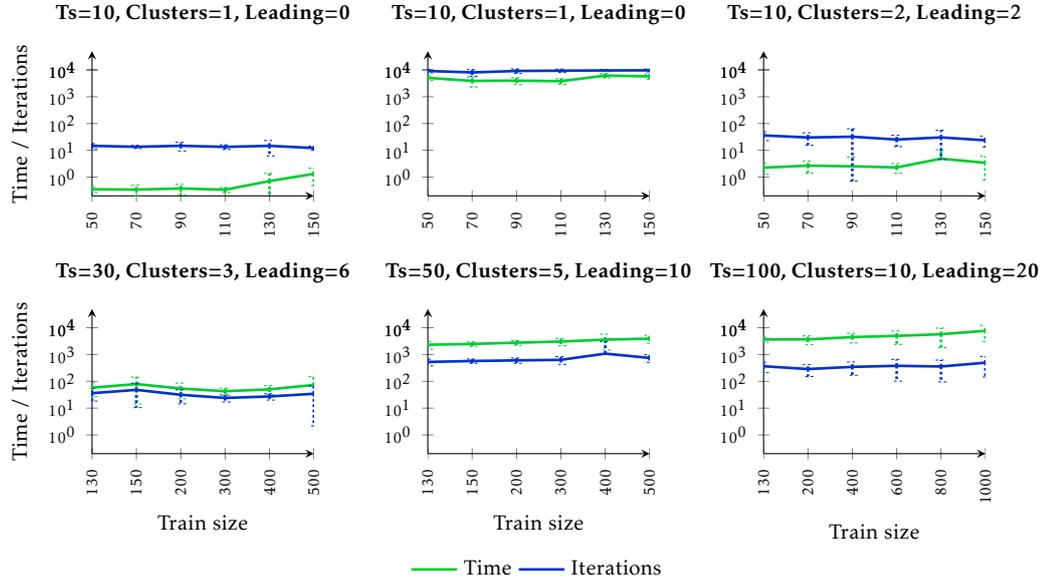

Figure 2.B.2: Synthetic experiments: Average runtime and number of iterations across the 20 experimental replications, the error bars are at 1 standard deviation.

Table 2.B.1: Synthetic experiments: Relative MSE over true model

| Size | AR | VARL2 | VARL1 | VARLG | CLVAR |
|------|-----|-------|-------|-------|-------|
| | | | Ts=10, Clusters=1, Leading=0 | | |
| 50 | 1.225 (0.102) | 3.113 (0.637) | 1.974 (0.310) | 2.513 (0.515) | 1.772 (0.643) |
| 70 | 1.143 (0.063) | 2.570 (0.474) | 1.503 (0.162) | 1.868 (0.304) | 1.395 (0.328) |
| 90 | 1.098 (0.042) | 2.186 (0.353) | 1.324 (0.125) | 1.545 (0.203) | 1.218 (0.082) |
| 110 | 1.071 (0.030) | 1.926 (0.268) | 1.232 (0.086) | 1.368 (0.133) | 1.160 (0.055) |
| 130 | 1.052 (0.020) | 1.745 (0.230) | 1.172 (0.062) | 1.267 (0.095) | 1.121 (0.038) |
| 150 | 1.043 (0.017) | 1.617 (0.178) | 1.143 (0.050) | 1.213 (0.076) | 1.102 (0.034) |
| | | | Ts=10, Clusters=1, Leading=10 | | |
| 50 | 12.563 (5.258) | 23.449 (18.273) | 42.347 (32.893) | 42.055 (32.115) | 22.319 (15.458) |
| 70 | 12.649 (4.507) | 12.712 (4.565) | 9.610 (3.681) | 9.634 (3.472) | 11.692 (4.769) |
| 90 | 12.281 (4.697) | 4.905 (1.545) | 4.943 (1.513) | 4.897 (1.547) | 4.934 (1.584) |
| 110 | 12.583 (4.316) | 3.569 (0.940) | 3.565 (0.939) | 3.562 (0.940) | 3.509 (0.917) |
| 130 | 12.028 (3.762) | 2.527 (0.480) | 2.523 (0.479) | 2.519 (0.477) | 2.477 (0.486) |
| 150 | 11.637 (3.157) | 2.195 (0.417) | 2.194 (0.416) | 2.189 (0.414) | 2.158 (0.432) |
| | | | Ts=10, Clusters=2, Leading=2 | | |
| 50 | 1.428 (0.167) | 2.730 (0.565) | 1.743 (0.246) | 2.124 (0.357) | 1.674 (0.308) |

<div align="right">continues in next page ...</div>

... continues from previous page

| Size | AR | VARL2 | VARL1 | VARLG | CLVAR |
|------|-----|-------|-------|-------|-------|
| 70 | 1.336 (0.096) | 2.296 (0.346) | 1.522 (0.206) | 1.689 (0.199) | 1.405 (0.139) |
| 90 | 1.295 (0.100) | 1.998 (0.253) | 1.398 (0.139) | 1.499 (0.146) | 1.289 (0.116) |
| 110 | 1.258 (0.063) | 1.824 (0.214) | 1.320 (0.082) | 1.396 (0.103) | 1.199 (0.062) |
| 130 | 1.237 (0.057) | 1.686 (0.181) | 1.283 (0.078) | 1.347 (0.084) | 1.156 (0.050) |
| 150 | 1.229 (0.054) | 1.566 (0.171) | 1.255 (0.076) | 1.300 (0.081) | 1.133 (0.043) |
| Ts=30, Clusters=3, Leading=6 | | | | | |
| 130 | 1.532 (0.142) | 2.424 (0.361) | 1.355 (0.090) | 1.490 (0.131) | 1.261 (0.081) |
| 150 | 1.512 (0.129) | 2.245 (0.308) | 1.319 (0.081) | 1.414 (0.111) | 1.225 (0.062) |
| 200 | 1.471 (0.113) | 1.861 (0.205) | 1.257 (0.062) | 1.289 (0.069) | 1.172 (0.048) |
| 300 | 1.451 (0.103) | 1.586 (0.135) | 1.164 (0.039) | 1.224 (0.051) | 1.100 (0.031) |
| 400 | 1.438 (0.100) | 1.375 (0.086) | 1.119 (0.028) | 1.151 (0.036) | 1.073 (0.024) |
| 500 | 1.434 (0.099) | 1.298 (0.068) | 1.097 (0.023) | 1.116 (0.027) | 1.059 (0.018) |
| Ts=50, Clusters=5, Leading=10 | | | | | |
| 130 | 3.327 (0.557) | 4.376 (0.792) | 2.024 (0.306) | 2.160 (0.277) | 1.746 (0.223) |
| 150 | 3.254 (0.544) | 3.920 (0.685) | 1.801 (0.205) | 1.915 (0.215) | 1.577 (0.196) |
| 200 | 3.200 (0.506) | 3.203 (0.522) | 1.583 (0.146) | 1.620 (0.145) | 1.324 (0.109) |
| 300 | 3.161 (0.491) | 2.280 (0.301) | 1.347 (0.082) | 1.392 (0.091) | 1.128 (0.034) |
| 400 | 3.123 (0.478) | 1.844 (0.192) | 1.228 (0.052) | 1.299 (0.070) | 1.091 (0.025) |
| 500 | 3.097 (0.468) | 1.621 (0.144) | 1.172 (0.039) | 1.214 (0.050) | 1.081 (0.025) |
| Ts=100, Clusters=10, Leading=20 | | | | | |
| 130 | 1.890 (0.202) | 3.362 (0.536) | 1.518 (0.118) | 1.807 (0.184) | 1.734 (0.178) |
| 200 | 1.845 (0.189) | 2.855 (0.420) | 1.350 (0.078) | 1.478 (0.106) | 1.415 (0.109) |
| 400 | 1.801 (0.177) | 2.196 (0.264) | 1.225 (0.050) | 1.268 (0.059) | 1.180 (0.049) |
| 600 | 1.783 (0.172) | 1.820 (0.181) | 1.137 (0.031) | 1.191 (0.043) | 1.108 (0.026) |
| 800 | 1.777 (0.170) | 1.643 (0.142) | 1.104 (0.023) | 1.129 (0.029) | 1.084 (0.021) |
| 1000 | 1.774 (0.170) | 1.501 (0.115) | 1.090 (0.020) | 1.100 (0.022) | 1.065 (0.019) |

Table 2.B.2 provides the numerical data behind the plots of figure 2.5.2 in the main text. The selection accuracy of the true G-causal links is measured by the average between the false negative and false positive rates. The table shows the averages and the standard deviations (in brackets) calculated over the 20 re-samples of the data for each experimental design.

Table 2.B.2: Synthetic experiments: Selection errors of true G-causal links

| Size | AR | VARL2 | VARL1 | VARLG | CLVAR |
|------|-----|--------|--------|--------|--------|
| | | Ts=10, Clusters=1, Leading=0 | | | |
| 50 | 0.000 (0.000) | 0.500 (0.000) | 0.268 (0.060) | 0.316 (0.071) | 0.151 (0.149) |
| 70 | 0.000 (0.000) | 0.500 (0.000) | 0.245 (0.026) | 0.318 (0.055) | 0.119 (0.091) |
| 90 | 0.000 (0.000) | 0.500 (0.000) | 0.207 (0.027) | 0.301 (0.026) | 0.118 (0.042) |
| 110 | 0.000 (0.000) | 0.500 (0.000) | 0.172 (0.026) | 0.285 (0.026) | 0.122 (0.033) |
| 130 | 0.000 (0.000) | 0.500 (0.000) | 0.141 (0.017) | 0.268 (0.028) | 0.138 (0.035) |
| 150 | 0.000 (0.000) | 0.500 (0.000) | 0.120 (0.018) | 0.245 (0.026) | 0.143 (0.040) |
| | | Ts=10, Clusters=1, Leading=10 | | | |
| 50 | 0.450 (0.000) | 0.000 (0.000) | 0.027 (0.082) | 0.031 (0.093) | 0.043 (0.109) |
| 70 | 0.450 (0.000) | 0.000 (0.000) | 0.013 (0.052) | 0.013 (0.057) | 0.010 (0.026) |
| 90 | 0.450 (0.000) | 0.000 (0.000) | 0.001 (0.004) | 0.000 (0.000) | 0.017 (0.041) |
| 110 | 0.450 (0.000) | 0.000 (0.000) | 0.000 (0.000) | 0.000 (0.000) | 0.000 (0.000) |
| 130 | 0.450 (0.000) | 0.000 (0.000) | 0.000 (0.000) | 0.000 (0.000) | 0.000 (0.000) |
| 150 | 0.450 (0.000) | 0.000 (0.000) | 0.000 (0.000) | 0.000 (0.000) | 0.000 (0.000) |
| | | Ts=10, Clusters=2, Leading=2 | | | |
| 50 | 0.222 (0.000) | 0.500 (0.000) | 0.190 (0.040) | 0.268 (0.040) | 0.168 (0.100) |
| 70 | 0.222 (0.000) | 0.500 (0.000) | 0.179 (0.070) | 0.248 (0.038) | 0.096 (0.060) |
| 90 | 0.222 (0.000) | 0.500 (0.000) | 0.159 (0.066) | 0.222 (0.029) | 0.094 (0.050) |
| 110 | 0.222 (0.000) | 0.500 (0.000) | 0.155 (0.083) | 0.200 (0.030) | 0.058 (0.043) |
| 130 | 0.222 (0.000) | 0.500 (0.000) | 0.192 (0.104) | 0.205 (0.074) | 0.067 (0.052) |
| 150 | 0.222 (0.000) | 0.500 (0.000) | 0.209 (0.114) | 0.196 (0.063) | 0.054 (0.033) |
| | | Ts=30, Clusters=3, Leading=6 | | | |
| 130 | 0.321 (0.000) | 0.500 (0.000) | 0.175 (0.015) | 0.204 (0.017) | 0.173 (0.036) |
| 150 | 0.321 (0.000) | 0.500 (0.000) | 0.165 (0.015) | 0.192 (0.017) | 0.176 (0.026) |
| 200 | 0.321 (0.000) | 0.500 (0.000) | 0.178 (0.045) | 0.183 (0.014) | 0.164 (0.028) |
| 300 | 0.321 (0.000) | 0.500 (0.000) | 0.225 (0.011) | 0.166 (0.027) | 0.107 (0.033) |
| 400 | 0.321 (0.000) | 0.500 (0.000) | 0.202 (0.013) | 0.242 (0.012) | 0.098 (0.028) |
| 500 | 0.321 (0.000) | 0.500 (0.000) | 0.184 (0.009) | 0.220 (0.013) | 0.095 (0.021) |
| | | Ts=50, Clusters=5, Leading=10 | | | |
| 130 | 0.321 (0.000) | 0.500 (0.000) | 0.226 (0.039) | 0.226 (0.010) | 0.123 (0.031) |
| 150 | 0.321 (0.000) | 0.500 (0.000) | 0.211 (0.038) | 0.217 (0.011) | 0.110 (0.034) |
| 200 | 0.321 (0.000) | 0.500 (0.000) | 0.212 (0.046) | 0.196 (0.009) | 0.077 (0.026) |
| 300 | 0.321 (0.000) | 0.500 (0.000) | 0.257 (0.011) | 0.169 (0.020) | 0.038 (0.008) |
| 400 | 0.321 (0.000) | 0.500 (0.000) | 0.236 (0.012) | 0.183 (0.046) | 0.034 (0.011) |
| 500 | 0.321 (0.000) | 0.500 (0.000) | 0.224 (0.011) | 0.218 (0.032) | 0.033 (0.011) |

| Size | AR | VARL2 | VARL1 | VARLG | CLVAR |
|------|-----|-------|-------|-------|-------|
| | | Ts=100, Clusters=10, Leading=20 | | | |
| 130 | 0.321 (0.000) | 0.500 (0.000) | 0.120 (0.006) | 0.164 (0.009) | 0.260 (0.052) |
| 200 | 0.321 (0.000) | 0.500 (0.000) | 0.098 (0.005) | 0.132 (0.006) | 0.159 (0.027) |
| 400 | 0.321 (0.000) | 0.500 (0.000) | 0.158 (0.005) | 0.094 (0.006) | 0.121 (0.015) |
| 600 | 0.321 (0.000) | 0.500 (0.000) | 0.129 (0.004) | 0.175 (0.005) | 0.103 (0.010) |
| 800 | 0.321 (0.000) | 0.500 (0.000) | 0.110 (0.004) | 0.154 (0.005) | 0.100 (0.011) |
| 1000 | 0.321 (0.000) | 0.500 (0.000) | 0.096 (0.004) | 0.137 (0.004) | 0.090 (0.012) |

### 2.B.2 Real-data experiments

Tables 2.B.3 and 2.B.4 provide the numerical data behind the plots of figure 2.5.4 in the main text. The predictive accuracy is measured by MSE of 1-step-ahead forecasts relative to the forecasts produced random walk model (= uses the last observed value as the 1-step-ahead forecast). The relative MSE is averaged over the 30 and 300 hold-out points for the Economic and the USGS dataset respectively (the models are fixed and the forecasts are produced by sliding forward over the dataset). The sparsity of the learned models is measured by the proportion of active edges in the learned G-causality graph. The table shows the averages and the standard deviations (in brackets) calculated over the 20 re-samples of the data for each experimental design.

Table 2.B.3: Real-data experiments: Relative MSE over random walk

| Size | AR | VARL2 | VARL1 | VARLG | CLVAR |
|------|-----|-------|-------|-------|-------|
| | | Economic Ts=20 | | | |
| 50 | 0.413 (0.035) | 0.573 (0.026) | 0.498 (0.016) | 0.498 (0.018) | 0.436 (0.036) |
| 70 | 0.419 (0.028) | 0.489 (0.025) | 0.456 (0.041) | 0.470 (0.027) | 0.409 (0.025) |
| 90 | 0.400 (0.025) | 0.455 (0.024) | 0.419 (0.023) | 0.442 (0.021) | 0.392 (0.025) |
| 110 | 0.384 (0.022) | 0.424 (0.023) | 0.379 (0.022) | 0.395 (0.029) | 0.370 (0.025) |
| 130 | 0.380 (0.023) | 0.406 (0.026) | 0.368 (0.022) | 0.368 (0.026) | 0.367 (0.024) |
| | | USGS Ts=17 | | | |

| Size | AR | VARL2 | VARL1 | VARLG | CLVAR |
|------|-----|-------|-------|-------|-------|
| 200 | 0.912 (0.013) | 1.857 (0.098) | 1.222 (0.058) | 1.222 (0.046) | 0.980 (0.069) |
| 300 | 0.881 (0.003) | 1.320 (0.040) | 0.876 (0.012) | 0.921 (0.019) | 0.746 (0.035) |
| 400 | 0.858 (0.001) | 0.934 (0.029) | 0.760 (0.026) | 0.755 (0.012) | 0.708 (0.019) |
| 500 | 0.855 (0.002) | 0.855 (0.021) | 0.754 (0.011) | 0.714 (0.007) | 0.675 (0.015) |
| 600 | 0.862 (0.002) | 0.858 (0.004) | 0.747 (0.004) | 0.729 (0.020) | 0.679 (0.024) |

Table 2.B.4: Real-data experiments: proportion of G-causal graph edges

| Size | AR | VARL2 | VARL1 | VARLG | CLVAR |
|------|-----|-------|-------|-------|-------|
| Economic Ts=20 | | | | | |
| 50 | 0.050 (0.000) | 1.000 (0.000) | 0.127 (0.092) | 0.207 (0.012) | 0.115 (0.024) |
| 70 | 0.050 (0.000) | 1.000 (0.000) | 0.394 (0.177) | 0.172 (0.012) | 0.109 (0.025) |
| 90 | 0.050 (0.000) | 1.000 (0.000) | 0.505 (0.100) | 0.166 (0.011) | 0.138 (0.053) |
| 110 | 0.050 (0.000) | 1.000 (0.000) | 0.502 (0.009) | 0.412 (0.204) | 0.182 (0.040) |
| 130 | 0.050 (0.000) | 1.000 (0.000) | 0.507 (0.007) | 0.570 (0.096) | 0.199 (0.042) |
| USGS Ts=17 | | | | | |
| 200 | 0.059 (0.000) | 1.000 (0.000) | 0.618 (0.117) | 0.465 (0.061) | 0.363 (0.056) |
| 300 | 0.059 (0.000) | 1.000 (0.000) | 0.581 (0.018) | 0.641 (0.010) | 0.369 (0.081) |
| 400 | 0.059 (0.000) | 1.000 (0.000) | 0.629 (0.143) | 0.625 (0.005) | 0.412 (0.060) |
| 500 | 0.059 (0.000) | 1.000 (0.000) | 0.823 (0.061) | 0.640 (0.009) | 0.431 (0.057) |
| 600 | 0.059 (0.000) | 1.000 (0.000) | 0.801 (0.014) | 0.632 (0.090) | 0.450 (0.100) |

# Chapter 3

# Forecasting and Granger Modelling with Non-linear Dynamical Dependencies

**Chapter abstract:** Traditional linear methods for forecasting multivariate time series are not able to satisfactorily model the non-linear dependencies that may exist in non-Gaussian series. We build on the theory of learning vector-valued functions in the reproducing kernel Hilbert space and develop a method for learning prediction functions that accommodate such non-linearities. The method not only learns the predictive function but also the matrix-valued kernel underlying the function search space directly from the data. Our approach is based on learning multiple matrix-valued kernels, each of those composed of a set of input kernels and a set of output kernels learned in the cone of positive semi-definite matrices. In addition to superior predictive performance in the presence of strong non-linearities, our method also recovers the hidden dynamic relationships between the series and thus is a new alternative to existing graphical Granger techniques.

## 3.1   Introduction

Traditional methods for forecasting stationary multivariate time series from their own past are derived from the classical linear vector autoregressive models (VARs). In these, the prediction of the next point in the future of the series is constructed as a linear function of the past observations. The use of linear functions as the predictors is in part based on the Wold representation theorem (e.g. Brockwell and Davis (1991)) and in part, probably more importantly, on the fact that the linear predictor is the best predictor (in the mean-square-error sense) in case the time series is Gaussian.

The Gaussian assumption is therefore often adopted in the analysis of time series to justify the simple linear modelling. However, it is indeed a simplifying assumption since for non-Gaussian series the best predictor may very well be a non-linear function of the past observations. A number of parametric non-linear models has been proposed in the literature, each adapted to capture specific sources of non-linearity (for example multiple forms of regime-switching models, e.g. Turkman et al. (2014)).

In our work we adopt an approach that does not rely on such prior assumptions for the function form. We propose to learn the predictor as a general vector-valued function $\mathbf{f}$ that takes as input the past observations of the multivariate series and outputs the forecast of the unknown next value (vector).

We have two principal requirements on the function $\mathbf{f}$. The first is the standard prediction accuracy requirement. That is, the function $\mathbf{f}$ shall be such that we can expect its outputs to be close (in the squared error sense) to the true future observations of the process. The second requirement is that the function $\mathbf{f}$ shall have a structure that will enable the analysis of the relationships amongst the subprocesses of the multivariate series. Namely, we wish to understand how parts of the series help in forecasting other parts of the multivariate series, a concept known in the time-series literature as graphical Granger modelling (Granger 1969; Eichler 2012).

To learn such a function $\mathbf{f}$ we employ the framework of regularised learning of vector-valued functions in the RKHS (Micchelli and Pontil 2005). Learning methods based on the RKHS theory have previously been considered for time series modelling (e.g. Franz and Schölkopf 2006; Sindhwani et al. 2013; Lim et al. 2014). Though, as Pillonetto et al. (2014) note in their survey, their adoption for the dynamical system analysis is not a commonplace.

A critical step in kernel-based methods for learning vector-valued functions is the specification of the operator-valued kernel that exploits well the relationships between the inputs and the outputs. A convenient and well-studied class of operator-valued kernels (e.g. in Caponnetto et al. 2008; Dinuzzo and Fukumizu 2011; Jawanpuria et al. 2015) are those decomposable into a product of a scalar kernel on the input space (input kernel) and a linear operator on the output space (output kernel).

The kernel uniquely determines the function space within which the function **f** is learned. It thus has significant influence on both our objectives described above. Instead of having to choose the input and the output kernels a priori, we introduce a method for learning the input and output kernels from the data together with learning the vector-valued function **f**.

Our method combines in a novel way the multiple kernel learning (MKL) approach (Lanckriet et al. 2004) with learning the output kernels within the space of positive semidefinite linear operators on the output space (Jawanpuria et al. 2015). MKL methods for operator-valued kernels have recently been developed in Kadri et al. (2012) and Sindhwani et al. (2013). The first learns a convex combination of a set of operator-valued kernels fixed in advance, the second combines a fixed set of input kernels with a single learned output kernel. To the best of our knowledge, ours is the first method in which the operator-valued kernel is learned by combining a set of input kernels with a set of multiple learned output kernels.

In accordance with our second objective stated above, we impose specific structural constraints on the function search space so that the learned function supports the graphical Granger analysis. We achieve this by working with matrix-valued kernels operating over input partitions restricted to single input scalar series (similar input partitioning has recently been used in Sindhwani et al. (2013)).

We impose diagonal structure on the output kernels to control the model complexity. Though this has a cost in the inability to model contemporaneous relationships, it addresses the strong over-parametrisation in a principled manner.

It also greatly simplifies the final structure of the problem, which, in result, suitably decomposes into a set of smaller independent problems solvable in parallel.

We develop two forms of sparsity-promoting regularisation approaches for learning the output kernels. These are based on the $\ell_1$ and $\ell_1/\ell_2$ norms respectively and are motivated by the search for Granger-causality relationships. As to our knowledge, the latter has not been previously used in the context of MKL.

Finally, we confirm on experiments the benefits our methods can bring to forecasting non-Gaussian series in terms of improved predictive accuracy and the ability to recover hidden dynamic dependency structure within the time series systems. This makes them valid alternatives to the state-of-the-art graphical Granger techniques.

## 3.2   Problem formulation

Given a realisation of a discrete stationary multivariate time series process $\left\{ \mathbf{y}_t \in \mathcal{Y} \subseteq \mathbb{R}^m : t \in \mathbb{N}_n \right\}$, our goal is to learn a vector-valued function $\mathbf{f} : \mathcal{Y}^p \to \mathcal{Y}$ that takes as input the $p$ past observations of the process and predicts its future vector value (one step ahead). The function $\mathbf{f}$ shall be such that

1. we can expect the prediction to be near (in the Euclidean distance sense) the unobserved future value
2. its structure allows to analyse if parts (subprocesses) of the series are useful for forecasting other subprocesses within the series or if some subprocesses can be forecast independently of the rest; in short, it allows Granger-causality analysis (Granger 1969).

For notational simplicity, from now on we indicate the output of the function $\mathbf{f}$ as $\mathbf{y} \in \mathcal{Y} \subseteq \mathbb{R}^m$ and the input as $\mathbf{x} \in \mathcal{X} \subseteq \mathbb{R}^{mp}$ (bearing in mind that $\mathcal{X} = \mathcal{Y}^p$ is in fact the $p$-th order Cartesian product of $\mathcal{Y}$ and that the inputs $\mathbf{x}$ and outputs $\mathbf{y}$ are the past and future observations of the same $m$-dimensional series). We also align the time indexes so that our data sample consists of input-output data pairs $\left\{ (\mathbf{y}_t, \mathbf{x}_t) : t \in \mathbb{N}_n \right\}$ (see also figure 1.2 in chapter 1).

Following the standard function learning theory, we will learn $\mathbf{f} \in \mathcal{H}_K$ by minimising the regularised empirical squared-error risk (with a regularization parameter $\lambda > 0$)

$$\widehat{\mathbf{f}} = \arg\min_{\mathbf{f} \in \mathcal{H}_K} R(\mathbf{f})$$
$$R(\mathbf{f}) := \sum_{t=1}^{T} \|\mathbf{y}_t - \mathbf{f}(\mathbf{x}_t)\|_2^2 + \lambda \|\mathbf{f}\|_{\mathcal{H}_K}^2 \quad . \tag{3.1}$$

Here $\mathcal{H}_K$ is the reproducing kernel Hilbert space (RKHS) of $\mathbb{R}^m$-valued functions endowed with the norm $\|.\|_{\mathcal{H}_K}$ and the inner product $\langle .,.\rangle_{\mathcal{H}_K}$. The RKHS is uniquely associated with a symmetric positive-semidefinite matrix-valued kernel $\mathbf{H} : \mathcal{X} \times \mathcal{X} \to \mathbb{R}^{m \times m}$ with the reproducing property

$$\langle \mathbf{y}, \mathbf{g}(\mathbf{x}) \rangle = \langle \mathbf{H_x y}, \mathbf{g} \rangle_{\mathcal{H}_K} \quad \forall (\mathbf{y}, \mathbf{x}, \mathbf{g}) \in (\mathcal{Y}, \mathcal{X}, \mathcal{H}_K) \ ,$$

where $\mathbf{H_x y} : \mathcal{X} \to \mathcal{Y}$ is the function centred at $(\mathbf{x}, \mathbf{y})$ and defined as $\mathbf{H_x y}(\mathbf{x}') := \mathbf{H}(\mathbf{x}, \mathbf{x}')\mathbf{y}$. From the classical result in Micchelli and Pontil (2005), the unique solution $\widehat{\mathbf{f}}$ of the variational problem (3.1) admits a finite dimensional representation

$$\widehat{\mathbf{f}} = \sum_{t=1}^{T} \mathbf{H_{x_t}} \mathbf{c}_t \ , \tag{3.2}$$

where the coefficients $\mathbf{c}_t \in \mathcal{Y}$ are the solutions of the system of linear equations

$$\sum_{t=1}^{T} \Big( \mathbf{H}(\mathbf{x}_s, \mathbf{x}_t) + \lambda \delta_{st} \Big) \mathbf{c}_t = \mathbf{y}_s, \quad \forall s \in \mathbb{N}_n \ , \tag{3.3}$$

where $\delta_{st} = 1$ if $s = t$ and is zero otherwise.

### 3.2.1 Granger-causality analysis

To study the dynamical relationships in time series processes, Granger (1969) proposed a practical definition of causality based on the accuracy of least-squares predictor functions. In brief, for two time series processes $\{\mathbf{y}_t\}$ and $\{\mathbf{z}_t\}$, $\{\mathbf{y}_t\}$ is said to Granger-cause $\{\mathbf{z}_t\}$ if given all the other relevant information we can predict the future of $\{\mathbf{z}_t\}$ better (in the mean-square-error sense) using the history of $\{\mathbf{y}_t\}$ than without it.

Eichler (2012) extended the concept to multivariate analysis through graphical models. The discussion in the paper focuses on the notion of Granger non-causality rather than causality and describes the specific Markov properties (conditional non-causality) encoded in the graphs of Granger-causal relationships. In this sense, the absence of a variable in a set of inputs is more informative of the Granger (non-)causality than its presence. In result, graphical Granger methods are typically based on (structured) sparse modelling (Bahadori and Liu 2013).

## 3.3 Function space and kernel specification

The function space $\mathcal{H}_K$ within which $\mathbf{f}$ is learned is fully determined by the reproducing kernel $\mathbf{H}$. Its specification is therefore critical for achieving the two objectives for the function $\mathbf{f}$ defined in section 3.2. We focus on the class of matrix-valued kernels decomposable into the product of input kernels, capturing the similarities in the inputs, and output kernels, encoding the relationships between the outputs.

To analyse the dynamical dependencies between the series, we need to be able to discern within the inputs of the learned function $\mathbf{f}$ the individual scalar series. Therefore we partition the elements of the input vectors according to the source scalar time series. In result, instead of a single kernel operating over the full vectors, we work with multiple partition-kernels, each of them operating over a single input series. We further propose to learn the partition-kernels by combining the MKL techniques with output kernel learning within the cone of positive semidefinite matrices $\mathbb{S}_+^m$.

More formally, the kernel we propose to use is constructed as a sum of kernels $\mathbf{H} = \sum_j^m \mathbf{H}^{(j)}$, where $m$ is the number of the individual scalar-valued series in the multivariate process (dimensionality of the output space $\mathcal{Y}$). Each $\mathbf{H}^{(j)} : \mathcal{X}^{(j)} \times \mathcal{X}^{(j)} \to \mathbb{R}^{m \times m}$ is a matrix-valued kernel that determines its own RKHS of vector-valued functions. The domains $\mathcal{X}^{(j)} \subseteq \mathbb{R}^p$ are sets of vectors constructed by selecting from the inputs $\mathbf{x}$ only the $p$ coordinates $i^{(j)} \in \mathbb{N}_{mp}$ that correspond to the past of a single scalar time series $j$.

$$\mathcal{X}^{(j)} = \{\mathbf{x}^{(j)} : x_i^{(j)} = x_{i(j)} \,\forall i, \, \mathbf{x} \in \mathcal{X}\}, \quad \cup_j \mathcal{X}^{(j)} = \mathcal{X}$$

Further, instead of choosing the individual matrix-valued functions $\mathbf{H}^{(j)}$, we propose to learn them. We construct each $\mathbf{H}^{(j)}$ again as a sum of kernels $\mathbf{H}^{(j)} = \sum_i^{s_j} \mathbf{H}^{(ji)}$ of possibly uneven number of summands $s_j$ of matrix-valued kernels $\mathbf{H}^{(ji)} : \mathcal{X}^{(j)} \times \mathcal{X}^{(j)} \to \mathbb{R}^{m \times m}$. For this lowest level $\mathbf{H}^{(ji)}$ we focus on the family of decomposable kernels $\mathbf{H}^{(ji)} = k^{(ji)} \mathbf{L}^{(ji)}$. Here, the input kernels $k^{(ji)} : \mathcal{X}^{(j)} \times \mathcal{X}^{(j)} \to \mathbb{R}$ capturing the similarity between the inputs are fixed in advance from a dictionary of valid scalar-valued kernels (e.g. Gaussian kernels with varying scales). The set $\mathcal{L} = \left\{ \mathbf{L}^{(ji)} : j = \mathbb{N}_m, i = \mathbb{N}_{s_j}, \sum_j^m s_j = l \right\}$ of output kernels $\mathbf{L}^{(ji)} : \mathcal{Y} \to \mathcal{Y}$ encoding the relations between the outputs is learned within the cone of symmetric positive semidefinite matrices $\mathbb{S}_+^m$.

$$\mathbf{H} = \sum_{j=1}^m \mathbf{H}^{(j)} = \sum_{j=1}^m \sum_{i=1}^{s_j} \mathbf{H}^{(ji)} = \sum_{j=1}^m \sum_{i=1}^{s_j} k^{(ji)} \mathbf{L}^{(ji)} \tag{3.4}$$

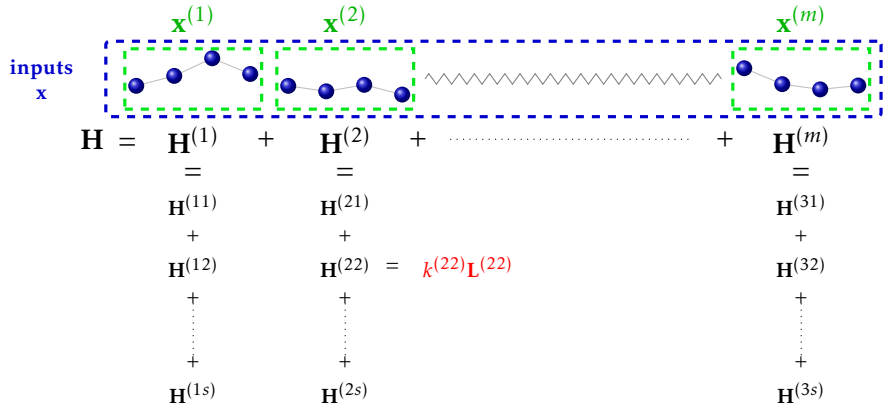

Figure 3.3.1: The matrix-valued kernel $\mathbf{H}$ is constructed from multiple kernels $\mathbf{H}^{(ji)}$ operating over input partitions. Each constituent is decomposable into input and output kernels $\mathbf{H}^{(ji)} = k^{(ji)} \mathbf{L}^{(ji)}$.

### 3.3.1 Kernel learning and function estimation

Learning all the output kernels $\mathbf{L}^{(ji)}$ as full PSD matrices implies learning more than $m^3$ parameters. To improve the generalization capability, we reduce the complexity of the problem drastically by restricting the search space for $\mathbf{L}$'s to PSD diagonal matrices $\mathbb{D}_+^m$. This essentially corresponds to the assumption of no contemporaneous relationships between the series. We return to this point in section 3.5.

As explained in section 3.2.1, Granger (non-)causality learning typically searches for sparse models. We bring this into our methods by imposing a further sparsity inducing regularizer $Q : (\mathbb{R}^{m \times m})^l \to \mathbb{R}$ on the set of the output kernels $\mathcal{L}$. We motivate and elaborate suitable forms of $Q$ in section 3.3.2.

The joint learning of the kernels and the function can now be formulated as the problem of finding the minimising solution $\mathbf{f} \in \mathcal{H}_K$ and $\mathbf{L}$'s $\in \mathbb{D}_+^m$ of the regularised functional

$$J(\mathbf{f}, \mathcal{L}) := R(\mathbf{f}) + \tau Q(\mathcal{L}), \quad \tau > 0 \ , \tag{3.5}$$

where $R(\mathbf{f})$ is the regularised risk from equation (3.1). By calling on the properties of the RKHS, we reformulate this as a finite dimensional problem that can be addressed by conventional finite-dimensional optimisation approaches. We introduce the gram matrices $\mathbf{K}^{(ji)} \in \mathbb{S}_+^n$ such that $K_{ts}^{(ji)} = k^{(ji)}(\mathbf{x}_t^{(j)}, \mathbf{x}_s^{(j)})$ for all $t, s \in \mathbb{N}_n$, the output data matrix $\mathbf{Y} \in \mathbb{R}^{n \times m}$ such that $\mathbf{Y} = (\mathbf{y}_1, \dots \mathbf{y}_n)^T$, and the coefficient matrix $\mathbf{C} \in \mathbb{R}^{n \times m}$ such that $\mathbf{C} = (\mathbf{c}_1, \dots \mathbf{c}_n)^T$.

Using these and the result in equation (3.2) it is easy to show that the minimisation of the regularised risk $R(\mathbf{f})$ in problem (3.1) with respect to $\mathbf{f} \in \mathcal{H}_K$ is equivalent to the minimisation with respect to $\mathbf{C} \in \mathbb{R}^{n \times m}$ of the objective

$$\widetilde{R}(\mathbf{C}) := \left\| \mathbf{Y} - \sum_{ji} \mathbf{K}^{(ji)} \mathbf{C} \mathbf{L}^{(ji)} \right\|_F^2 + \lambda \sum_{ji} \left\langle \mathbf{C}^T \mathbf{K}^{(ji)} \mathbf{C}, \mathbf{L}^{(ji)} \right\rangle_F \ . \tag{3.6}$$

The finite dimensional equivalent of the minimisation problem (3.5) is thus the joint minimisation of

$$\widetilde{J}(\mathbf{C}, \mathcal{L}) := \widetilde{R}(\mathbf{C}, \mathcal{L}) + \tau Q(\mathcal{L}) \ . \tag{3.7}$$

### 3.3.2 Sparse regularization

The construction of the kernel $\mathbf{H}$ and the function space $\mathcal{H}_K$ described in section 3.3 imposes on the function $\mathbf{f}$ the necessary structure that allows the Granger-causality analysis (as per our 2nd objective set-out in section 3.2). As explained in section 3.2.1, the other ingredient we need to identify the Granger non-causalities is sparsity within the structure of the learned function.

In our methods, the sparsity is introduced by the regularizer $Q$. By construction of the function space, we can examine the elements of the output kernels $\mathbf{L}^{(ij)}$ (their diagonals) to make statements about the Granger non-causality. We say the $j$-th scalar time series is non-causal for the $s$ series (given all the remaining series in the process) if $\mathbf{L}^{(ji)}_{ss} = 0$ for all $i \in \mathbb{N}_{s_j}$.

Essentially, any of the numerous regularizers that exist for sparse or structured sparse learning (e.g. Bach et al. (2012)) could be used as $Q$, possibly based on some prior knowledge about the underlying dependencies within the time-series process.

We elaborate here two cases that do not assume any special structure in the dependencies as the base scenarios. The first is the entry-wise $\ell_1$ norm across all the output kernels so that

$$Q_1(\mathcal{L}) = \sum_{ji} \left\| \mathbf{L}^{(ji)} \right\|_1 = \sum_{ji} \sum_{s}^{m} |L^{(ji)}_{ss}| \ . \tag{3.8}$$

The second is the $\ell_1/\ell_2$ grouped norm

$$Q_{1/2}(\mathcal{L}) = \sum_{js} \sqrt{\sum_i \left( L^{(ji)}_{ss} \right)^2} \ . \tag{3.9}$$

After developing the learning strategy for these in sections 3.4.1 and 3.4.2, we provide some more intuition of their effects on the models and link to some other known graphical Granger techniques in section 3.5.

## 3.4 Learning strategy

First of all, we simplify the final formulation of problem (3.7) in section 3.3.1. Rather than working with a set of diagonal matrices $\mathbf{L}^{(ji)}$, we merge the diagonals into a single matrix $\mathbf{A}$. We then re-formulate the problem with respect to this single matrix in place of the set and show how this reformulation can be suitably decomposed into smaller independent sub-problems.

We develop fit-to-purpose approaches for our two regularisers in sections 3.4.1 and 3.4.2. The first - based on the decomposition of the kernel matrices into the corresponding empirical features and on the variational formulation of norms (Bach et al. 2012) - shows the equivalence of the problem with group lasso (Yuan and Lin 2006; Zhao and Rocha 2006). The second proposes a simple alternating minimisation algorithm to obtain the two sets of parameters.

We introduce the non-negative matrix $\mathbf{A} \in \mathbb{R}_+^{l \times m}$ such that

$$\mathbf{A} = \left( diag(\mathbf{L}^{11}), \ldots, diag(\mathbf{L}^{ms_m}) \right)^T \tag{3.10}$$

(each row in $\mathbf{A}$ corresponds to the diagonal of one output kernel; if $s_j = 1$ for all $j$ we have $A_{js} = \mathbf{L}_{ss}^{(j1)}$). Using this change of variable, the optimisation problem (3.7) can be written equivalently as

$$\arg\min_{\mathbf{A},\mathbf{C}} \ddot{J}(\mathbf{C}, \mathbf{A})$$
$$\ddot{J}(\mathbf{C}, \mathbf{A}) := \ddot{R}(\mathbf{C}, \mathbf{A}) + \tau \ddot{Q}(\mathbf{A}) \quad , \tag{3.11}$$

where

$$
\begin{aligned}
\ddot{R}(\mathbf{C}, \mathbf{A}) &= \sum_s^m \left( \left\| \mathbf{Y}_{:s} - \sum_{ji} A_{(ji)s} \mathbf{K}^{ij} \mathbf{C}_{:s} \right\|_2^2 + \lambda \sum_{ji} A_{(ji)s} \mathbf{C}_{:s}^T \mathbf{K}^{(ji)} \mathbf{C}_{:s} \right) \\
&= \sum_s^m \left( \ddot{R}_s(\mathbf{C}_{:s}, \mathbf{A}_{:s}) \right) , 
\end{aligned}
\tag{3.12}
$$

and $\ddot{Q}(\mathbf{A})$ is the equivalent of $Q(\mathcal{L})$ so that

$$\ddot{Q}_1(\mathbf{A}) = \|\mathbf{A}\|_1 = \sum_{rs} |A_{rs}| \tag{3.13}$$

and

$$\ddot{Q}_{1/2}(\mathbf{A}) = \sum_{js} \sqrt{\sum_i \left( A_{(ji)s} \right)^2} \tag{3.14}$$

In equations (3.12) and (3.14) we somewhat abuse the notation by using $\sum_{ji} A_{(ji)s}$

to indicate the sum across the rows of the matrix $\mathbf{A}$.

From equations (3.12)-(3.14) we observe that, with both of our regularizers, problem (3.11) is conveniently separable along $s$ into the sum of $m$ smaller independent problems, one per scalar output series. These can be efficiently solved in parallel, which makes our method scalable to very large multivariate systems. The final complexity depends on the choice of the regulariser $Q$ and the appropriate algorithm. The overhead cost can be significantly reduced by precalculating the gram matrices $\mathbf{K}^{(ij)}$ in a single preprocessing step and sharing these in between the $m$ parallel tasks.

### 3.4.1 Learning with $\ell_1$ norm

To unclutter the notation we replace the bracketed double superscripts $(ij)$ by a single superscript $d = 1, \ldots, l$. We also drop the regularization parameter $\tau$ (fix it to $\tau = 1$) as it is easy to show (see the appendix) that any other value can be absorbed into the rescaling of $\lambda$ and the $\mathbf{C}$ and $\mathbf{A}$ matrices. For each of the $s$ parallel tasks we indicate $\mathbf{A}_{:s} = \mathbf{a}$, $\mathbf{C}_{:s} = \mathbf{c}$ and $\mathbf{Y}_{:s} = \mathbf{y}$ so that the individual problems are the minimisations with respect to $\mathbf{a} \in \mathbb{R}^l_+$ and $\mathbf{c} \in \mathbb{R}^n$ of

$$P(\mathbf{c}, \mathbf{a}) := \left\| \mathbf{y} - \sum_d a_d \mathbf{K}^d \mathbf{c} \right\|_2^2 + \lambda \sum_{ji} a_d \mathbf{c}^T \mathbf{K}^d \mathbf{c} + \sum_d a_d \ . \tag{3.15}$$

We decompose (for example by eigendecomposition) each of the gram matrices as $\mathbf{K}^d = \mathbf{\Phi}^d (\mathbf{\Phi}^d)^T$, where $\mathbf{\Phi}^d \in \mathbb{R}^{n \times n}$ is the matrix of the empirical features, and we introduce the variables $\mathbf{z}^d = a_d (\mathbf{\Phi}^d)^T \mathbf{c} \in \mathbb{R}^n$ and the set $\mathcal{Z} = \{\mathbf{z}^d : \mathbf{z}^d \in \mathbb{R}^n, d \in \mathbb{N}_l\}$. Using these we rewrite[1] equation (3.15)

$$\widetilde{P}(\mathcal{Z}, \mathbf{a}) := \left\| \mathbf{y} - \sum_d \mathbf{\Phi}^d \mathbf{z}^d \right\|_2^2 + \sum_d \left( \frac{\lambda \|\mathbf{z}^d\|_2^2}{a_d} + a_d \right) \ . \tag{3.16}$$

We first find the closed form of the minimising solution for $\mathbf{a}$ as $a_d = \sqrt{\lambda} \|\mathbf{z}^d\|_2$

---

[1] We extend the function $x^2/y : \mathbb{R} \times \mathbb{R}_+ \to \mathbb{R}_+$ to the point $(0,0)$ by taking the convention $0/0 = 0$.

for all $d$. Plugging this back to (3.16) we obtain

$$\min_{\mathbf{a}} \widetilde{P}(\mathcal{Z}, \mathbf{a}) = \left\| \mathbf{y} - \sum_d \mathbf{\Phi}^d \mathbf{z}^d \right\|_2^2 + 2\sqrt{\lambda} \sum_d \left\| \mathbf{z}^d \right\|_2 \ . \tag{3.17}$$

Seen as a minimisation with respect to the set $\mathcal{Z}$ this is the classical group-lasso formulation with the empirical features $\mathbf{\Phi}^d$ as inputs. Accordingly, it can be solved by any standard method for group-lasso problems such as the proximal gradient descent method, e.g. Bach et al. (2012), which we employ in our experiments. After solving for $\mathcal{Z}$ we can directly recover $\mathbf{a}$ from the above minimising identity and then obtain the parameters $\mathbf{c}$ from the set of linear equations

$$\left( \sum_d a_d \mathbf{K}^d + \lambda \mathbf{I}_n \right) \mathbf{c} = \mathbf{y} \ . \tag{3.18}$$

The algorithm outlined above takes advantage of the convex group-lasso reformulation (3.17) and has the standard convergence and complexity properties of proximal gradient descent. The empirical features $\mathbf{\Phi}^d$ can be pre-calculated and shared amongst the $m$ tasks to reduce the overhead cost.

### 3.4.2 Learning with $\ell_1/\ell_2$ norm

For the $\ell_1/\ell_2$ regularization, we need to return to the double indexation $(ji)$ to make clear how the groups are created. As above, for each of the $s$ parallel tasks we use the vectors $\mathbf{a}, \mathbf{c}$ and $\mathbf{y}$. However, for vector $\mathbf{a}$ we will keep the $(ji)$ notation for its elements. The individual problems are the minimisations with respect to $\mathbf{a} \in \mathbb{R}_+^l$ and $\mathbf{c} \in \mathbb{R}^n$ of

$$P(\mathbf{c}, \mathbf{a}) := \left\| \mathbf{y} - \sum_{ji} a_{(ji)} \mathbf{K}^{(ji)} \mathbf{c} \right\|_2^2 + \lambda \sum_{ji} a_{(ji)} \mathbf{c}^T \mathbf{K}^{(ji)} \mathbf{c} + \sum_j \sqrt{\sum_i a_{(ji)}^2} \tag{3.19}$$

We propose to use the alternating minimisation with a proximal gradient step. At each iteration, we alternatively solve for $\mathbf{c}$ and $\mathbf{a}$. For fixed $\mathbf{a}$ we obtain $\mathbf{c}$ from the set of linear equations (3.18). With fixed $\mathbf{c}$, problem (3.19) is a group lasso for **a** with groups defined by the sub-index $j$ within the double $(ji)$ indexation of the elements of **a**. Here, the proximal gradient step takes place to move along the descend direction for **a**. Though convex in **a** and **c** individually, the problem (3.19) is jointly non-convex and therefore can converge to local minima.

## 3.5  Interpretation and crossovers

To help the understanding of the inner workings of our methods and especially the effects of the two regularizers, we discuss here the crossovers to other existing methods for MKL and Granger modelling.

$\ell_1$ **norm**   The link to group-lasso demonstrated in section 3.4.1 is not in itself too surprising. The formulation in (3.15) can be recognised as a sparse MKL problem which has been previously shown to relate to group-lasso (eg. Bach (2008), Xu et al. (2010)). We derive this link in section 3.4.1 using the empirical feature representation to i) provide better intuition for the structure of the learned function $\widehat{\mathbf{f}}$, ii) develop an efficient algorithm for solving problem (3.15).

The re-formulation in terms of the empirical features $\mathbf{\Phi}^d$ creates an intuitive bridge to the classical linear models. Each $\mathbf{\Phi}^d$ can be seen as a matrix of features generated from a subset $\mathcal{X}^{(j)}$ of the input coordinates relating to the past of a single scalar time series $j$. The group-lasso regularizer in equation (3.17) has a sparsifying effect at the level of these subsets zeroing out (or not) the whole groups of parameters $\mathbf{z}^d$. In the context of linear methods, this approach is known as the grouped graphical Granger modelling (Lozano et al. 2009).

Within the non-linear approaches to time series modelling, Sindhwani et al. (2013) recently derived a similar formulation. There the authors followed a strategy of multiple kernel learning from a dictionary of input kernels combined with a single learned output kernel (as opposed to our multiple output kernels). They obtain their IKL model, which is in its final formulation equivalent to problem (3.15), by fixing the output kernel to identity.

Though we initially formulate our problem quite differently, the diagonal constraint we impose on the output kernels essentially prevents the modelling of any contemporaneous relationships between the series (as does the identity output kernel matrix in IKL). What remains in our methods are the diagonal elements, which are non-constant and sparse, and which can be interpreted as the weights of the input kernels in the standard MKL setting.

$\ell_1/\ell_2$ **norm** The more complex $\ell_1/\ell_2$ regularisation discussed in section 3.4.2 is to the best of our knowledge novel in the context of multiple kernel learning. It has again a strong motivation and clear interpretation in terms of the graphical Granger modelling. The norm has a sparsifying effect not only at the level of the individual kernels but at the level of the groups of kernels operating over the same input partitions $\mathcal{X}^{(j)}$. In this respect our move from the $\ell_1$ to the $\ell_1/\ell_2$ norm has a parallel in the same move in linear graphical Granger techniques. The $\ell_1$ norm lasso-Granger method (Arnold et al. 2007) imposes the sparsity on the individual elements of the parameter matrices in a linear model, while the $\ell_1/\ell_2$ of the grouped-lasso-Granger (Lozano et al. 2009) works with groups of the corresponding parameters of a single input series across the multiple lags $p$.

## 3.6 Experiments

To document the performance of our method, we have conducted a set of experiments on real and synthetic datasets. In these we simulate real-life forecasting exercise by splitting the data into a training and a hold-out set which is unseen by the algorithm when learning the function $\widehat{\mathbf{f}}$ and is only used for the final performance evaluation.

We compare our methods with the output kernel $\ell_1$ regularization (NVARL1) and and $\ell_1/\ell_1$ (NVARL12) with simple baselines (which nevertheless are often hard to beat in practical time series forecasting) as well as with the state-of-the-art techniques for forecasting and Granger modelling. Namely, we compare with simple mean and univariate linear autoregressive models (LAR), multivariate linear vector autoregressive model with $\ell_2$ penalty (LVARL2), the group-lasso Granger method (Lozano et al. 2009) (LVARL1), and a sparse MKL without the $\mathcal{X}^{(j)}$ input partitioning (NVAR). Of these, the last two are the most relevant competitors. LVARL1, similarly to our methods, aims at recovering the Granger structure but is strongly constrained to linear modelling only. NVAR has no capability to capture the Granger relationships but, due to the lack of structural constraints, it is the most flexible of all the models.

We evaluate our results with respect to the two objectives for the function $\widehat{f}$ defined in section 3.2. We measure the accuracy of the one-step ahead forecasts by the mean squared error (MSE) for the whole multivariate process averaged over 500 hold-out points. The structural objective allowing the analysis of dependencies between the sub-processes is wired into the method itself (see sections 3.3 and 3.2.1) and is therefore satisfied by construction. We produce adjacency matrices of the graphs of the learned dependencies, compare these with the ones produced by the linear Granger methods and comment on the observed results.

### 3.6.1 Technical considerations

For each experiment we preprocessed the data by removing the training sample mean and rescaling with the training sample standard deviation. We fix the number of kernels for each input partition to six ($s_j = 6$ for all $j$) and use the same kernel functions for all experiments: a linear, 2nd order and 3rd polynomial, and Gaussian kernels with width 0.5, 1 and 2. We normalise the kernels so that the training Gram matrices have trace equal to the size of the training sample.

We search for the hyper-parameter $\lambda$ by a 5-fold cross-validation within a 15-long logarithmic grid $\lambda \in \{10^{-3}, \dots, 10^4\}\sqrt{n}l$, where $n$ is the training sample size and $l$ is the number of kernels or groups (depending on the method). In each grid search, we use the previous parameter values as warm starts. We do not perform an exhaustive search for the optimal lag for each of the scalar input series by some of the classical testing procedures (based on AIC, BIC etc.). We instead fix it to $p = 5$ for all series in all experiments and rely on the regularization to control any excess complexity.

We implemented our own tools for all the tested methods based on variations of proximal gradient descent with ISTA line search Beck and Teboulle (2009). The full Matlab code is available at https://bitbucket.org/dmmlgeneva/nonlinear-granger

### 3.6.2 Synthetic experiments

We have simulated data from a five dimensional non-Gaussian time-series process generated through a linear filter of a 5-dimensional IID exponential white noise $\mathbf{e}_t$ with identity covariance matrix (re-centered to zero and re-scaled to unit variance). The matrix $\mathbf{\Psi} = [0.7, 1.3, 0, 0, 0; 0, 0.6, -1.5, 0, 0; 0, -1.2, 1.46, 0, 0; 0, 0, 0, 0.6, 1.4; 0, 0, 0, 1.3, -0.5]$ in the filter $\mathbf{y}_t = \mathbf{e}_t + \mathbf{\Psi}\mathbf{e}_{t-1}$ is such that the process consists of two independent internally interrelated sub-processes, one composed of the first 3 scalar series, the other of the remaining two series. This structural information, though known to us, is unknown to the learning methods (not considered in the learning process).

We list in Table 3.6.1 the predictive performance of the tested methods in terms of the average hold-out MSE based on training samples of varying size. Our methods clearly outperform all the linear models. The functionally strongly constrained linear LVARL1 performs roughly on par with our methods for the small sample sizes. But for larger sample sizes, the higher flexibility of the function space in our methods yields significantly more accurate forecasts (as much as 10% MSE improvement).

The structural constraints in our methods also help the performance when competing with the unstructured NVAR method, which has mostly less accurate forecasts. At the same time, as illustrated in figure 3.6.1, our methods are able to correctly recover the Granger-causality structure (splitting the process into the two independent subprocesses by the zero off-diagonal blocks), which NVAR by construction cannot.

### 3.6.3 Real data experiments

We use data on water physical discharge publicly available from the website of the Water Services of the US geological survey (http://www.usgs.gov/). Our

Table 3.6.1: Synthetic experiments: MSE (std) for 1-step ahead forecasts (hold-out sample average)

| Train size | 300 | | 700 | | 1000 | |
|---|---|---|---|---|---|---|
| Mean | 0.925 | (0.047) | 0.923 | (0.047) | 0.923 | (0.047) |
| LAR | 0.890 | (0.045) | 0.890 | (0.044) | 0.890 | (0.044) |
| LVAR | 0.894 | (0.045) | 0.836 | (0.041) | 0.763 | (0.035) |
| LVARL1 | 0.787 | (0.037) | 0.737 | (0.031) | 0.722 | (0.030) |
| NVAR | 0.835 | (0.041) | 0.735 | (0.032) | 0.719 | (0.030) |
| NVARL1 | *0.754* | (0.034) | 0.706 | (0.030) | **0.679** | (0.028) |
| NVARL12 | 0.808 | (0.040) | 0.710 | (0.031) | 0.684 | (0.029) |
| Train size | 1500 | | 2000 | | 3000 | |
| Mean | 0.923 | (0.047) | 0.922 | (0.047) | 0.922 | (0.047) |
| LAR | 0.888 | (0.045) | 0.889 | (0.045) | 0.888 | (0.045) |
| LVAR | 0.751 | (0.034) | 0.741 | (0.033) | 0.687 | (0.028) |
| LVARL1 | 0.710 | (0.029) | 0.701 | (0.028) | 0.693 | (0.028) |
| NVAR | 0.699 | (0.028) | 0.682 | (0.027) | 0.662 | (0.026) |
| NVARL1 | ***0.654*** | (0.026) | ***0.640*** | (0.025) | **0.626** | (0.025) |
| NVARL12 | **0.659** | (0.027) | 0.685 | (0.028) | **0.657** | (0.027) |

In brackets is the average standard deviation (std) of the MSEs. Results for NVARL1 and NVARL12 in **bold** are significantly better than **all** the linear competitors, in *italics* are significantly better than the non-linear NVAR (using one-sided paired-sample t-test at 10% significance level).

dataset consists of 9 time series of daily rates of year-on-year growth at measurement sites along the streams of Connecticut and Columbia rivers.

The prediction accuracy of the tested methods is listed in Table 3.6.2. Our non-linear methods perform on par with the state-of-the-art linear models. This on one hand suggests that for the analysed dataset the linear modelling seems sufficient. On the other hand, it confirms that our methods, which in general have the ability to learn more complex relationships by living in a richer functional space, are well behaved and can capture simpler dependencies as well. The structure encoded into our methods, however, benefits the learning since the unstructured NVAR tends to perform less accurately.

The learned dynamical dependence structure of the time series is depicted in figure 3.6.1. In the dataset (and the adjacency matrices), the first 4 series are the Connecticut measurement sites starting from the one highest up the stream and moving down to the mouth of the river. The next 5 our the

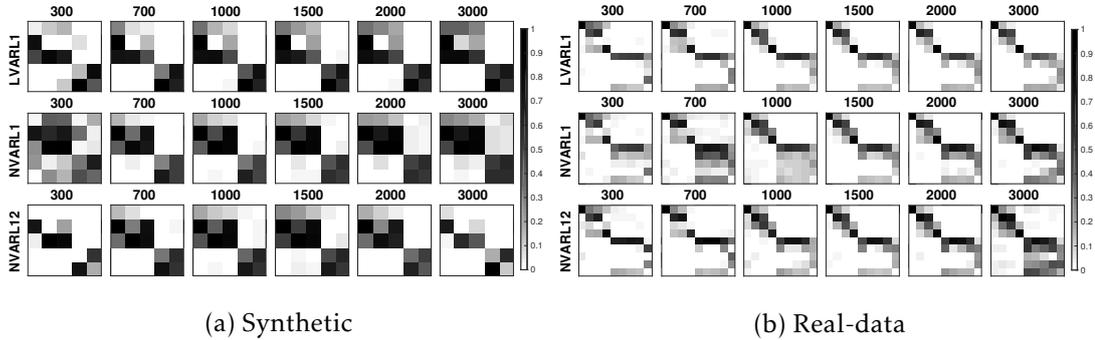

(a) Synthetic                      (b) Real-data

Figure 3.6.1: Schematics of the learned adjacency matrices of Granger-causality graphs for the three sparse learning methods across varying training sample size. A scalar time series $y_i$ does not Granger-cause series $y_j$ (given all the other series) if the element $e_{ij}$ in the adjacency matrix is zero (white). The displayed adjacency matrices were derived from the learned matrices **A** by summing the respective elements across individual kernels. The values are rescaled so that the largest element in each matirx is equal to 1 (black).

Columbia measurement sites ordered in the same manner.

From inspecting the learned adjacency matrices, we observe that all the sparse methods recover similar Granger-causal structures. Since we do not know the ground truth in this case, we can only speculate about the accuracy of the structure recovery. Nevertheless, it seems plausible that there is little dynamical cross-dependency between the Connecticut and Columbia measurements as the learned graphs suggest (the two rivers are at the East and West extremes of the US).

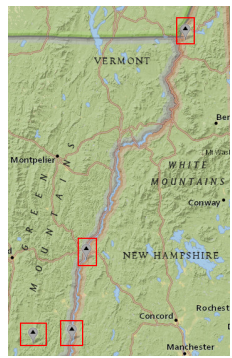                                          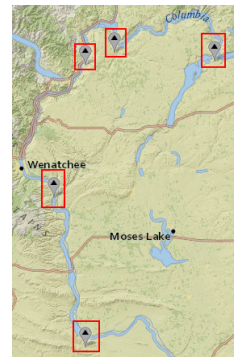

(a) Connecticut                          (b) Columbia

Figure 3.6.2: Measurements sites along the river streams for USGS water data.

Table 3.6.2: Real-data experiments: MSE (std) for 1-step ahead forecasts (hold-out sample average)

| Train size | 300 | | 700 | | 1000 | |
|---|---|---|---|---|---|---|
| Mean | 0.780 | (0.053) | 0.795 | (0.054) | 0.483 | (0.026) |
| LAR | 0.330 | (0.023) | 0.340 | (0.024) | 0.152 | (0.013) |
| LVARL2 | 0.302 | (0.021) | 0.311 | (0.022) | 0.140 | (0.012) |
| LVARL1 | 0.310 | (0.022) | 0.310 | (0.023) | 0.140 | (0.012) |
| NVAR | 0.328 | (0.023) | 0.316 | (0.023) | 0.148 | (0.012) |
| NVARL1 | 0.308 | (0.023) | 0.317 | (0.024) | 0.140 | (0.012) |
| NVARL12 | 0.321 | (0.023) | 0.322 | (0.024) | 0.141 | (0.012) |
| Train size | 1500 | | 2000 | | 3000 | |
| Mean | 0.504 | (0.03) | 0.464 | (0.027) | 0.475 | (0.017) |
| LAR | 0.181 | (0.015) | 0.179 | (0.013) | 0.187 | (0.008) |
| LVARL2 | 0.167 | (0.014) | 0.164 | (0.013) | 0.170 | (0.007) |
| LVARL1 | 0.165 | (0.014) | 0.163 | (0.013) | 0.170 | (0.008) |
| NVAR | 0.169 | (0.014) | 0.166 | (0.012) | 0.173 | (0.007) |
| NVARL1 | 0.164 | (0.014) | 0.161 | (0.013) | 0.167 | (0.007) |
| NVARL12 | 0.162 | (0.014) | 0.160 | (0.012) | 0.166 | (0.007) |

In brackets is the average standard deviation (std) of the MSEs.

## 3.7 Conclusions

We have developed a new method for forecasting and Granger-causality modelling in multivariate time series that does not rely on prior assumptions about the shape of the dynamical dependencies (other than being sparse). The method is based on learning a combination of multiple operator-valued kernels in which the multiple output kernels are learned as sparse diagonal matrices. We have documented on experiments that our method outperforms linear competitors in the presence of strong non-linearities and is able to correctly recover the Granger-causality structure which non-structured kernel methods cannot do.

# Appendix

**In the output kernel learning problem, the second hyperparameter $\tau$ can be fixed to a constant as its values can be absorbed into the scaling of $\lambda$, and the $\mathbf{C}$ and $\mathbf{A}$ matrices.**

*Proof.* The general output kernel learning problem can be written as

$$\min_{\mathbf{C},\mathbf{L}} J(\mathbf{C}, \mathbf{L}) = \|\mathbf{Y} - \mathbf{KCL}\|_F^2 + \lambda \left\langle \mathbf{C}'\mathbf{KC}, \mathbf{L} \right\rangle_F + \tau \, Q(\mathbf{L}), \tag{3.20}$$

where $\mathbf{Y}$ is the $n \times m$ output data matrix, $\mathbf{K}$ is the $n \times n$ input-kernel Gram matrix, $\mathbf{L}$ is the $m \times m$ output-kernel Gram matrix, $\mathbf{C}$ is the $n \times m$ parameters matrix and $Q(.)$ is the regularizer on $\mathbf{L}$ which we assume to be homogenous so that $Q(\alpha \mathbf{L}) = \alpha^k Q(\mathbf{L})$ for some $k$.

We introduce the change of variables $\tau^{1/k}\mathbf{L} = \widetilde{\mathbf{L}}$ and $\mathbf{C} = \tau^{1/k}\widetilde{\mathbf{C}}$ where the matrices with tildes are simply scaled versions of the original matrices. Using these we can rewrite problem (3.20) into an equivalent minimization

$$
\begin{aligned}
\min_{\widetilde{\mathbf{C}},\widetilde{\mathbf{L}}} J(\widetilde{\mathbf{C}}, \widetilde{\mathbf{L}}) &= \left\| \mathbf{Y} - \mathbf{K}\tau^{1/k}\widetilde{\mathbf{C}}\frac{1}{\tau^{1/k}}\widetilde{\mathbf{L}} \right\|_F^2 + \lambda \left\langle \tau^{1/k}\widetilde{\mathbf{C}}\mathbf{K}\tau^{1/k}\widetilde{\mathbf{C}}, \frac{1}{\tau^{1/k}}\widetilde{\mathbf{L}} \right\rangle_F + Q(\widetilde{\mathbf{L}}) \\
&= \left\| \mathbf{Y} - \mathbf{K}\widetilde{\mathbf{C}}\widetilde{\mathbf{L}} \right\|_F^2 + \lambda\tau^{1/k} \left\langle \widetilde{\mathbf{C}}\mathbf{K}\widetilde{\mathbf{C}}, \widetilde{\mathbf{L}} \right\rangle_F + Q(\widetilde{\mathbf{L}}), \tag{3.21}
\end{aligned}
$$

where the 2nd regularization parameter $\tau$ has been absorbed into the first regularization parameter and the scaling of the $\mathbf{C}$ and $\mathbf{L}$ matrices. □

From the above we see that we can fix $\tau$ arbitrarily (and hence for example set it to $\tau = 1$) and only grid search for $\lambda$ to find the optimal combination of the parameter and output-kernel matrices. If we changed the value of $\tau$ we could get the same regularization path by adjusting the $\lambda$ grid accordingly. The minimizing solutions $\widetilde{\mathbf{C}}$ and $\widetilde{\mathbf{L}}$ would be the scaled version of $\mathbf{C}$ and $\mathbf{L}$ but would yield the same objective values $J(.)$.

In consequence, not only we *can* drop the second regularization parameter but we *should* drop it (unless we fix the scale of $\mathbf{C}$). Otherwise, for every combination $\lambda$ and $\tau$ we can find a combination $\tilde{\lambda}$, $\tilde{\tau}$ which will yield the same minimum of the objective with different scalings of the learned matrices $\mathbf{C}$ and $\mathbf{L}$.

# Chapter 4

# Structured Nonlinear Variable Selection

**Chapter abstract:** We investigate structured sparsity methods for variable selection in regression problems where the target depends nonlinearly on the inputs. We focus on general nonlinear functions not limiting a priori the function space to additive models. We propose two new regularizers based on partial derivatives as nonlinear equivalents of group lasso and elastic net. We formulate the problem within the framework of learning in reproducing kernel Hilbert spaces and show how the variational problem can be reformulated into a more practical finite dimensional equivalent. We develop a new algorithm derived from the ADMM principles that relies solely on closed forms of the proximal operators. We explore the empirical properties of our new algorithm for Nonlinear Variable Selection based on Derivatives (NVSD) on a set of experiments and confirm favourable properties of our structured-sparsity models and the algorithm in terms of both prediction and variable selection accuracy.

## 4.1 Introduction

We are given a set of $n$ input-output pairs $\{(\mathbf{x}^i, y^i) \in (\mathcal{X} \times \mathcal{Y}) : \mathcal{X} \subseteq \mathbb{R}^d, \mathcal{Y} \subseteq \mathbb{R}, i \in \mathbb{N}_n\}$ sampled i.i.d. according to an unknown probability measure $\rho$. Our task is to learn a regression function $f : \mathcal{X} \to \mathcal{Y}$ with minimal expected squared error loss $\mathrm{L}(f) = \mathrm{E}\,(y - f(\mathbf{x}))^2 = \int (y - f(\mathbf{x}))^2\,\mathrm{d}\rho(\mathbf{x}, y)$.

We follow the standard theory of regularised learning where $\widehat{f}$ is learned by minimising the regularised empirical squared error loss $\widehat{\mathrm{L}}(f) = \frac{1}{n} \sum_i^n \left( y^i - f(\mathbf{x}^i) \right)^2$

$$\widehat{f} = \operatorname*{arg\,min}_f \widehat{\mathrm{L}}(f) + \tau \mathrm{R}(f) \ . \tag{4.1}$$

In the above, $\mathrm{R}(f)$ is a suitable penalty typically based on some prior assumption about the function space (e.g. smoothness), and $\tau > 0$ is a suitable regularization hyper-parameter. The principal assumption we consider in our work is that the function $f$ is sparse with respect to the original input space $\mathcal{X}$, that is it depends only on $l \ll d$ input variables.

Learning with variable selection is a well-established and rather well-explored problem in the case of linear models $f(\mathbf{x}) = \sum_a^d x_a w_a$, e.g. Hastie et al. (2015). The main ideas from linear models have been successfully transferred to additive models $f(\mathbf{x}) = \sum_a^d f_a(x_a)$, e.g. Ravikumar et al. (2007), Bach (2009), Koltchinskii and Yuan (2010), and Yin et al. (2012), or to additive models with interactions $f(\mathbf{x}) = \sum_a^d f_a(x_a) + \sum_{a<b}^d f_{a,b}(x_a, x_b)$, e.g Lin and Zhang (2006) and Tyagi et al. (2016).

However, sparse modelling of general non-linear functions is more intricate. A promising stream of works focuses on the use of non-linear (conditional) crosscovariance operators arising from embedding probability measures into Hilbert function spaces, e.g. Yamada et al. (2014) and Chen et al. (2017).

In this work, we follow an alternative approach proposed in Rosasco et al. (2013) based on partial derivatives and develop new regularizers to promote structured sparsity with respect to the original input variables. We stress that our objective here is not to learn new data representations nor learn sparse models in some latent feature space, e.g. Gurram and Kwon (2014). Nor is it to learn models sparse in the data instances (in the sense of support vectors, e.g. Chan et al. (2007)). We aim at selecting the relevant input variables, the relevant dimensions of the input vectors $\mathbf{x} \in \mathbb{R}^d$.

After a brief review of the regularizers used in Rosasco et al. (2013) for individual variable selection in non-linear model learning (similar in spirit to lasso of Tibshirani (1996)) we propose two extensions motivated by the linear structured-sparsity learning literature. Using suitable norms of the partial derivatives we propose the non-linear versions of the group lasso Yuan and Lin (2006) and the elastic net Zou and Hastie (2005) (see also section 1.2.2).

We pose our problem into the framework of learning in the reproducing kernel Hilbert space (RKHS). We extend the representer theorem to show that the minimiser of (4.1) with our new regularizers R$(f)$ can be conveniently written as a linear combination of kernel functions and their partial derivatives evaluated over the training set.

We further propose a new reformulation of the equivalent finite dimensional learning problem, which allows us to develop a new algorithm (NVSD) based on the alternating direction method of multipliers (ADMM) (Boyd 2010). This is a generic algorithm that can be used (with small alterations) for all regularizers we discuss here. At each iteration, the algorithm needs to solve a single linear problem, perform a proximal step resulting in a soft-thresholding operation, and do a simple additive update of the dual variables. Unlike Rosasco et al. (2013), which uses approximations of the proximal operator, our algorithm is based on proximals admitting closed forms for all the discussed regularizers, including the one suggested previously in Rosasco et al. (2013). Furthermore, by avoiding the approximations in the proximal step, the algorithm directly provides also the learned sparsity patterns over the training set (up to the algorithmic convergence precision).

We explore the effect of the proposed regularizers on model learning on synthetic and real-data experiments, and confirm the superior performance of our methods in comparison to a range of baseline methods when learning structured-sparse problems. Finally, we conclude by discussing the advantages and shortcomings of the current proposal and outline some directions for future work.

## 4.2 Regularizers for variable selection

In Rosasco et al. (2013) the authors propose to use the partial derivatives of the function with respect to the input vector dimensions $\{\partial_a f : a \in \mathbb{N}_d\}$ to construct a regularizer promoting sparsity. The partial derivative evaluated at an input point $\partial_a f(\mathbf{x})$ is the rate of change of the function at that point with respect to $x_a$ holding the other input dimensions fixed. Intuitively, when the function does not dependent on an input variable (input dimension $a$), its evaluations do not change with changes in the input variable: $\partial_a f(\mathbf{x}) = 0$ at all points $\mathbf{x} \in \mathcal{X}$. A natural measure of the size of the partial derivatives across the space $\mathcal{X}$ is the $L_2$ norm

$$\|\partial_a f\|_{L_2} = \sqrt{\int_{\mathcal{X}} |\partial_a f(\mathbf{x})|^2 \, \mathrm{d}\rho_x(\mathbf{x})} \tag{4.2}$$

**Remark 1.** *At this point we wish to step back and make a link to the linear models $f(\mathbf{x}) = \sum_a^d x_a w_a$. The partial derivatives with respect to any of the $d$ dimensions of the input vector $\mathbf{x}$ are the individual elements of the $d$-dimensional parameter vector $\mathbf{w}$, $\partial_a f(\mathbf{x}) = w_a$, and this at every point $\mathbf{x} \in \mathcal{X}$. For the linear model we thus have $\|\partial_a f\|_{L_2} = |w_a|$. Sparsity inducing norms or constraints operating over the parameter vectors $\mathbf{w}$ can therefore be seen as special cases of the same norms and constraints imposed on the partial-derivative norms (4.2).*

### 4.2.1 Sparsity inducing norms

The sparsity objective over a vector $\mathbf{v} \in \mathbb{R}^d$ can be cast as the minimization of the $\ell_0$ norm $\|\mathbf{v}\|_0 = \#\{a = 1, \ldots, d : v_a \neq 0\}$ which counts the number of non-zero elements of the vector. Since it is well known from the linear sparse learning literature that finding the $\ell_0$ solutions is computationally difficult in higher dimensions (NP-hard, Weston et al. (2003)), the authors in Rosasco et al. (2013) suggest to use its tightest convex relaxation, the $\ell_1$ norm $\|\mathbf{v}\|_1 = \sum_a^d |v_a|$. They apply the $\ell_1$ norm over the partial-derivative norms (4.2) so that the lasso-like sparsity regularizer in (4.1) is

$$\mathrm{R}^L(f) = \sum_{a=1}^d \|\partial_a f\|_{L_2} \ . \tag{4.3}$$

In this chapter we explore two extensions inspired by the linear sparse learning, opening the doors to many of the other sparsity and structured sparsity inducing norms that have been proposed in the abundant literature on this topic. Namely, we focus here on the structured sparsity induced by the mixed $\ell_1/\ell_2$ norm known in the context of linear least squares as the group lasso Yuan and Lin (2006). For a vector $\mathbf{v}$ composed of $G$ groups $\mathbf{v}_g$ (non-overlapping but not necessarily consecutive) with $p_g$ number of elements each, the mixed $\ell_1/\ell_2$ norm is $\|\mathbf{v}\|_{1,2} = \sum_g^G p_g \|\mathbf{v}_g\|_2$. The corresponding group-lasso-like regularizer to be used in (4.1) is

$$R^{GL}(f) = \sum_{g=1}^{G} p_g \sqrt{\sum_{a \in g} \|\partial_a f\|_{L_2}^2} \ .$$ (4.4)

Second, we look at the elastic net penalty proposed initially in Zou and Hastie (2005). This uses a convex combination of the $\ell_1$ and square of the $\ell_2$ norm and has been shown to have better selection properties over the vanilla $\ell_1$ norm regularization in the presence of highly correlated features. Unlike the $\ell_1$ penalty, the combined elastic net is also strictly convex. The corresponding elastic-net-like regularizer to be used in (4.1) is

$$R^{EN}(f) = \mu \sum_{a=1}^{d} \|\partial_a f\|_{L_2} + (1-\mu) \sum_{a=1}^{d} \|\partial_a f\|_{L_2}^2, \quad \mu \in [0,1] \ .$$ (4.5)

### 4.2.2 Empirical versions of regularizers

A common problem of the regularizers introduced above is that in practice they cannot be evaluated due to the unknown probability measure $\rho_x$ on the input space $\mathcal{X}$. Therefore instead of the partial-derivative norms defined in expectation in (4.2)

$$\|\partial_a f\|_{L_2} = \sqrt{\mathrm{E}\left[|\partial_a f(\mathbf{x})|^2\right]}$$ (4.6)

we use their sample estimates replacing the expectation by the training sample average

$$\|\partial_a f\|_{2_n} = \sqrt{\frac{1}{n} \sum_i^n |\partial_a f(\mathbf{x}^i)|^2} \ .$$ (4.7)

This corresponds to the move from expected loss to the empirical loss introduced in section 4.1 and is enabled by the IID sample assumptions.

In result, the regression function is learned from the empirical version of (4.1)

$$\widehat{f} = \arg\min_{f \in \mathcal{F}} \widehat{L}(f) + \tau \widehat{R}(f) \ , \tag{4.8}$$

where $\widehat{R}(f)$ are the empirical analogues of the regularizers (4.3), (4.4) and (4.5) replacing the population partial-derivative norms $\|\partial_a f\|_{L_2}$ by their sample estimates $\|\partial_a f\|_{2_n}$. The function space $\mathcal{F}$ is discussed next.

## 4.3 Learning in RKHS

The hypothesis space $\mathcal{F}$ within which we learn the function $f$ is a reproducing kernel Hilbert space (RKHS). We recall (section 1.3) that a RKHS is a function space $\mathcal{H}_K$ of real-valued functions over $\mathcal{X}$ endowed with an inner product $\langle \cdot, \cdot \rangle_{\mathcal{H}_K}$ and the induced norm $\|.\|_{\mathcal{H}_K}$ that is uniquely associated with a positive semidefinite kernel $k : \mathcal{X} \times \mathcal{X} \to \mathbb{R}$. The kernel $k$ has the reproducing property $\langle k_{\mathbf{x}}, f \rangle_{\mathcal{H}_K} = f(\mathbf{x})$ and, in particular, $\langle k_{\mathbf{x}}, k_{\mathbf{x}'} \rangle_{\mathcal{H}_K} = k(\mathbf{x}, \mathbf{x}')$, where $k_{\mathbf{x}} \in \mathcal{H}_K$ is the kernel section centred at $\mathbf{x}$ such that $k_{\mathbf{x}}(\mathbf{x}') = k(\mathbf{x}, \mathbf{x}')$ for any two $\mathbf{x}, \mathbf{x}' \in \mathcal{X}$. Furthermore, the space $\mathcal{H}_K$ is the completion of the linear span of the functions $\{k_{\mathbf{x}} : \mathbf{x} \in \mathcal{X}\}$.

In addition to these fairly well known properties of the RKHS and its kernel, the author in Zhou (2008) has shown that if $k$ is continuous and sufficiently smooth the kernel partial-derivative functions belong to the RKHS and have a partial-derivative reproducing property. More specifically, we define the kernel partial-derivative function $[\partial_a k_{\mathbf{x}}] : \mathcal{X} \to \mathbb{R}$ as

$$[\partial_a k_{\mathbf{x}}](\mathbf{x}') = \frac{\partial}{\partial x_a} k(\mathbf{x}, \mathbf{x}') \quad \forall \mathbf{x}, \mathbf{x}' \in \mathcal{X} \ . \tag{4.9}$$

The function $[\partial_a k_{\mathbf{x}}] \in \mathcal{H}_K$ has the reproducing property $\langle [\partial_a k_{\mathbf{x}}], f \rangle_{\mathcal{H}_K} = \partial_a f(\mathbf{x})$. In particular $\langle [\partial_a k_{\mathbf{x}}], k_{\mathbf{x}'} \rangle_{\mathcal{H}_K} = \partial_a k_{\mathbf{x}'}(\mathbf{x})$ and $\langle [\partial_a k_{\mathbf{x}}], [\partial_b k_{\mathbf{x}'}] \rangle_{\mathcal{H}_K} = \frac{\partial^2}{\partial x_a \partial x_b'} k(\mathbf{x}, \mathbf{x}')$.

**Remark 2.** *Since the notation above may seem somewhat knotty at first, we invite the reader to appreciate the difference between the function $[\partial_a k_{\mathbf{x}}]$ and the partial* *derivative of the kernel section with respect to the a-th dimension $\partial_a k_{\mathbf{x}}$. Clearly, $[\partial_a k_{\mathbf{x}}](\mathbf{x}') \neq \partial_a k_{\mathbf{x}}(\mathbf{x}')$ for any $\mathbf{x} \neq \mathbf{x}' \in \mathcal{X}$. However, due to the symmetry of the kernel we do have $[\partial_a k_{\mathbf{x}}](\mathbf{x}') = \partial_a k_{\mathbf{x}'}(\mathbf{x}) = \frac{\partial}{\partial x_a} k(\mathbf{x}, \mathbf{x}')$.*

### 4.3.1 Solution representation

The variational (infinite-dimensional) problem (4.8) is difficult to handle as is. However, it has been previously shown for a multitude of RKHS learning problems that their solutions $\widehat{f}$ can be expressed as finite linear combinations of the kernel evaluations over the training data (Argyriou and Dinuzzo 2014). This property, known as *representer theorem*, renders the problems amenable to practical computations.

**Proposition 1.** *The minimising solution $\widehat{f}$ of the variational problem*

$$\widehat{f} = \underset{f \in \mathcal{H}_K}{\arg\min} \, \widehat{L}(f) + \tau \widehat{R}(f) + \nu \|f\|^2_{\mathcal{H}_K} \quad , \tag{4.10}$$

*where $\tau, \nu \geq 0$ and $\widehat{R}(f)$ is any of the empirical versions of the three formulations (4.3), (4.4), (4.5) can be represented as*

$$\widehat{f} = \sum_{i}^{n} \alpha_i k_{\mathbf{x}^i} + \sum_{i}^{n} \sum_{a}^{d} \beta_{ai} [\partial_a k_{\mathbf{x}^i}] \quad . \tag{4.11}$$

The proof (available in the appendix) follows the classical approach (e.g. Schölkopf et al. (2001) and used also in chapter 1) of decomposition of $\mathcal{H}_K$ into the space spanned by the representation and its orthogonal complement.

The proposition extends the representer theorem of Rosasco et al. (2013) to the new regularizers (4.4) and (4.5). Note that we included the induced Hilbert norm $\|f\|_{\mathcal{H}_K}$ into (4.10) as a useful generalization that reduces to our original problem (4.8) if $\nu = 0$. On the other hand, when $\tau = 0$ we recover a classical kernel regression problem which is known to have another simpler representation consisting just of the first term in (4.11).

## 4.4 Algorithm

In this section we describe the new algorithm we developed to solve problem (4.10) with the three sparse regularizers introduced in section 4.2. The algorithm is versatile so that it requires only small alterations in specific steps to move from one regularizer to the other. Importantly, unlike the algorithm proposed in Rosasco et al. (2013) for solving only the lasso-like problem, our algorithm does not need to rely on proximal approximations since all the proximal steps can be evaluated in closed forms. Our algorithm also directly provides values of the partial derivatives of the learned function indicating the learned sparsity.

### 4.4.1 Finite dimensional formulation

To be able to develop a practical algorithm we first need to reformulate the variational optimisation problem (4.10) into its finite dimensional equivalent. For this we introduce the following objects: the $n$-long vector $\boldsymbol{\alpha} = [\alpha_1, \ldots, \alpha_n]^T$, the $dn$-long vector $\boldsymbol{\beta} = [\beta_{11}, \ldots, \beta_{1n}, \beta_{21} \ldots \beta_{dn}]^T$, the $n \times n$ symmetric PSD kernel matrix $\mathbf{K}$ such that $K_{ij} = k(\mathbf{x}^i, \mathbf{x}^j)$, the $n \times n$ (non-symmetric) kernel derivative matrices $\mathbf{D}^a$ and $\widetilde{\mathbf{D}}^a$, $a \in \mathbb{N}_d$ such that $D_{ij}^a = [\partial_a k_{\mathbf{x}^i}](\mathbf{x}^j) = \partial_a k_{\mathbf{x}^j}(\mathbf{x}^i) = \widetilde{D}_{ji}^a$, the $n \times n$ (non-symmetric) kernel 2nd derivative matrices $\mathbf{L}^{ab}$, $a, b \in \mathbb{N}_d$ such that $\mathbf{L}_{ij}^{ab} = \frac{\partial^2}{\partial x_a^i \partial x_b^j} k(\mathbf{x}^i, \mathbf{x}^j) = \frac{\partial}{\partial x_b^j}[\partial_a k_{\mathbf{x}^i}](\mathbf{x}^j) = \mathbf{L}_{ji}^{ba}$. Further, we need the following concatenations:

$$\mathbf{D} = \begin{bmatrix} \mathbf{D}^1 \\ \ldots \\ \mathbf{D}^d \end{bmatrix} \quad \mathbf{L}^a = [\mathbf{L}^{a1} \ldots \mathbf{L}^{ad}] \quad \mathbf{L} = \begin{bmatrix} \mathbf{L}^1 \\ \ldots \\ \mathbf{L}^d \end{bmatrix}$$

and specifically for the groups $g$ in $\mathrm{R}^{GL}$ the partitions

$$\check{\mathbf{D}}^g = \begin{bmatrix} \mathbf{D}^{g_1} \\ \ldots \\ \mathbf{D}^{g_{p_g}} \end{bmatrix} \quad \check{\mathbf{L}}^g = \begin{bmatrix} \mathbf{L}^{g_1} \\ \ldots \\ \mathbf{L}^{g_{p_g}} \end{bmatrix},$$

where the subscripts $g_i$ are the corresponding indexes of the input dimensions.

**Proposition 2.** *The variational problem* (4.10) *is equivalent to the finite dimen-*

*sional problem*

$$\underset{\boldsymbol{\alpha}, \boldsymbol{\beta}}{\arg\min}\, \mathrm{J}1(\boldsymbol{\alpha}, \boldsymbol{\beta}) + \tau \mathrm{J}2(\boldsymbol{\alpha}, \boldsymbol{\beta}) + \nu \mathrm{J}3(\boldsymbol{\alpha}, \boldsymbol{\beta}), \qquad (4.12)$$

*where*

$$\mathrm{J}1(\boldsymbol{\alpha}, \boldsymbol{\beta}) = \frac{1}{n} \left\| \mathbf{y} - \mathbf{K}\boldsymbol{\alpha} - \mathbf{D}^T \boldsymbol{\beta} \right\|_2^2$$

$$\mathrm{R}^L : \mathrm{J}2(\boldsymbol{\alpha}, \boldsymbol{\beta}) = \frac{1}{\sqrt{n}} \sum_a^d \left\| \mathbf{D}^a \boldsymbol{\alpha} + \mathbf{L}^a \boldsymbol{\beta} \right\|_2$$

$$\mathrm{R}^{GL} : \mathrm{J}2(\boldsymbol{\alpha}, \boldsymbol{\beta}) = \frac{1}{\sqrt{n}} \sum_g^G p_g \left\| \mathbf{\check{D}}^g \boldsymbol{\alpha} + \mathbf{\check{L}}^g \boldsymbol{\beta} \right\|_2$$

$$\mathrm{R}^{EN} : \mathrm{J}2(\boldsymbol{\alpha}, \boldsymbol{\beta}) = \frac{\mu}{\sqrt{n}} \sum_a^d \left\| \mathbf{D}^a \boldsymbol{\alpha} + \mathbf{L}^a \boldsymbol{\beta} \right\|_2 + \frac{1-\mu}{n} \sum_a^d \left\| \mathbf{D}^a \boldsymbol{\alpha} + \mathbf{L}^a \boldsymbol{\beta} \right\|_2^2$$

$$\mathrm{J}3(\boldsymbol{\alpha}, \boldsymbol{\beta}) = \boldsymbol{\alpha}^T \mathbf{K} \boldsymbol{\alpha} + 2\boldsymbol{\alpha}^T \mathbf{D}^T \boldsymbol{\beta} + \boldsymbol{\beta}^T \mathbf{L} \boldsymbol{\beta}$$

The proof (available in the appendix) is based on the finite dimensional representation (4.11) of the minimising function, and the kernel and derivative reproducing properties stated in section 4.3.

The problem reformulation (4.12) is instructive in terms of observing the roles of the kernel and the derivative matrices and is reminiscent of the classical finite dimensional reformulation of Hilbert-norm regularised least squares. However, for the development of our algorithm we derive a more convenient equivalent form.

**Proposition 3.** *The variational problem* (4.10) *is equivalent to the finite dimensional problem*

$$\underset{\boldsymbol{\omega}}{\arg\min}\, \frac{1}{n} \left\| \mathbf{y} - \mathbf{F}\boldsymbol{\omega} \right\|_2^2 + \tau \mathrm{J}(\boldsymbol{\omega}) + \nu \boldsymbol{\omega}^T \mathbf{Q} \boldsymbol{\omega}, \qquad (4.13)$$

*where*

$$R^L : J(\boldsymbol{\omega}) = \frac{1}{\sqrt{n}} \sum_a^d \|\mathbf{Z}^a \boldsymbol{\omega}\|_2$$

$$R^{GL} : J(\boldsymbol{\omega}) = \frac{1}{\sqrt{n}} \sum_g^G p_g \left\|\mathbf{\breve{Z}}^g \boldsymbol{\omega}\right\|_2$$

$$R^{EN} : J(\boldsymbol{\omega}) = \frac{\mu}{\sqrt{n}} \sum_a^d \|\mathbf{Z}^a \boldsymbol{\omega}\|_2 + \frac{1-\mu}{n} \sum_a^d \|\mathbf{Z}^a \boldsymbol{\omega}\|_2^2 \quad,$$

*with*

$$\boldsymbol{\omega} = \begin{bmatrix} \boldsymbol{\alpha} \\ \boldsymbol{\beta} \end{bmatrix} \quad \begin{matrix} \mathbf{F} = [\mathbf{KD}^T] \\ \mathbf{Z}^a = [\mathbf{D}^a \mathbf{L}^a] \\ \mathbf{\breve{Z}}^g = [\mathbf{\breve{D}}^g \mathbf{\breve{L}}^g] \end{matrix} \quad \mathbf{Q} = \begin{bmatrix} \mathbf{K} & \mathbf{0} \\ 2\mathbf{D} & \mathbf{L} \end{bmatrix}$$

The proof is trivial using (4.12) as an intermediate step.

### 4.4.2 Development of generic algorithm

Problem (4.13) is convex though its middle part $J(\boldsymbol{\omega})$ is non-differentiable for all three discussed regularizers. Indeed, it is the singularities of the norms at zero points that yield the sparse solutions. A popular approach for solving convex non-differentiable problems is the proximal gradient descent (Parikh and Boyd 2013). At every step it requires evaluating the proximal operator defined for any function $f : \mathbb{R}^m \to \mathbb{R}^m$ and any vector $\mathbf{v} \in \mathbb{R}^m$ as

$$prox_f(\mathbf{v}) = \arg\min_{\mathbf{x}} f(\mathbf{x}) + \frac{1}{2} \|\mathbf{x} - \mathbf{v}\|_2^2 \quad. \tag{4.14}$$

However, proximal operators for the functions $J$ in (4.13) do not have closed forms or fast methods for solving which makes the proximal gradient descent algorithm difficult to use.

We therefore propose to introduce a linearizing change of variables $\mathbf{Z}^a \boldsymbol{\omega} = \boldsymbol{\varphi}_a$ and cast the problem in a form amenable for the ADMM method (Boyd 2010)

$$\min \ \Omega(\boldsymbol{\omega}) + \tau \Upsilon(\boldsymbol{\varphi}), \quad \text{s.t. } \mathbf{Z}\boldsymbol{\omega} - \boldsymbol{\varphi} = 0 \quad. \tag{4.15}$$

In the above

$$\boldsymbol{\varphi} = \begin{bmatrix} \boldsymbol{\varphi}_1 \\ \dots \\ \boldsymbol{\varphi}_d \end{bmatrix} \qquad \mathbf{Z} = \begin{bmatrix} \mathbf{Z}^1 \\ \dots \\ \mathbf{Z}^d \end{bmatrix} ,$$

(or concatenation of the grouped versions for $\mathbf{R}^{GL}$), $\Omega : \mathbb{R}^{n+nd} \to \mathbb{R}$ is the convex differentiable function

$$\Omega(\boldsymbol{\omega}) = \frac{1}{n} \|\mathbf{y} - \mathbf{F}\boldsymbol{\omega}\|_2^2 + \nu \boldsymbol{\omega}^T \mathbf{Q} \boldsymbol{\omega} ,$$

and $\Upsilon : \mathbb{R}^{nd} \to \mathbb{R}$ is the convex non-differentiable function corresponding to each regularizer such that $\Upsilon(\boldsymbol{\varphi}) = J(\boldsymbol{\omega})$ for every $\mathbf{Z}\boldsymbol{\omega} = \boldsymbol{\varphi}$.

At each iteration the ADMM algorithm consists of the following three update steps (the standard approach of augmented Lagrangian with $\boldsymbol{\lambda}$ as the scaled dual variable and $\kappa$ as the step size):

$$S1 : \boldsymbol{\omega}^{(k+1)} = \underset{\boldsymbol{\omega}}{\arg\min} \ \Omega(\boldsymbol{\omega}) + \frac{\kappa}{2} \left\| \mathbf{Z}\boldsymbol{\omega} - \boldsymbol{\varphi}^{(k)} + \boldsymbol{\lambda}^{(k)} \right\|_2^2$$

$$S2 : \boldsymbol{\varphi}^{(k+1)} = \underset{\boldsymbol{\varphi}}{\arg\min} \ \tau\Upsilon(\boldsymbol{\varphi}) + \frac{\kappa}{2} \left\| \mathbf{Z}\boldsymbol{\omega}^{(k+1)} - \boldsymbol{\varphi} + \boldsymbol{\lambda}^{(k)} \right\|_2^2$$

$$S3 : \boldsymbol{\lambda}^{(k+1)} = \boldsymbol{\lambda}^{(k)} + \mathbf{Z}\boldsymbol{\omega}^{(k+1)} - \boldsymbol{\varphi}^{(k+1)}$$

The first step $S1$ is a convex quadratic problem with a closed form solution

$$S1 : (\nu\mathbf{Q} + \nu\mathbf{Q}^T + 2n^{-1}\mathbf{F}^T\mathbf{F} + \kappa\mathbf{Z}^T\mathbf{Z})\boldsymbol{\omega}^{(k+1)} = 2n^{-1}\mathbf{F}^T\mathbf{y} + \kappa\mathbf{Z}^T(\boldsymbol{\varphi}^{(k)} - \boldsymbol{\lambda}^{(k)})$$

By comparing with (4.14) we observe that the second step $S2$ is a proximal update. The advantage of our problem reformulation and our algorithm is that this has a closed form for all the three discussed regularizers.

**Proposition 4.** *The proximal problem in step $S2$ is decomposable by the $d$ partitions of vector $\boldsymbol{\varphi}$ (or G partition in case of the group structure) and the minimising*

*solution is*

$$\text{R}^L: \boldsymbol{\varphi}_a^{(k+1)} = (\mathbf{Z}^a \boldsymbol{\omega}^{(k+1)} + \boldsymbol{\lambda}_a^{(k)})\left(1 - \frac{\tau}{\kappa \sqrt{n} \left\| \mathbf{Z}^a \boldsymbol{\omega}^{(k+1)} + \boldsymbol{\lambda}_a^{(k)} \right\|_2}\right)_+$$

$$\text{R}^{GL}: \boldsymbol{\varphi}_g^{(k+1)} = (\check{\mathbf{Z}}^g \boldsymbol{\omega}^{(k+1)} + \check{\boldsymbol{\lambda}}_g^{(k)})\left(1 - \frac{\tau p_g}{\kappa \sqrt{n} \left\| \mathbf{Z}^g \boldsymbol{\omega}^{(k+1)} + \boldsymbol{\lambda}_g^{(k)} \right\|_2}\right)_+$$

$$\text{R}^{EN}: \boldsymbol{\varphi}_a^{(k+1)} = \frac{\mathbf{Z}^a \boldsymbol{\omega}^{(k+1)} + \boldsymbol{\lambda}_a^{(k)}}{2\tau(1-\mu)/(\kappa n)+1}\left(1 - \frac{\tau \mu}{\kappa \sqrt{n} \left\| \mathbf{Z}^a \boldsymbol{\omega}^{(k+1)} + \boldsymbol{\lambda}_a^{(k)} \right\|_2}\right)_+ .$$

*Here $(v)_+ = \min(0, v)$ is the thresholding operator.*

The decomposability comes from the additive structure of $\Upsilon$. The derivation follows similar techniques as used for classical $\ell_1$ and $\ell_2$ proximals.[1]

### 4.4.3 Practical implementation

In practice, the $\mathbf{Q}, \mathbf{F}$ and $\mathbf{Z}$ matrices are precomputed in a preprocessing step and passed onto the algorithm as inputs. The matrices are directly computable using the kernel function $k$ and its first and second order derivatives evaluated at the training points (following the matrix definitions introduced in section 4.4.1).

The algorithm converges to a global minimum by the standard properties of ADMM. In our implementation (available at https://bitbucket.org/dmmlgeneva/nvsd_uai2018/) we follow a simple updating rule (Boyd 2010, sec. 3.4.1) for the step size $\kappa$. We use inexact minimization for the most expensive step $S1$, gradually increasing the number of steepest descent steps, each with complexity $\mathcal{O}\big((nd)^2\big)$.

Furthermore, we use $S2$ to get the values of the training sample partial-derivative norms defined in equation (4.7) as $\left\| \partial_a f^{(k)} \right\|_{2_n} = \left\| \boldsymbol{\varphi}_a^{(k)} \right\|_2 / \sqrt{n}$. The sparsity pattern is obtained by examining for which of the dimensions $a \in \mathbb{N}_d$ the norm is zero $\left\| \partial_a f^{(k)} \right\|_{2_n} = 0$.

---

[1] For $\text{R}^{EN}$ it is more practical to add the quadratic term into $\Omega(\boldsymbol{\omega})$ in $S1$ and use the corresponding scaled version of the $\text{R}^L$ proximal in $S2$.

## 4.5 Empirical evaluation

We conducted a set of synthetic and real-data experiments to document the efficacy of our structured-sparsity methods and the new algorithm under controlled and more realistic conditions. We compare our methods NVSD(L), NVSD(GL) and NVSD(EN) in terms of their predictive accuracy and their selection ability to the simple (non-sparse) kernel regularised least squares (Krls) (see section 1.3.1), to the sparse additive model (SpAM) of Ravikumar et al. (2007), to the non-linear crosscovariance based method using the Hilbert Schmidt independence criterion in a lasso-like manner (HSIC) of Yamada et al. (2014), and to the derivative-based lasso-like method (Denovas) of Rosasco et al. (2013).[2] We compared also to simple mean and linear sparse and non-sparse models. All of these performed considerably worse than the non-linear models and therefore are not listed in the summary results. For all the sparse kernel methods we consider a two-step debiasing procedure based on variable selection via the base algorithm followed by a simple kernel regularised least squares on the selected variables.[3]

### 4.5.1 Synthetic experiments

We motivate each synthetic experiment by a realistic story-line and explain the data generating process here below. In all the synthetic experiments we fix the input dimension to $d = 18$ with only 6 input variables $\{1, 2, 3, 7, 8, 9\}$ relevant for the model and the other 12 irrelevant.

**E1** In the first experiment we focus on the NVSD(GL) which assumes the input variables can be grouped a priori by some domain knowledge (e.g. each group describes a *type* of input data such as a different biological process) and the groups are expected to be completely in or out of the model. The input variables are generated independently from a standard normal distribution and they are grouped by three into 6 groups. The output is

---

[2] For HISC and Denovas we used the author's code, for SpAM the R implementation of Zhao et al. 2014. For all algorithms we kept the default settings.

[3] This is native to Denovas and necessary for HSIC which otherwise does not produce a predictive model.

generated from the 1st and the 3rd group as

$$y = \sum_{i=1}^{3} \sum_{j=i}^{3} \sum_{k=j}^{3} x_i x_j x_k + \sum_{q=7}^{9} \sum_{r=q}^{9} \sum_{s=r}^{9} x_q x_r x_s + \epsilon \ ,$$

with $\epsilon \sim N(0, 0.01)$. For learning we fix the kernel to 3rd order polynomial.

**E2** In the second experiment we do not assume any a priori grouping of the variables. Instead some of the variables are strongly correlated (perhaps relating to a single phenomenon), a case for NVSD(EN). The input variables are generated similarly as in E1 but with the pairs $\{1, 7\}, \{2, 8\}$ and $\{3, 9\}$ strongly correlated (Pearson's population correlation coefficient 0.95). The remaining (irrelevant) input variables are also pair-wise correlated and the output is generated as

$$y = \sum_{i,j,k=1}^{3} x_i x_j x_k + \sum_{q,r,s=7}^{9} x_q x_r x_s + \epsilon \ ,$$

with $\epsilon \sim N(0, 0.01)$. For learning we fix the kernel to 3rd order polynomial.

**E3** In the third experiment we assume the inputs are noisy measurements of some true phenomenon (e.g. repeated measurements, measurements from multiple laboratories) for which there is no reason to prefer one over the other in the model. We first generate the true data $z_i \sim N(0, 1), i = 1, \ldots, 6$ and use these to generate the outputs as

$$y = 10(z_1^2 + z_3^2)e^{-2(z_1^2 + z_3^2)} + \epsilon \ ,$$

with $\epsilon \sim N(0, 0.01)$. We then generate the noisy measurements that will be used as inputs for the learning: for each $z_i$ we create three noisy measurements $x_{ij} = z_i + N(0, 0.1), j = 1, 2, 3$ (a group for the NVSD(GL) method); the input vector is the concatenation of all $x_{ij}$ so that from the 18 long concatenated input vector $\mathbf{x}$ again only the set $\{1, 2, 3, 7, 8, 9\}$ of the dimensions is relevant for predicting the output $y$. For learning we fix the kernel to Gaussian with width $\sigma = 4$.

**Remark 3.** *In all the synthetic experiments we use the same experimental proto-col. We split the data into train sets varying the size in $n = \{30, 50, 70, 90, 110\}$, a validation set of length 1000, and a test set of length 1000. We train the models over the train sets and use the validation set to select the regularization hyper-parameters (and therefore the models) based on the minimal validation MSE. We use dense grids of 50 points for the $\tau$ search (automatically established by the al-gorithm) and 5 points grid for $\mu \in \{0.1, \ldots, 0.9\}$. Complete settings (also for the baseline methods) are detailed in the replication files publicly available at* https://bitbucket.org/dmmlgeneva/nvsd_uai2018/.

We report the average results across 50 independent replications of the exper-iments in table 4.5.1. We measure the prediction accuracy by the root mean squared error (RMSE) over the test sets and the selection accuracy by the Tani-moto distance between the true sparsity and the learned sparsity patterns (sec-tion 4.4.3).

Our structured-sparsity methods clearly outperform all the non-structured sparse learning methods achieving better prediction accuracy based on more precise variable selection, typically with statistically significant differences. Also, the prediction and selection accuracy generally increases (errors decrease) for larger training sample sizes suggesting our methods are well-behaved in terms of the standard statistical learning paradigms. In the E3 experiment, NVSD(GL) performs the best having the benefit of the prior knowledge of the variable groupings. Remarkably, NVSD(EN) follows closely after even without such prior information, learning about the groups of correlated variables from the data when building the model.

Krls can only learn full models and therefore performs rather poorly on these by-construction sparse problems. From the other three baselines, HSIC typi-cally achieves the second best results (after our NVSD methods). SpAM is not particularly suitable for the non-additive structures of our experiments. Fi-nally, in all the experiments our NVSD(L) outperforms Denovas though they share the same lasso-like problem formulation. We attribute this to our new algorithm developed in section 4.4 which, unlike Denovas, does not rely on approximations of the proximal operators.

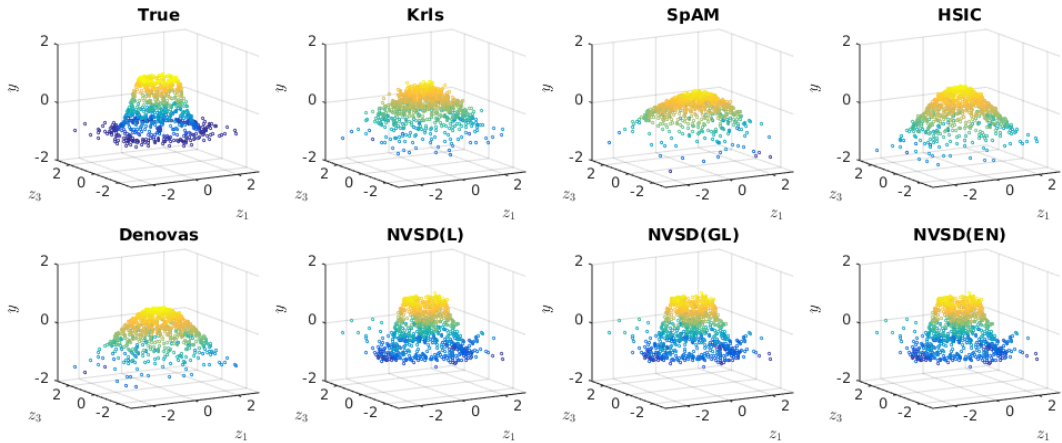

Figure 4.5.1: Predictions for the E3 experiment over the test data. We picked an example for the model trained with 110 instances (the 17th replication) which illustrates well the advantage our NVSD methods have over the baselines in capturing the True complex non-linear structure.

### 4.5.2 Real-data experiments

For the real-data experiments we used a collection of regression datasets from UCI[4] and LIACC[5] repositories listed in table 4.5.2.

We report the average results across 50 replications of the experiments in tables 4.5.3 and 4.5.4. We use RMSE over the test data for measuring the prediction accuracy. For the real datasets we do not know the ground-truth sparsity patterns. Instead of measuring the selection error we therefore count the number of input variables selected by each method. Krls has no selection ability, its support size is hence equal to the total number of input variables in each problem.

**Remark 4.** *We followed similar experimental protocol as for the synthetic experiments. We fixed the training sample size for all experiments to 100 instances and used 200-1000 instances for the validation and test sets (depending on the total number of available observations). We pre-processed the data by normalizing the inputs and centering the outputs. For all the experiments we used a Gaussian kernel with the width set to the median distance calculated over the nearest 20 neigh-*

---

[4]http://archive.ics.uci.edu/ml

[5]http://www.dcc.fc.up.pt/$\sim$ltorgo/Regression/DataSets.html

*bours, and the 3rd order polynomial kernel. With the exception of the EN dataset, the Gaussian kernel yielded better results and was therefore kept for the final evaluation. Full details of the settings can be found in the replication files publicly available at* https://bitbucket.org/dmmlgeneva/nvsd_uai2018/.

Results in table 4.5.3 are for the original data for which we have no prior knowledge about possible variable groupings. Therefore we only use the non-structured methods and our NVSD(EN) that do not rely on any such prior information.

Our NVSD methods learned sparse non-linear models achieving better or comparable results than the baselines in 4 out of the 5 experiments (BH, CP, EN, EL). For CC reducing the number of input dimensions does not seem to bring any advantages and the methods tend to learn full models. For several experiments SpAM finished with errors and therefore the results in the table are missing.

To explore the performance and benefits of NVSD(GL) method we had to construct variable groups that could potentially help the model learning. We adopted two strategies:

1. For CP and EL datasets we constructed the groups based on the NVSD(EN) results. For CP we grouped together the 5 most often selected variables across the 50 replications of the experiment and created 3 other groups from the remaining variables. For EL we created five groups by 3-4 elements putting together variables with similar frequencies of occurrence in the support of the learned NVSD(EN) models over the 50 replications.

2. For AI, CC, and KN datasets we doubled the original input data dimensions by complementing the input data by a copy of each input variable with permuted instance order. We then constructed two groups, the first over the original data, the second over the permuted copy.

Table 4.5.4 confirms that our NVSD(GL) is able to use the grouping information based on prior knowledge to select better, more relevant subset of variables than the non-structured baselines. Thanks to this it achieves significantly better prediction accuracy in all the experiments.

## 4.6 Conclusions and future work

In this work we addressed the problem of variable selection in non-linear regression problems. We followed up from the work of Rosasco et al. 2013 arguing for the use of partial derivatives as an indication of the pertinence of an input variable for the model. Extending the existing work, we proposed two new derivative-based regularizers for learning with structured sparsity in non-linear regression similar in spirit to the linear elastic net and group lasso.

After posing the problems into the framework of RKHS learning, we designed a new NVSD algorithm for solving these. Unlike the previously proposed Denovas our new algorithm does not rely on proximal approximations. This is most likely the main reason why our NVSD(L) method achieved systematically better predictive performance than Denovas on a broad set of experiments. We also empirically demonstrated the advantages our structured sparsity methods NVSD(GL) and NVSD(EN) bring for learning tasks with a priori known group structures or correlation in the inputs.

These promising results point to questions requiring further attention:

Our NVSD algorithm achieves better results in terms of prediction accuracy than Denovas, however, at the cost of longer training times. Its $\mathcal{O}\left((nd)^2\right)$ complexity is not favourable for scaling in neither instances nor dimensions. Exploring avenues for speeding up, possibly along the lines of random features construction, is certainly an important next step in making the algorithm operational for more practical real-life problems.

The method is based on the partial-derivative arguments and therefore assumes the functions (and therefore the kernels) are at least 2nd order differentiable (and square-integrable). We use here the polynomial and Gaussian kernel as the most commonly used examples. What other properties of the kernels are necessary to ensure good performance and how the methods could be extended to other, more complex kernels are relevant questions.

The full problem formulation (e.g. equation (4.10) in proposition 1) combines the sparse regularizers with the function Hilbert-norm. This combination has been proposed in Rosasco et al. (2013) to ensure that the regularization part of the problem is strongly convex and the problem is well-posed in terms of the generalization properties.

However, interactions of the Hilbert norm with the sparsity inducing regularizers of section 4.2 and the effects on the learning and selection properties are not yet fully clear. Empirically (from Rosasco et al. 2013 and our own experiments) the models are often little sensitive to variations in $\nu$[6].

In addition, the $R^{EN}$ regularizer is already strongly convex even without the Hilbert norm. To what degree combining it with the Hilbert norm is necessary to guarantee good generalization for outside the training needs to be further investigated. So does its behaviour and the possible improvements it can bring when learning from inputs with non-linear dependencies. In view of the above considerations, our work is posing the motivations, foundations and principles for further studies on partial derivative-based regularizations.

---

[6]We fix it based on a small subset of replications instead of including it into the full hyper-parameter search.

Table 4.5.1: Results of synthetic experiments

| | Train size | 30 | 50 | 70 | 90 | 110 |
|---|---|---|---|---|---|---|
| E1 | **RMSE** | | | | | |
| | Krls | 12.79 | 11.66 | 10.99 | 10.43 | 9.80 |
| | SpAM | 11.41 | 9.47 | 8.66 | 8.22 | 7.75 |
| | HSIC | 11.37 | 10.00 | 8.58 | 7.28 | 5.68 |
| | Denovas | 11.66 | 10.87 | 12.37 | 13.28 | 11.78 |
| | NVSD(L) | 11.55 | 10.22 | 9.36 | 7.90 | 7.13 |
| | NVSD(GL) | **9.92** | **7.89** | **6.34** | **1.94** | **2.41** |
| | **Selection error** | | | | | |
| | Krls | 0.67 | 0.67 | 0.67 | 0.67 | 0.67 |
| | SpAM | 0.54 | 0.56 | 0.59 | 0.57 | 0.58 |
| | HSIC | 0.50 | 0.48 | 0.42 | 0.35 | 0.32 |
| | Denovas | 0.49 | 0.50 | 0.53 | 0.67 | 0.73 |
| | NVSD(L) | 0.49 | 0.47 | 0.48 | 0.39 | 0.32 |
| | NVSD(GL) | **0.28** | **0.24** | **0.22** | **0.05** | **0.11** |
| E2 | **RMSE** | | | | | |
| | Krls | 27.69 | 24.83 | 22.53 | 19.14 | 18.04 |
| | SpAM | 31.24 | 29.21 | 29.25 | 27.11 | 26.03 |
| | HSIC | 21.74 | 15.50 | 12.02 | 9.42 | 7.67 |
| | Denovas | 24.23 | 34.33 | 17.51 | 8.89 | 11.20 |
| | NVSD(L) | 21.24 | 16.59 | 11.79 | 8.61 | 7.35 |
| | NVSD(EN) | **17.53** | **10.05** | **5.67** | **4.29** | **3.29** |
| | **Selection error** | | | | | |
| | Krls | 0.67 | 0.67 | 0.67 | 0.67 | 0.67 |
| | SpAM | 0.57 | 0.55 | 0.49 | 0.52 | 0.46 |
| | HSIC | 0.52 | 0.42 | 0.42 | 0.35 | 0.32 |
| | Denovas | 0.46 | 0.54 | 0.40 | 0.30 | 0.26 |
| | NVSD(L) | 0.46 | 0.43 | 0.36 | 0.31 | 0.29 |
| | NVSD(EN) | **0.35** | **0.20** | **0.14** | **0.09** | **0.08** |
| E3 | **RMSE** | | | | | |
| | Krls | 0.65 | 0.55 | 0.54 | 0.53 | 0.50 |
| | SpAM | 0.51 | 0.49 | 0.47 | 0.47 | 0.46 |
| | HSIC | 0.52 | 0.47 | 0.45 | 0.44 | 0.43 |
| | Denovas | 0.55 | 0.51 | 0.50 | 0.51 | 0.50 |
| | NVSD(L) | 0.51 | 0.44 | 0.44 | 0.41 | 0.34 |
| | NVSD(GL) | 0.51 | **0.41** | **0.39** | **0.33** | 0.31 |
| | NVSD(EN) | **0.50** | 0.43 | 0.42 | **0.36** | **0.30** |
| | **Selection error** | | | | | |
| | Krls | 0.67 | 0.67 | 0.67 | 0.67 | 0.67 |
| | SpAM | 0.65 | 0.61 | 0.60 | 0.58 | 0.59 |
| | HSIC | 0.59 | 0.51 | 0.53 | 0.47 | 0.44 |
| | Denovas | 0.49 | 0.45 | 0.47 | 0.45 | 0.41 |
| | NVSD(L) | 0.33 | 0.30 | 0.40 | 0.34 | 0.23 |
| | NVSD(GL) | **0.26** | **0.20** | **0.24** | **0.15** | **0.14** |
| | NVSD(EN) | 0.30 | 0.33 | 0.35 | **0.25** | 0.16 |

Best results in bold; underlined when structured-sparsity methods significantly better than all other methods using Wilcoxon signed-rank test at 5% significance level.

Table 4.5.2: Real datasets desription

| Code | Name | Inputs | Test Size | Source |
|------|------|--------|-----------|--------|
| AI | Airfoil Self Noise | 5 | 700 | UCI |
| BH | Boston Housing | 10 | 200 | UCI |
| CC | Concrete Compressive | 8 | 450 | UCI |
| EN | Energy Efficiency | 8 | 300 | UCI |
| CP | Computer Activity | 21 | 1000 | LIACC |
| EL | F16 Elevators | 17 | 1000 | LIACC |
| KN | Kynematics | 8 | 1000 | LIACC |

Table 4.5.3: Results of real-data experiments

| | Experiment | BH | CP | CC | EN | EL |
|------|-----------|------|-------|-------|------|------|
| **RMSE** | Krls | 4.00 | 12.27 | 8.70 | 1.83 | 5.10 |
| | SpAM | 4.33 | ~ | 12.70 | ~ | ~ |
| | HSIC | 4.02 | 9.39 | 8.73 | **1.19** | 9.07 |
| | Denovas | 4.02 | 9.21 | 12.07 | 3.02 | 6.01 |
| | NVSD(L) | 3.96 | 8.43 | **8.67** | 1.50 | <u>4.81</u> |
| | NVSD(EN) | <u>**3.93**</u> | **7.88** | 8.70 | 1.20 | <u>**4.67**</u> |
| **Support size** | Krls | 10.00 | 21.00 | 8.00 | 8.00 | 17.00 |
| | SpAM | 9.00 | ~ | 2.82 | ~ | ~ |
| | HSIC | 6.12 | 8.26 | 5.88 | 5.08 | 0.00 |
| | Denovas | 8.80 | 4.76 | 4.38 | 4.96 | 10.52 |
| | NVSD(L) | 8.20 | 3.78 | 7.36 | 7.26 | 14.06 |
| | NVSD(EN) | 8.06 | 4.58 | 7.98 | 6.66 | 13.00 |

Best results in bold; underlined when NVSD methods significantly better than all the baselines using Wilcoxon signed-rank test at 5% significance level. For several experiments SpAM finished with errors.

Table 4.5.4: Results of real-data experiments with groups

| | Experiment | AI | CP | CC | KN | EL |
|---|---|---|---|---|---|---|
| **RMSE** | Krls | 5.08 | 12.27 | 10.34 | 2.07 | 5.10 |
| | SpAM | ~ | ~ | 13.31 | 2.20 | ~ |
| | HSIC | 4.64 | 9.39 | 9.29 | 2.05 | 9.07 |
| | Denovas | 5.12 | 9.21 | 11.49 | 2.10 | 6.01 |
| | NVSD(L) | <u>4.45</u> | 8.43 | 9.58 | 2.03 | <u>4.81</u> |
| | NVSD(GL) | **<u>4.16</u>** | **<u>7.43</u>** | **<u>8.79</u>** | **<u>1.96</u>** | **<u>4.76</u>** |
| **Support size** | Krls | 10.00 | 21.00 | 16.00 | 16.00 | 17.00 |
| | SpAM | ~ | ~ | 2.60 | 11.32 | ~ |
| | HSIC | 5.08 | 8.26 | 6.16 | 11.82 | 0.00 |
| | Denovas | 5.94 | 4.76 | 6.96 | 9.72 | 10.52 |
| | NVSD(L) | 4.76 | 3.78 | 8.16 | 13.58 | 14.06 |
| | NVSD(GL) | 5.00 | 5.84 | 8.00 | 11.84 | 13.82 |

Best results in bold; underlined when NVSD methods significantly better than all the baselines using Wilcoxon signed-rank test at 5% significance level. For several experiments SpAM finished with errors.

# Appendix

## 4.A    Proofs of propositions from the main text

*Proof of Proposition 1.* We may decompose any function $f \in \mathcal{H}_K$ as $f = f_\| + f_\perp$, where $f_\|$ lies in the span of the kernel sections $k_{\mathbf{x}^i}$ and its partial derivatives $[\partial_a k_{\mathbf{x}^i}]$ centred at the $n$ training points, and $f_\perp$ lies in its orthogonal complement.

The 1st term $\widehat{L}(f)$ depends on the function $f$ only through its evaluations at the training points $f(\mathbf{x}^i), i \in \mathbb{N}_n$. For each training point $\mathbf{x}^i$ we have

$$f(\mathbf{x}^i) = \langle f, k_{\mathbf{x}^i} \rangle_{\mathcal{H}_K} = \langle f_\| + f_\perp, k_{\mathbf{x}^i} \rangle_{\mathcal{H}_K} = \langle f_\|, k_{\mathbf{x}^i} \rangle_{\mathcal{H}_K} \ ,$$

where the last equality is the result of the orthogonality of the complement $\langle f_\perp, k_{\mathbf{x}^i} \rangle_{\mathcal{H}_K} = 0$. By this the term $\widehat{L}(f)$ is independent of $f_\perp$.

The 2nd term $\widehat{R}(f)$ depends on the function $f$ only through the evaluations of its partial derivatives at the training points $\partial_a f(\mathbf{x}^i), i \in \mathbb{N}_i, a \in \mathbb{N}_d$. For each training point $\mathbf{x}^i$ and dimension $a$ we have

$$\partial_a f(\mathbf{x}^i) = \langle f, [\partial_a k_{\mathbf{x}^i}] \rangle_{\mathcal{H}_K} = \langle f_\|, [\partial_a k_{\mathbf{x}^i}] \rangle_{\mathcal{H}_K} \ ,$$

by the orthogonality of the complement $\langle f_\perp, [\partial_a k_{\mathbf{x}^i}] \rangle_{\mathcal{H}_K} = 0$. By this the term $\widehat{R}(f)$ is independent of $f_\perp$ for the empirical versions of all three considered regularizers $R^L, R^{GL}, R^{EN}$. For the 3rd term we have $\|f\|^2_{\mathcal{H}_K} = \|f_\| + f_\perp\|^2_{\mathcal{H}_K} = \|f_\|\|^2_{\mathcal{H}_K} + \|f_\perp\|^2_{\mathcal{H}_K}$ because $\langle f_\|, f_\perp \rangle_{\mathcal{H}_K} = 0$. Trivially, this is minimised when $f_\perp = 0$. $\qquad\square$

*Proof of Proposition 2.* Using the matrices and vector introduced in section 4.1 and proposition 1 we have

$$f(\mathbf{x}^i) = \sum_{j=1}^n \alpha_j K_{ji} + \sum_{j=1}^n \sum_{a=1}^d \beta_{aj} \check{D}^a_{ij}$$

$$\partial_a f(\mathbf{x}^i) = \sum_{j=1}^{n} \alpha_j \tilde{D}_{ij}^a + \sum_{j=1}^{n} \sum_{c=1}^{d} \beta_{cj} L_{ji}^{ca}$$

For the 1st term $\widehat{L}(f)$ we have

$$\widehat{L}(f) = \sum_{i=1}^{n} \left(y^i - f(\mathbf{x}^i)\right)^2 = \sum_{i=1}^{n} \left(y^i - \sum_{j=1}^{n} \alpha_j K_{ji} - \sum_{j=1}^{n} \sum_{a=1}^{d} \beta_{aj} \tilde{D}_{ij}^a\right)^2$$

$$= \sum_{i=1}^{n} \left((y^i)^2 - 2y^i \sum_{j=1}^{n} \alpha_j K_{ji} - 2y^i \sum_{j=1}^{n} \sum_{a=1}^{d} \beta_{aj} \tilde{D}_{ij}^a + \sum_{j,l}^{n} \alpha_j \alpha_l K_{ji} K_{l,i} + 2 \sum_{j,l}^{n} \sum_{a=1}^{d} \beta_{aj} \alpha_l \tilde{D}_{ij}^a K_{l,i}\right.$$

$$\left. + \sum_{j,l}^{n} \sum_{a,b}^{d} \beta_{aj} \beta_{bl} \tilde{D}_{ij}^a \tilde{D}_{i,l}^b\right)$$

$$= \mathbf{y}^T \mathbf{y} - 2\mathbf{y}^T \mathbf{K}\boldsymbol{\alpha} - 2\sum_{a}^{d} \mathbf{y}^T \widetilde{\mathbf{D}}^a \mathbf{B}_{a,:}^T + \boldsymbol{\alpha}^T \mathbf{K}\mathbf{K}\boldsymbol{\alpha} + 2\sum_{a}^{d} \boldsymbol{\alpha}^T \mathbf{K}\widetilde{\mathbf{D}}^a \mathbf{B}_{a,:}^T + \sum_{a,b}^{d} \mathbf{B}_{a,:} \mathbf{D}^a \widetilde{\mathbf{D}}^b \mathbf{B}_{b,:}^T$$

$$= \mathbf{y}^T \mathbf{y} - 2\mathbf{y}^T \mathbf{K}\boldsymbol{\alpha} - 2\mathbf{y}^T \mathbf{D}^T \boldsymbol{\beta} + \boldsymbol{\alpha}^T \mathbf{K}\mathbf{K}\boldsymbol{\alpha} + 2\boldsymbol{\alpha}^T \mathbf{K}\mathbf{D}^T \boldsymbol{\beta} + \sum_{a,b}^{d} \boldsymbol{\beta}^T \mathbf{D}\mathbf{D}^T \boldsymbol{\beta}$$

$$= \left\| \mathbf{y} - \mathbf{K}\boldsymbol{\alpha} - \mathbf{D}^T \boldsymbol{\beta} \right\|_2^2 \quad,$$

where $\mathbf{B}$ is the $d \times n$ matrix with the $\beta$ coefficients $\boldsymbol{\beta} = \text{vec}(\mathbf{B}^T)$

For the 2nd term we have

$$\widehat{R}^L(f) = \sum_{a=1}^{d} \sqrt{\frac{1}{n} \sum_{i=1}^{n} \left(\partial_a f(\mathbf{x}^i)\right)^2} = \sum_{a=1}^{d} \left[\frac{1}{n} \sum_{i=1}^{n} \left(\sum_{j=1}^{n} \alpha_j \tilde{D}_{ji}^a + \sum_{j=1}^{n} \sum_{c=1}^{d} \beta_{cj} L_{ji}^{ca}\right)^2\right]^{0.5}$$

$$= \sum_{a=1}^{d} \left[\frac{1}{n} \sum_{i=1}^{n} \left(\sum_{j,l}^{n} \alpha_j \alpha_l \tilde{D}_{ji}^a \tilde{D}_{l,i}^a + 2\sum_{j,l}^{n} \sum_{c=1}^{d} \alpha_j \beta_{cl} \tilde{D}_{ji}^a L_{l,i}^{ca} + \sum_{j,l}^{n} \sum_{c,r}^{d} \beta_{cj} \beta_{rl} L_{ji}^{ca} L_{l,i}^{ra}\right)\right]^{0.5}$$

$$= \sum_{a=1}^{d} \frac{1}{\sqrt{n}} \left[\boldsymbol{\alpha}^T \widetilde{\mathbf{D}}^a \mathbf{D}^a \boldsymbol{\alpha} + 2\sum_{c=1}^{d} \boldsymbol{\alpha}^T \widetilde{\mathbf{D}}^a \mathbf{L}^{ac} \mathbf{B}_{c:}^T + \sum_{c,r}^{d} \mathbf{B}_{c:} \mathbf{L}^{ca} \mathbf{L}^{ar} \mathbf{B}_{r:}^T\right]^{0.5}$$

$$= \sum_{a=1}^{d} \frac{1}{\sqrt{n}} \left[\boldsymbol{\alpha}^T \widetilde{\mathbf{D}}^a \mathbf{D}^a \boldsymbol{\alpha} + 2\boldsymbol{\alpha}^T \widetilde{\mathbf{D}}^a \mathbf{L}^a \boldsymbol{\beta} + \boldsymbol{\beta}^T \mathbf{L}^{aT} \mathbf{L}^a \boldsymbol{\beta}\right]^{0.5} = \sum_{a=1}^{d} \frac{1}{\sqrt{n}} \left\| \mathbf{D}^a \boldsymbol{\alpha} + \mathbf{L}^a \boldsymbol{\beta} \right\|_2$$

$\widehat{R}^{GL}(f)$ and $\widehat{R}^{EN}(f)$ follow in analogy.

For the 3rd term we have

$$
\begin{aligned}
\|f\|_{\mathcal{H}_K}^2 &= \left\| \sum_{j=1}^n \alpha_j k_{\mathbf{x}^j} + \sum_{j=1}^n \sum_{a=1}^d \beta_{aj}[\partial_a k_{\mathbf{x}^j}] \right\|_{\mathcal{H}_K}^2 \\
&= \left\langle \sum_{j=1}^n \alpha_j k_{\mathbf{x}^j}, \sum_{i=1}^n \alpha_i k_{\mathbf{x}^i} \right\rangle_{\mathcal{H}_K} + 2 \left\langle \sum_{j=1}^n \alpha_j k_{\mathbf{x}^j}, \sum_{i=1}^n \sum_{a=1}^d \beta_{ai}[\partial_a k_{\mathbf{x}^i}] \right\rangle_{\mathcal{H}_K} \\
&\quad + \left\langle \sum_{j=1}^n \sum_{a=1}^d \beta_{aj}[\partial_a k_{\mathbf{x}^j}], \sum_{i=1}^n \sum_{c=1}^d \beta_{ci}[\partial_c k_{\mathbf{x}^i}] \right\rangle_{\mathcal{H}_K} \\
&= \boldsymbol{\alpha}^T \mathbf{K} \boldsymbol{\alpha} + 2 \sum_{ij}^n \sum_a^d \alpha_j \beta_{ai} \, \partial_a k_{\mathbf{x}^j}(\mathbf{x}^i) + \sum_{ij}^n \sum_{ac}^d \beta_{aj} \beta_{ci} \frac{\partial^2}{\partial x_a^j \partial x_c^i} k(\mathbf{x}^j, \mathbf{x}^i) \\
&= \boldsymbol{\alpha}^T \mathbf{K} \boldsymbol{\alpha} + 2 \sum_{ij}^n \sum_a^d \alpha_j \beta_{ai} \tilde{D}_{ji}^a + \sum_{ij}^n \sum_{ac}^d \beta_{aj} \beta_{ci} L_{ji}^{ac} \\
&= \boldsymbol{\alpha}^T \mathbf{K} \boldsymbol{\alpha} + 2 \sum_a^d \boldsymbol{\alpha}^T \widetilde{\mathbf{D}}^a \mathbf{B}_{a:}^T + \sum_{ac}^d \mathbf{B}_{:j} \mathbf{L}^{ac} \mathbf{B}_{c:}^T \\
&= \boldsymbol{\alpha}^T \mathbf{K} \boldsymbol{\alpha} + 2 \boldsymbol{\alpha}^T \mathbf{D}^T \boldsymbol{\beta} + \sum_a^d \mathbf{B}_{a:} \mathbf{L}^a \boldsymbol{\beta} \\
&= \boldsymbol{\alpha}^T \mathbf{K} \boldsymbol{\alpha} + 2 \boldsymbol{\alpha}^T \mathbf{D}^T \boldsymbol{\beta} + \boldsymbol{\beta}^T \mathbf{L} \boldsymbol{\beta}
\end{aligned}
$$

$\square$

*Proof of Proposition 4.* The proximal problem in step $S2$ for $R^L$ for a single partition $\boldsymbol{\varphi}_a$ is

$$
R^L : \quad \boldsymbol{\varphi}_a^{(k+1)} = \underset{\boldsymbol{\varphi}_a}{\arg\min} \frac{\tau}{\sqrt{n}} \|\boldsymbol{\varphi}_a\|_2 + \frac{\rho}{2} \left\| \mathbf{Z}^a \boldsymbol{\omega}^{(k+1)} - \boldsymbol{\varphi}_a + \boldsymbol{\lambda}_a^{(k)} \right\|_2^2
$$

This convex problem is non-differentiable at the point $\boldsymbol{\varphi} = \mathbf{0}$. It is, however, sub-differentiable with the optimality condition for the minimizing $\boldsymbol{\varphi}^*$

$$
\mathbf{0} \in \partial \frac{\tau}{\sqrt{n}} \|\boldsymbol{\varphi}_a^*\|_2 - \rho \left( \mathbf{Z}^a \boldsymbol{\omega}^{(k+1)} - \boldsymbol{\varphi}_a + \boldsymbol{\lambda}_a^{(k)} \right) ,
$$

where for any function $f : \mathbb{R}^d \to \mathbb{R}$, $\partial f(\mathbf{x}) \subset \mathbb{R}^d$ is the sub-differential of $f$ at $x$ defined as

$$\partial f(\mathbf{x}) = \{\mathbf{g} \,|\, f(\mathbf{z}) \geq f(\mathbf{x}) + \mathbf{g}^T (\mathbf{z} - \mathbf{x})\} \ .$$

For notational simplicity, in what follows we introduce the variable $\mathbf{v} = \mathbf{Z}^a \boldsymbol{\omega}^{(k+1)} + \boldsymbol{\lambda}_a^{(k)}$, and we drop the sub-/super-scripts of the partitions $a$ and the iterations $k$.

**Part A**  For all points other than $\boldsymbol{\varphi}^* = \mathbf{0}$ the optimality condition reduces to

$$\mathbf{0} = \frac{\tau}{\sqrt{n}} \frac{\boldsymbol{\varphi}^*}{\|\boldsymbol{\varphi}^*\|_2} - \rho (\mathbf{v} - \boldsymbol{\varphi}^*) \ ,$$

From which we get

$$
\begin{aligned}
\left( \frac{\tau}{\rho \sqrt{n} \|\boldsymbol{\varphi}^*\|_2} + 1 \right) \boldsymbol{\varphi}^* &= \mathbf{v} \\
\left( \frac{\tau}{\rho \sqrt{n} \|\boldsymbol{\varphi}^*\|_2} + 1 \right) \|\boldsymbol{\varphi}^*\|_2 &= \|\mathbf{v}\|_2 \\
\|\boldsymbol{\varphi}^*\|_2 &= \|\mathbf{v}\|_2 - \frac{\tau}{\rho \sqrt{n}} \ .
\end{aligned}
$$

We use this result in the optimality condition

$$
\begin{aligned}
\mathbf{0} &= \frac{\tau}{\sqrt{n}} \frac{\boldsymbol{\varphi}^*}{\|\mathbf{v}\|_2 - \frac{\tau}{\rho \sqrt{n}}} - \rho (\mathbf{v} - \boldsymbol{\varphi}^*) \\
\frac{\tau}{\sqrt{n}} \boldsymbol{\varphi}^* &= \rho (\mathbf{v} - \boldsymbol{\varphi}^*)(\|\mathbf{v}\|_2 - \frac{\tau}{\rho \sqrt{n}}) \\
\frac{\tau}{\sqrt{n}} \boldsymbol{\varphi}^* &= (\rho \|\mathbf{v}\|_2 - \frac{\tau}{\sqrt{n}}) \mathbf{v} - \rho \|\mathbf{v}\|_2 \boldsymbol{\varphi}^* + \frac{\tau}{\sqrt{n}} \boldsymbol{\varphi}^* \\
\boldsymbol{\varphi}^* &= \left( 1 - \frac{\tau}{\rho \sqrt{n} \|\mathbf{v}\|_2} \right) \mathbf{v}
\end{aligned}
$$

**Part B**  For the point $\boldsymbol{\varphi}^* = \mathbf{0}$ we have $\partial \|\boldsymbol{\varphi}^*\|_2 = \{\mathbf{g} \,|\, \|\mathbf{g}\|_2 \leq 1\}$ (from the definition of sub-differential and the Cauchy-Schwarz inequality).

From the optimality condition

$$
\begin{aligned}
\mathbf{0} &= \frac{\tau}{\sqrt{n}}\mathbf{g} - \rho\,\mathbf{v} && (\boldsymbol{\varphi}^* = \mathbf{0}) \\
\rho\,\mathbf{v} &= \frac{\tau}{\sqrt{n}}\mathbf{g} \\
\rho\,\|\mathbf{v}\|_2 &= \frac{\tau}{\sqrt{n}}\|\mathbf{g}\|_2 \\
\|\mathbf{v}\|_2 &\leq \frac{\tau}{\rho\sqrt{n}} && (\|\mathbf{g}\|_2 \leq 1)
\end{aligned}
$$

Putting the results from part A and B together we obtain the final result

$$
\boldsymbol{\varphi}^* = \left(1 - \frac{\tau}{\rho\sqrt{n}\,\|\mathbf{v}\|_2}\right)_+ \mathbf{v}
$$

The proofs for $R^{GL}$ and $R^{EN}$ follow similarly. $\qquad\square$

## 4.B   Examples of kernel partial derivatives

We list here the 1st and 2nd order partial derivatives which form the elements of the derivative matrices $\mathbf{D}$ and $\mathbf{L}$ introduced in section 4.1 for some common kernel functions $k$.

**Linear kernel**

Kernel gram matrix

$$
K_{i,j} = k(\mathbf{x}^i, \mathbf{x}^j) = \left\langle \mathbf{x}^i, \mathbf{x}^j \right\rangle
$$

1st order partial-derivative matrix

$$
D_{i,j}^a = \frac{\partial k(\mathbf{s}, \mathbf{x}^j)}{\partial s_a}\big|_{\mathbf{s}=\mathbf{x}^i} = x_a^j
$$

2nd order partial-derivative matrix

$$L_{i,j}^{ab} = \frac{\partial^2 k(\mathbf{s}, \mathbf{r})}{\partial s_a \partial r_b}\big|_{\substack{\mathbf{s}=\mathbf{x}^i \\ \mathbf{r}=\mathbf{x}^j}} = \begin{cases} 0 & \text{if } a \neq b \\ 1 & \text{if } a = b \end{cases}$$

**Polynomial of order $p > 1$**

Kernel gram matrix

$$K_{i,j} = (\langle \mathbf{x}^i, \mathbf{x}^j \rangle + c)^p$$

1st order partial-derivative matrix

$$D_{i,j}^a = p \left( \langle \mathbf{x}^i, \mathbf{x}^j \rangle + c \right)^{p-1} x_a^j$$

2nd order partial-derivative matrix

$$L_{i,j}^{ab} = \begin{cases} p(p-1)(\langle \mathbf{x}^i, \mathbf{x}^j \rangle + c)^{p-2} x_b^i x_a^j & \text{if } a \neq b \\ p(p-1)(\langle \mathbf{x}^i, \mathbf{x}^j \rangle + c)^{p-2} x_a^i x_a^j + p(\langle \mathbf{x}^i, \mathbf{x}^j \rangle + c)^{p-1} \\ & \text{if } a = b \end{cases}$$

**Gaussian kernel**

Kernel gram matrix

$$K_{i,j} = \exp\left( -\frac{\left\| \mathbf{x}^i - \mathbf{x}^j \right\|_2^2}{2\sigma^2} \right)$$

1st order partial-derivative matrix

$$D_{i,j}^a = \exp\left( -\frac{\left\| \mathbf{x}^i - \mathbf{x}^j \right\|_2^2}{2\sigma^2} \right) \frac{x_a^j - x_a^i}{\sigma^2}$$

2nd order partial-derivative matrix

$$
L_{i,j}^{ab} = \begin{cases} \exp\left(-\dfrac{\left\|\mathbf{x}^i - \mathbf{x}^j\right\|_2^2}{2\sigma^2}\right)\dfrac{(x_a^j - x_a^i)(x_b^i - x_b^j)}{\sigma^4} & \text{if } a \neq b \\[3ex] \exp\left(-\dfrac{\left\|\mathbf{x}^i - \mathbf{x}^j\right\|_2^2}{2\sigma^2}\right)\dfrac{(x_a^i - x_a^j)^2 - \sigma^2}{-\sigma^4} & \text{if } a = b \end{cases}
$$

# Chapter 5

# Large-scale Nonlinear Variable Selection via Kernel Random Features

**Chapter abstract:** We propose a new method for input variable selection in nonlinear regression. The method is embedded into a kernel regression machine that can model general nonlinear functions, not being a priori limited to additive models. This is the first kernel-based variable selection method applicable to large datasets. It sidesteps the typical poor scaling properties of kernel methods by mapping the inputs into a relatively low-dimensional space of random features. The algorithm discovers the variables relevant for the regression task together with learning the prediction model through learning the appropriate nonlinear random feature maps. We demonstrate the outstanding performance of our method on a set of large-scale synthetic and real datasets.

## 5.1 Introduction

It has been long appreciated in the machine learning community that learning sparse models can bring multiple benefits such as better interpretability, improved accuracy by reducing the curse of dimensionality, computational efficiency at prediction times, reduced costs for gathering and storing measurements, etc. A plethora of sparse learning methods has been proposed for linear models (Hastie, Tibshirani, and Wainwright 2015). However, developing similar methods in the nonlinear setting proves to be a challenging task.

Generalized additive models (Hastie and Tibshirani 1990) can use similar sparse techniques as their linear counterparts. However, the function class of linear combinations of nonlinear transformations is too limited to represent general nonlinear functions. Kernel methods (Schölkopf and Smola 2002) have for long been the workhorse of nonlinear modelling. Recently, a substantial effort has been invested into developing kernel methods with feature selection capabilities (Bolón-Canedo et al. 2013). The most successful approaches within the filter methods are based on mapping distributions into the reproducing kernel Hilbert spaces (RKHSs) (Muandet et al. 2017). Amongst the embedded methods, multiple algorithms use the feature-scaling weights proposed in Weston et al. (2003). The authors in Rosasco et al. (2013) follow an alternative strategy based on the function and kernel partial derivatives.

All the kernel-based approaches above suffer from a common problem: they do not scale well for large data sets. The kernel methods allow for nonlinear modelling by applying high dimensional (possibly infinite-dimensional) nonlinear transformations $\phi : \mathcal{X} \to \mathcal{H}$ to the input data. Due to what is known as the kernel trick, these transformations do not need to be explicitly evaluated. Instead, the kernel methods operate only over the inner products between pairs of data points that can be calculated quickly by the use of positive definite kernel functions $k : \mathcal{X} \times \mathcal{X} \to \mathbb{R}$, $k(\mathbf{x}, \bar{\mathbf{x}}) = \langle \phi(\mathbf{x}), \phi(\bar{\mathbf{x}}) \rangle$. Given that these inner products need to be calculated for all data-point pairs, the kernel methods are generally costly for datasets with a large number $n$ of training points both in terms of computation and memory. This is further exacerbated for the kernel variable selection methods, which typically need to perform the $\mathcal{O}(n^2)$ kernel evaluations multiple times (per each input dimension, or with each it-erative update).

In this work we propose a novel kernel-based method for input variable selection in nonlinear regression that can scale to datasets with large numbers of training points. The method builds on the idea of approximating the kernel evaluations by Fourier random features (Rahimi and Recht 2007). Instead of fixing the distributions generating the random features a priori, it learns them together with the predictive model such that they degenerate for the irrelevant input dimensions. The method falls into the category of embedded approaches that seek to improve predictive accuracy of the learned models through sparsity (Blum and Langley 1997). This is the first kernel-based variable selection method for general nonlinear functions that can scale to large datasets of tens of thousands of training data.

## 5.2 Background

We formulate the problem of nonlinear regression as follows: given a training set of $n$ input-output pairs $\mathcal{S}_n = \{(\mathbf{x}_i, y_i) \in (\mathcal{X} \times \mathcal{Y}) : \mathcal{X} \subseteq \mathbb{R}^d, \mathcal{Y} \subseteq \mathbb{R}, i \in \mathbb{N}_n\}$ sampled IID according to some unknown probability measure $\rho$, our task is to estimate the regression function $f : \mathcal{X} \to \mathcal{Y}$, $f(\mathbf{x}) = \mathrm{E}(y|\mathbf{x})$ that minimizes the expected squared error loss $\mathrm{L}(f) = \mathrm{E}\,(y - f(\mathbf{x}))^2 = \int (y - f(\mathbf{x}))^2 \,\mathrm{d}\rho(\mathbf{x}, y)$.

In the variable selection setting, we assume that the regression function does not depend on all the $d$ input variables. Instead, it depends only on a subset $\mathcal{I}$ of these of size $l < d$, so that $f(\mathbf{x}) = f(\bar{\mathbf{x}})$ if $x^s = \bar{x}^s$ for all dimensions $s \in \mathcal{I}$.

We follow the standard theory of regularised kernel learning and estimate the regression function as the solution to the following problem

$$\widehat{f} = \operatorname*{arg\,min}_{f \in \mathcal{H}} \widehat{\mathrm{L}}(f) + \lambda \|f\|_{\mathcal{H}}^2 \ . \tag{5.1}$$

Here the function hypothesis space $\mathcal{H}$ is a reproducing kernel Hilbert spaces (RKHSs), $\|f\|_{\mathcal{H}_K}$ is the norm induced by the inner product in that space, and $\widehat{\mathrm{L}}(f) = \frac{1}{n}\sum_i^n (y_i - f(\mathbf{x}_i))^2$ is the empirical loss replacing the intractable expected loss above.

From the standard properties of the RKHS, the classical result, e.g. Schölkopf and Smola (2002), states that the evaluation of the minimizing function $\widehat{f}$ at any point $\tilde{\mathbf{x}} \in \mathcal{X}$ can be represented as a linear combination of the kernel functions $k$ over the $n$ training points

$$\widehat{f}(\tilde{\mathbf{x}}) = \sum_{i}^{n} c_i \, k(\mathbf{x}_i, \tilde{\mathbf{x}}) \ . \tag{5.2}$$

The parameters $\mathbf{c}$ are obtained by solving the linear problem

$$(\mathbf{K} + \lambda \mathbf{I}_n)\mathbf{c} = \mathbf{y} \ , \tag{5.3}$$

where $\mathbf{K}$ is the $n \times n$ kernel matrix with the elements $K_{ij} = k(\mathbf{x}_i, \mathbf{x}_j)$ for all $\mathbf{x}_i, \mathbf{x}_j \in \mathcal{S}_n$.

### 5.2.1 Random Fourier features

Equations (5.2) and (5.3) point clearly to the scaling bottlenecks of the kernel regression. In principal, at training it needs to construct and keep in memory the $(n \times n)$ kernel matrix and solve an $n$ dimensional linear system ($\propto \mathcal{O}(n^3)$). Furthermore, the whole training set $\mathcal{S}_n$ needs to be stored and accessed at test time so that the predictions are of the order $\mathcal{O}(n)$.

To address these scaling issues, the authors in Rahimi and Recht (2007) proposed to map the data into a low-dimensional Euclidean space $\mathbf{z} : \mathcal{X} \to \mathbb{R}^D$ so that the inner products in this space are close approximations of the corresponding kernel evaluation $\langle \mathbf{z}(\mathbf{x}), \mathbf{z}(\tilde{\mathbf{x}}) \rangle_{\mathbb{R}^D} \approx \langle \phi(\mathbf{x}), \phi(\tilde{\mathbf{x}}) \rangle_{\mathcal{H}_K} = k(\mathbf{x}, \tilde{\mathbf{x}})$. Using the nonlinear features $\mathbf{z}(\mathbf{x}) \in \mathbb{R}^D$ the evaluations of the minimising function can be approximated by

$$\widehat{f}(\tilde{\mathbf{x}}) \approx \langle \mathbf{z}(\tilde{\mathbf{x}}), \mathbf{a} \rangle_{\mathbb{R}^D} \ , \tag{5.4}$$

where the coefficients $\mathbf{a}$ are obtained from solving the linear system

$$(\mathbf{Z}^T \mathbf{Z} + \lambda \mathbf{I}_D)\mathbf{a} = \mathbf{Z}^T \mathbf{y} \ , \tag{5.5}$$

where $\mathbf{Z}$ is the $(n \times D)$ matrix of the random features for all the data points. The above approximation requires the construction of the $\mathbf{Z}$ matrix and solving the $D$-dimensional linear problem, hence significantly reducing the training costs if $D \ll n$. Moreover, access to training points is no longer needed at test time and the predictions are of the order $\mathcal{O}(D) \ll \mathcal{O}(n)$.

To construct well-approximating features, the authors in Rahimi and Recht (2007) called upon Bochner's theorem which states that a continuous function $g : \mathbb{R}^d \to \mathbb{R}$ with $g(\mathbf{0}) = 1$ is positive definite if and only if it is a Fourier transform of some probability measure on $\mathbb{R}^d$. For translation-invariant positive definite kernels we thus have

$$k(\mathbf{x}, \tilde{\mathbf{x}}) = g(\mathbf{x} - \tilde{\mathbf{x}}) = g(\boldsymbol{\lambda}) = \int_{\mathbb{R}^d} e^{i \boldsymbol{\omega}^T \boldsymbol{\lambda}} \, \mathrm{d}\mu(\boldsymbol{\omega}) \ , \tag{5.6}$$

where $\mu(\boldsymbol{\omega})$ is the probability measure on $\mathbb{R}^d$. In the above, $g$ is the characteristic function of the multivariate random variable $\boldsymbol{\omega}$ defined by the expectation

$$g(\boldsymbol{\lambda}) = \mathrm{E}_{\boldsymbol{\omega}}(e^{i \boldsymbol{\omega}^T \boldsymbol{\lambda}}) = \mathrm{E}_{\boldsymbol{\omega}}(e^{i \boldsymbol{\omega}^T (\mathbf{x} - \tilde{\mathbf{x}})}) = \mathrm{E}_{\boldsymbol{\omega}}(e^{i \boldsymbol{\omega}^T \mathbf{x}} e^{-i \boldsymbol{\omega}^T \tilde{\mathbf{x}}}) = k(\mathbf{x}, \tilde{\mathbf{x}}) \ . \tag{5.7}$$

It is straightforward to show (see the appendix) that the expectation over the complex exponential can be decomposed into an expectation over an inner product

$$\mathrm{E}_{\boldsymbol{\omega}}(e^{i \boldsymbol{\omega}^T (\mathbf{x} - \tilde{\mathbf{x}})}) = \mathrm{E}_{\boldsymbol{\omega}} \langle \psi_{\boldsymbol{\omega}}(\mathbf{x}), \psi_{\boldsymbol{\omega}}(\tilde{\mathbf{x}}) \rangle \ , \tag{5.8}$$

where the nonlinear mappings are defined as

$$\psi_{\boldsymbol{\omega}} : \mathcal{X} \to \mathbb{R}^2, \quad \psi_{\boldsymbol{\omega}}(\mathbf{x}) = [\cos(\boldsymbol{\omega}^T \mathbf{x}), \sin(\boldsymbol{\omega}^T \mathbf{x})]^T \ . \tag{5.9}$$

In Rahimi and Recht (2007) the authors proposed an even lower-dimensional transformation

$$\varphi_{\boldsymbol{\omega}, b}(\mathbf{x}) : \mathcal{X} \to \mathbb{R}, \quad \varphi_{\boldsymbol{\omega}, b}(\mathbf{x}) = \sqrt{2} \cos(\boldsymbol{\omega}^T \mathbf{x} + b) \ , \tag{5.10}$$

where $b$ is sampled uniformly from $[0, 2\pi]$ and that satisfies the expectation equality

$$\mathrm{E}_{\boldsymbol{\omega}}(e^{i \boldsymbol{\omega}^T (\mathbf{x} - \tilde{\mathbf{x}})}) = \mathrm{E}_{\boldsymbol{\omega}, b} \langle \varphi_{\boldsymbol{\omega}, b}(\mathbf{x}), \varphi_{\boldsymbol{\omega}, b}(\tilde{\mathbf{x}}) \rangle \ . \tag{5.11}$$

We chose to work with the mapping $\varphi$ (dropping the subscripts $\boldsymbol{\omega}, b$ when there is no risk of confusion) in the remainder of the text. The approximating nonlinear feature $\mathbf{z}(\mathbf{x})$ for each data-point $\mathbf{x}$ is obtained by concatenating $D$ instances of the random mappings $\mathbf{z}(\mathbf{x}) = [\varphi^1(\mathbf{x}),\ldots,\varphi^D(\mathbf{x})]^T$ with $\boldsymbol{\omega}$ and $b$ sampled according to their probability distribution so that the expectation is approximated by the sample sum.

### 5.2.2 Variable Selection Methods

In this section we position our research with respect to other nonlinear methods for variable selection with an emphasis on kernel methods.

In the class of generalized additive models, lessons learned from the linear models can be reused to construct sparse linear combinations of the nonlinear functions of each variable or, taking into account also possible interactions, of all possible pairs, triplets, etc., e.g. Ravikumar et al. (2007), Yin et al. (2012), Lin and Zhang (2006), and Tyagi et al. (2016). Closely related to these are the multiple kernel learning methods that seek to learn a sparse linear combination of kernel bases, e.g. Bach (2008), Bach (2009), and Koltchinskii and Yuan (2010). While these methods have shown some encouraging results, their simplifying additive assumption and the fast increasing complexity when higher-order interactions shall be considered (potentially $2^d$ additive terms for $d$ input variables) clearly present a serious limitation.

Recognising these shortcomings, multiple alternative approaches for general nonlinear functions were explored in the literature. They can broadly be grouped into three categories (Blum and Langley 1997): filters, wrappers and embedded methods.

The filter methods consider the variable selection as a preprocessing step that is then followed by an independent algorithm for learning the predictive model. Many traditional methods based on information-theoretic or statistical measures of dependency (e.g. information gain, Fisher-score, etc.) fall into this category (Bolón-Canedo et al. 2015). More recently, significant advancement has been achieved in formulating criteria more appropriate for non-trivial nonlinear dependencies (Gretton et al. 2008; Song et al. 2007; Fukumizu and Leng 2012; Yamada et al. 2014; Ren et al. 2015; Chen et al. 2017). These are based on the use of (conditional) crosscovariance operators arising from embedding probability measures into the RKHS (Muandet et al. 2017). However, they are still largely disconnected from the predictive model learning procedure and oblivious of the effects the variable selection has on the final predictive performance.

The wrapper methods perform variable selection on top of the learning algorithm treating it as a black box. These are practical heuristics (such as greedy forward or backward elimination) for the search in the $2^d$ space of all possible subsets of the $d$ input variables (Kohavi and John 1997). Classical example in this category is the SVM with Recursive Feature Elimination (Guyon et al. 2002). The wrappers are universal methods that can be used on top of any learning algorithm but they can become expensive for large dimensionalities $d$, especially if the complexity of the underlying algorithm is high.

Finally, the embedded methods link the variable selection to the training of the predictive model with the view to achieve higher predictive accuracy stemming from the learned sparsity. Our method falls into this category. There are essentially just two branches of kernel-based methods here: methods based on feature rescaling (Grandvalet and Canu 2002; Weston et al. 2003; Rakotomamonjy 2003; Maldonado et al. 2011; Allen 2013), and derivative-based methods (Rosasco et al. 2013; Gregorová et al. 2018). We discuss the feature rescaling methods in more detail in section 5.3.2. The derivative based methods use regularizers over the partial derivatives of the function and exploit the derivative reproducing property (Zhou 2008) to arrive at an alternative finite-dimensional representation of the function. Though theoretically intriguing, these methods scale rather badly as in addition to the ($n \times n$) kernel matrix they construct also the ($nd \times n$) and ($nd \times nd$) matrices of first and second order partial kernel derivatives and use their concatenations to formulate the sparsity constrained optimisation problem.

There exist two other large groups of *sparse* nonlinear methods. These address the sparsity in either the latent feature representation, e.g. Gurram and Kwon (2014), or in the data instances, e.g. Chan et al. (2007). While their motivation partly overlaps with ours (control of overfitting, lower computational costs at prediction), their focus is on a different notion of sparsity that is out of the scope of our discussion.

## 5.3 Towards sparsity in input dimensions

As stated above, our objective in this chapter is learning a regression function that is sparse with respect to the input variables. Stated differently, the function shall be insensitive to the values of the inputs in the $d - l$ dimensional complement set $\mathcal{I}^c$ of the irrelevant dimensions so that $f(\mathbf{x}) = f(\bar{\mathbf{x}})$ if $x^s = \bar{x}^s$ for all $s \in \mathcal{I}$.

From equation (5.4) we observe that the function evaluation is a linear combination of the D random features $\varphi$. The random features (5.10) are in turn constructed from the input $\mathbf{x}$ through the inner product $\boldsymbol{\omega}^T \mathbf{x}$. Intuitively, if the function $\widehat{f}$ is insensitive to an input dimension $s$, the value of the corresponding input $x^s$ shall not enter the inner product $\boldsymbol{\omega}^T \mathbf{x}$ generating the D random features. Formally, we require $\omega^s x^s = 0$ for all $s \in \mathcal{I}^c$ which is obviously achieved by $\omega^s = 0$. We therefore identify the problem of learning sparse predictive models with sparsity in vectors $\boldsymbol{\omega}$.

### 5.3.1 Learning through random sampling

Though in equation (5.10) $\boldsymbol{\omega}$ appears as a parameter of the nonlinear transformation $\varphi$, it cannot be learned directly as it is the result of random sampling from the probability distribution $\mu(\boldsymbol{\omega})$. In order to ensure the same sparse pattern in the D random samples of $\boldsymbol{\omega}$, we use a procedure similar to what is known as the reparametrization trick in the context of variational auto-encoders (Kingma and Welling 2014).

We begin by expanding equation (5.6) of the Bochner's theorem into the marginals across the $d$ dimensions[1]

$$g(\boldsymbol{\lambda}) = \int_{\mathbb{R}^d} e^{i\boldsymbol{\omega}^T\boldsymbol{\lambda}} \, \mathrm{d}\mu(\boldsymbol{\omega}) = \int_{\mathbb{R}} e^{i\omega^1\lambda^1} \mathrm{d}\mu(\omega^1)\ldots \int_{\mathbb{R}} e^{i\omega^d\lambda^d} \mathrm{d}\mu(\omega^d) \ . \quad (5.12)$$

To ensure that $\omega^s = 0$ when $s \in \mathcal{I}^c$ in all the D random samples, the corresponding probability measure (distribution) $\mu(\omega^s)$ needs to degenerate to $\delta(\omega^s)$. The distribution $\delta(\omega^s)$ has all its mass concentrated at the point $\omega^s = 0$, and has the property $\int_{\mathcal{X}} h(\omega^s) \mathrm{d}\delta(\omega^s) = h(0)$. In particular for $h$ the complex exponential we have $\int_{\mathcal{X}} e^{i \omega^s \lambda^s} \, \mathrm{d}\delta(\omega^s) = 1$ so that the value of $\lambda^s$ has no impact on the product in equation (5.12), and therefore no impact on $g(\boldsymbol{\lambda})$.[2]

---

[1] This is possible due to the independence of the $d$ dimensions of the r.v. $\boldsymbol{\omega}$.

**Reparametrization trick**

To ensure that all the D random samples of $\boldsymbol{\omega}$ have the same sparse pattern we need to be able to optimise through its random sampling. For each element $\omega$ of the vector $\boldsymbol{\omega}$, we parametrize the sampling distributions $\mu_\gamma(\omega)$ by its scale $\gamma$ so that $\lim_{\gamma \to 0} \mu_\gamma(\omega) = \delta(\omega)$. We next express each of the univariate random variables $\omega$ as a deterministic transformation of the form $\omega = q_\gamma(\epsilon) = \gamma \epsilon$ (scaling) of an auxiliary random variable $\epsilon$ with a fixed probability distribution $\mu_1(\epsilon)$ with the scale parameter $\gamma = 1$. For example, for the Gaussian and Laplace kernels the auxiliary distribution $\mu_1(\epsilon)$ are the standard Gaussian and Cauchy respectively.

By the above reparametrization of the random variable $\boldsymbol{\omega}$ we disconnect the sampling operation over $\boldsymbol{\epsilon}$ from the rescaling operation $\boldsymbol{\omega} = \mathbf{q}_\gamma(\boldsymbol{\epsilon}) = \boldsymbol{\epsilon} \odot \boldsymbol{\gamma}$ with a deterministic parameter vector $\boldsymbol{\gamma}$. Sparsity in $\boldsymbol{\omega}$ (and therefore the learned model) can now be achieved by learning sparse parameter vector $\boldsymbol{\gamma}$.

Though in principle it would be possible to learn the sparsity in the sampled $\boldsymbol{\omega}$'s directly, this would mean sparsifying instead of one vector $\boldsymbol{\gamma}$ the D sampled vectors $\boldsymbol{\omega}$. Moreover, the procedure would need to cater for the additional constraint that all the samples have the same sparse pattern. While theoretically possible, we find our reparametrization approach more elegant and practical.

### 5.3.2 Link to feature scaling

In the previous section we have built our strategy for sparse learning using the inverse Fourier transform of the kernels and the degeneracy of the associated probability measures. When we plug the rescaling operation into the random feature mapping (5.10)

$$\varphi(\mathbf{x}) = \sqrt{2} \cos(\boldsymbol{\omega}^T \mathbf{x} + b) = \sqrt{2} \cos((\boldsymbol{\epsilon} \odot \boldsymbol{\gamma})^T \mathbf{x} + b) = \sqrt{2} \cos(\boldsymbol{\epsilon}^T (\boldsymbol{\gamma} \odot \mathbf{x}) + b) \ , \ \ (5.13)$$

---

[2]And from (5.6) and (5.4) it neither impacts the kernel and regression function evaluation.

we see that the parameters $\boldsymbol{\gamma}$ can be interpreted as weights scaling the input variables $\mathbf{x}$. This makes a link to the variable selection methods based on feature scaling. These are rather straightforward when the kernel is simply linear, or when the nonlinear transformations $\phi(\mathbf{x})$ can be evaluated explicitly (e.g. polynomial) (Grandvalet and Canu 2002; Weston et al. 2003). In essence, instead of applying the weights to the input features, they are applied to the associated model parameters and suitably constrained to approximate the zero-norm problem.

More complex kernels, for which the nonlinear features $\phi(\mathbf{x})$ cannot be directly evaluated (may be infinite dimensional), are considered in Rakotomamonjy (2003), Maldonado et al. (2011), and Allen (2013). Here the scaling is applied within the kernel function $k(\boldsymbol{\gamma} \odot \mathbf{x}, \boldsymbol{\gamma} \odot \tilde{\mathbf{x}})$. The methods typically apply a two-step iterative procedure: they fix the rescaling parameters $\boldsymbol{\gamma}$ and learn the corresponding $n$-long model parameters vector $\mathbf{c}$ (equation (5.2)); fix $\mathbf{c}$ and learn the $d$-long rescaling vector $\boldsymbol{\gamma}$ under some suitable zero-norm approximating constraint. The naive formulation for $\boldsymbol{\gamma}$ is a nonconvex problem that requires calculating derivatives of the kernel functions with respect to $\boldsymbol{\gamma}$ (which depending on the kernel function may become rather expensive). In Allen (2013), the author proposed a convex relaxation based on linearization of the kernel function. Nevertheless, all the existing methods applying the feature scaling within the kernel functions scale badly with the number of instances as they need to recalculate the $(n \times n)$ kernel matrix and solve the corresponding optimisation (typically $\mathcal{O}(n^3)$) with every update of the weights $\boldsymbol{\gamma}$.

## 5.4 Sparse random Fourier features algorithm

In this section we present our algorithm for learning with Sparse Random Fourier Features (SRFF).

Similarly to the feature scaling methods we propose a two-step alternative procedure to learn the model parameters $\mathbf{a}$ and the distribution scalings $\boldsymbol{\gamma}$. For a fixed $\boldsymbol{\gamma}$ we generate the random features for all the input training points $\mathcal{O}(nD)$, and solve the linear problem (5.5) $\mathcal{O}(D^3)$ to get the $D$-long model parameters $\mathbf{a}$. Given that in our large-sample settings we assume $D \ll n$, this

**Input** : training data $(\mathbf{X}, \mathbf{y})$; hyper-parameters $\lambda, D$, size of $\Delta$ simplex
**Output** : model parameters $\mathbf{a}$, scale vector $\boldsymbol{\gamma}$
**Initialise**: $\boldsymbol{\gamma}$ evenly on simplex $\Delta$, $\boldsymbol{\epsilon}_j \sim \mu_I(\boldsymbol{\epsilon})$ and $b_j \sim U[0, 2\pi]$, $\forall j \in \mathbb{N}_D$
**Objective**: $J(\mathbf{a}, \boldsymbol{\gamma}) = \|\mathbf{y} - \mathbf{Z}\mathbf{a}\|_2^2 + \lambda \|\mathbf{a}\|_2^2$

**repeat**                                                          // alternating descent
  **begin** Step 1: Solve for $\mathbf{a}$
    rescalings $\boldsymbol{\omega}_j = \boldsymbol{\gamma} \odot \boldsymbol{\epsilon}_j$, $\forall j \in \mathbb{N}_D$
    random features $\mathbf{z}(\mathbf{x}) = [\varphi^1(\mathbf{x}), \ldots, \varphi^D(\mathbf{x})]$, $\forall \mathbf{x} \in \mathcal{S}_n$       // equation (5.13)
    $\mathbf{a} \leftarrow \arg\min_a \|\mathbf{y} - \mathbf{Z}\mathbf{a}\|_2^2 + \lambda \|\mathbf{a}\|_2^2$                    // equation (5.5)
  **end**
  **begin** Step 2: Solve for $\boldsymbol{\gamma}$
    $\boldsymbol{\gamma} \leftarrow \arg\min_{\boldsymbol{\gamma} \in \Delta} \|\mathbf{y} - \mathbf{Z}\mathbf{a}\|_2^2$              // projected gradient descent
  **end**
**until** *objective convergence*;

**Algorithm 3:** Sparse Random Fourier Features (SRFF) algorithm

step is significantly cheaper than the corresponding step for learning the $\mathbf{c}$ parameters in the existing kernel feature scaling methods described in section 5.3.2.

In the second step, we fix the model parameters $\mathbf{a}$ and learn the $d$-long vector of the distribution scalings $\boldsymbol{\gamma}$. We formulate the optimisation problem as the minimisation of the empirical squared error loss with $\boldsymbol{\gamma}$ constrained on the probability simplex $\Delta$ to encourage the sparsity.

$$\arg\min_{\boldsymbol{\gamma} \in \Delta} J(\boldsymbol{\gamma}), \qquad J(\boldsymbol{\gamma}) := \|\mathbf{y} - \mathbf{Z}\mathbf{a}\|_2^2 \tag{5.14}$$

Here the $(n \times D)$ matrix $\mathbf{Z}$ is constructed by concatenating the D random features $\varphi$ with the $\boldsymbol{\gamma}$ rescaling (5.13).

We solve problem (5.14) by the projected gradient method with accelerated FISTA line search (Beck and Teboulle 2009). The gradient can be constructed from the partial derivatives as follows

$$\frac{\partial J(\boldsymbol{\gamma})}{\gamma^s} = -(\mathbf{y} - \mathbf{Z}\,\mathbf{a})^T \frac{\partial \mathbf{Z}}{\partial \gamma^s} \mathbf{a} \qquad \forall s \in \mathbb{N}_d$$

$$\frac{\partial Z_{ij}}{\partial \gamma^s} = -\sqrt{2}\sin(\boldsymbol{\epsilon}^T(\boldsymbol{\gamma} \odot \mathbf{x}) + b)\,\epsilon^s x^s, \qquad \epsilon^s = \omega_s/\gamma_s \;. \tag{5.15}$$

Unlike in the other kernel feature scaling methods, the form of the gradient (5.15) is always the same irrespective of the choice of the kernel. The particular kernel choice is reflected only in the probability distribution from which the auxiliary variable $\epsilon$ is sampled and has no impact on the gradient computations. In our implementation (https://bitbucket.org/dmmlgeneva/srff_pytorch), we leverage the automatic differentiation functionality of pytorch in order to obtain the gradient values directly from the objective formulation.

## 5.5 Empirical evaluation

We implemented our algorithm in pytorch and made it executable optionally on CPUs or GPUs. All of our experiments were conducted on GPUs (single p100). The code including the settings of our experiments amenable for replication is publicly available at https://bitbucket.org/dmmlgeneva/srff_pytorch.

In our empirical evaluation we compare to multiple baseline methods. We included the nonsparse random Fourier features method (RFF) of Rahimi and Recht (2007) in our main SRFF code as a call option. For the naive mean and ridge regression we use our own matlab implementation. For the linear lasso we use the matlab PASPAL package (Mosci et al. 2010). For the nonlinear Sparse Aditive Model (SPAM) (Ravikumar et al. 2007) we use the R implementation of (Zhao et al. 2014). For the Hilberth-Schmidt independece criterion lasso method (HSIC) (Yamada et al. 2014), and the derivative-based embedded method of (Rosasco et al. 2013) (Denovas) we use the authors' matlab implementation.

Except SPAM, all of the baseline sparse learning methods use a two step procedure for arriving at the final model. They first learn the sparsity using either predictive-model-dependent criteria (lasso, Denovas) or in a completely disconnected fashion (HSIC). In the second step (sometimes referred to as debiasing (Rosasco et al. 2013)), they use a base non-sparse learning method (ridge, or kernel ridge) to learn a model over the selected variables (including hyper-parameter search and cross-validation). For HSIC, which is a filter method that does not natively predict the regression outputs, we use as the second step our implementation of the RFF. It searches through the candidate sparsity patterns HSIC produces and uses the validation mean square error as a criteria for the final model selection. In contrast to these, our SRFF method is a *single step procedure* that does not necessitate this extra re-learning phase.

**Experimental protocol**

In all our experiments we use the same protocol. We randomly split the data into three independent subsets: train, validation and test. We use the train subset for training the models, we use the validation subset to perform the hyper-parameter search, and we use the test set to evaluate the predictive performance. We repeat all the experiments 30 times, each with a different random train/validation/test split.

We measure the predictive performance in terms of the root mean squared error (RMSE) over the test samples, averaged over the 30 random replications of the experiments. The regularization hyper-parameter $\lambda$ (exists in ridge, lasso, Denovas, HSIC, RFF and SRFF) is searched within a 50-long data-dependent grid (automatically established by the methods). The smoothing parameter in Denovas is fixed to 10 following the authors' default (Rosasco et al. 2013). We use the Gaussian kernel for all the experiments with the width $\sigma$ set as the median of the Euclidean distances amongst the 20 nearest neighbour instances. We use the same kernel in all the kernel methods and the corresponding scale parameter $\gamma = 1/\sigma$ in the random feature methods for comparability of results. We fix the number of random features to $D = 300$ for all the experiments in both RFF and SRFF.

We provide the results of the baseline nonlinear sparse methods (SPAM, HSIC, Denovas) only for the smallest experiments. As explained in the previous sections, the motivation for our work is to address the poor scaling properties of the existing methods. Indeed, none of the baseline kernel sparse methods scales to the problems we consider here. HSIC (Yamada et al. 2014) creates a $(n \times n)$ kernel matrix per each dimension $d$ and solves a linear lasso problem over the concatenated vectorization of these with memory requirements ($n^2 \times d$) and complexity $\mathcal{O}(n^4)$. In our tests, it did not finish within 24hrs running on 20 CPUs (Dual Core Intel Xeon E5-2680 v2 / 2.8GHz) for the smallest training size of 1000 instances in our SE3 experiment. Within the same time span it did not arrive at a solution for any of the experiments with $n > 1000$. Denovas constructs, stores in memory, and operates over the $(n \times n)$, $(nd \times n)$ and $(nd \times nd)$ kernel matrix and the matrices of the first and second order derivatives. In our tests the method finished with an out-of-memory error (with 32GB RAM) for the SE1 with 5k training samples and for SE2 problem already with 1k training instances. SPAM finished with errors for most of the real-data experiments.

### 5.5.1 Synthetic experiments

We begin our empirical evaluation by exploring the performance over a set of synthetic experiments. The purpose of these is to validate our method under controlled conditions when we understand the true sparsity of the generating model. We also experiment with various nonlinear functions and increasing data sizes in terms of both the sample numbers $n$ and the dimensionality $d$.

Table 5.5.1: Summary of synthetic experiments

| Exp code | Train size | Test size | Total dims | Relevant dims | Generative function |
|---|---|---|---|---|---|
| SE1 | 1k - 50k | 1k | 18 | 5 | $y = \sin\left((x_1 + x_3)^2\right)\sin(x_7 x_8 x_9) + N(0, 0.1)$ |
| SE2 | 1k - 50k | 1k | 100 | 5 | $y = \log\left((\sum_{s=11}^{15} x_s)^2\right) + N(0, 0.1)$ |
| SE3 | 1k - 50k | 10k | 1000 | 10 | $y = 10(z_1^2 + z_3^2)e^{-2(z_1^2 + z_3^2)} + N(0, 0.01)$ |

We use the same size for the test and validation samples. In all the experiments, the data instances are generated from a standard normal distribution. In the functions, subscripts are dimensions, superscripts are exponents. For more detailed description of the generative function of SE3 see the appropriate section in the text.

**SE1:**

The very first of our experiments is a rather small problem with only $d = 18$ input dimensions of which only 5 are relevant for the regression function. In Table 5.5.2 we compare our SRFF method to the baselines for the smallest sample setting with $n = 1000$. Most of the methods (linear, additive or non-sparse) do not succeed in learning a model for the complex nonlinear relationships between the inputs and outputs and fall back to predicting simple mean.

The general nonlinear models with sparsity (HSIC, Denovas and SRFF) divert from the simple mean prediction. They all discover and use in the predictive model the same sparse pattern (see Fig. 5.5.1 for SRFF). Denovas and SRFF achieve almost identical results which confirms that our method is competitive with the state of the art methods in terms of predictive accuracy and variable selection.[3]

Table 5.5.2: SE1 - Test RMSE for $n = 1000$

|  | Mean | Ridge | Lasso | RFF | SPAM | HSIC | Denovas | SRFF |
|---|---|---|---|---|---|---|---|---|
| RMSE | 0.287 | 0.287 | 0.287 | 0.287 | 0.0287 | 0.341 | **0.272** | **0.272** |
| std | 0.009 | 0.009 | 0.009 | 0.009 | 0.009 | 0.060 | 0.009 | 0.009 |

Predictive performance in terms of root mean squared error (RMSE) over independent test sets for the SE1 dataset with training size $n = 1000$. The std line is the standard deviation of the RMSE across the 30 resamples.

In Table 5.5.3 we document how the increasing training size contributes to improving the predictive performance even in the case of several thousands instances. The performance of the SRFF method for the largest 50k sample is by about 6% better than for the 1k training size. For the other methods the problem remains out of their reach[4] and they stick to the mean prediction even for higher training sizes.[5] We do not provide any comparisons with the nonlinear sparse methods because, as explained above, they do not scale to the sample sizes we consider here.

The improved predictive performance for the larger training sizes goes hand in hand with the variable selection, Figure 5.5.1. For the smallest 1k training sample, SRFF identifies only the 7th, 8th and 9th relevant dimensions. They enter the sine in the generative function in a product and therefore have a larger combined effect on the function outcome than the squared sum of dimensions 1 and 3. These two dimensions are picked up by the method from the larger training sets and this contributes to the increase in the predictive performance.

---

[3]The low predictive performance of HSIC is the result of the 2nd model fitting step. It could potentially be improved with an additional kernel learning step. However, as we keep the kernel fixed for all the other methods, we do not perform the kernel search for HSIC either.

[4]The class of linear functions is too limited and the nonlinear function with all the variables consid-

Table 5.5.3: SE1 - Test RMSE for increasing train size $n$

| $n$ | Mean | Ridge | Lasso | RFF | SRFF |
|-----|------|-------|-------|-----|------|
| 1k  | 0.287 (0.009) | 0.287 (0.009) | 0.287 (0.009) | 0.287 (0.009) | **0.272** (0.009) |
| 5k  | 0.284 (0.011) | 0.284 (0.011) | 0.284 (0.011) | 0.284 (0.011) | **0.263** (0.010) |
| 10k | 0.285 (0.010) | 0.285 (0.010) | 0.285 (0.010) | 0.286 (0.010) | **0.261** (0.011) |
| 50k | 0.283 (0.010) | 0.283 (0.010) | 0.283 (0.010) | 0.283 (0.010) | **0.255** (0.009) |

Predictive performance in terms of root mean squared error (RMSE) over independent test sets for the SE1 dataset with increasing training size $n$. The standard deviation of the RMSE across the 30 resamples is in the brackets.

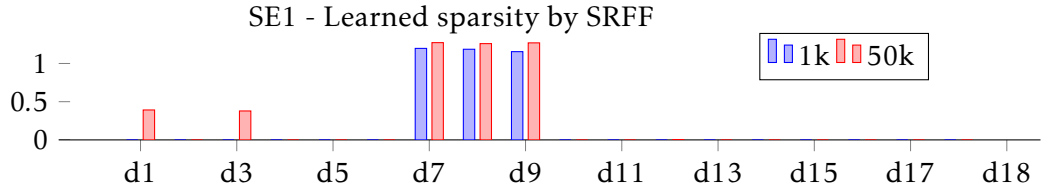

Figure 5.5.1: Learned sparsity pattern $\gamma$ by the SRFF method for the 1k and 50k training size in the SE1 experiment (the median of the 30 replications). The other nonlinear sparse methods learn the same pattern for the 1k problem but cannot solve the 50k problem.

**SE2:**

In the second experiment we increase the dimensionality to $d = 100$ and change the nonlinear function (see Table 5.5.1). The overall outcomes are rather similar to the SE1 experiment. Again, it's only the nonlinear sparse models that predict something else than mean, SPAM marginally better, HSIC marginally worse. Our SRFF method clearly outperforms all the other methods in the predictive accuracy. It also correctly discovers the 5 relevant variables with the median value of $\gamma$ for these dimensions between $0.92 - 1.04$ while the maximum for all the irrelevant variables is 0.06.[6] The advantage of SRFF over the baselines for large sample sizes (Table 5.5.5) is even more striking than in the SE1 experiment.

---

ered by RFF is too complex.

[5] The small variations in the error stem from using different training sets to estimate the mean.

[6] SPAM and HSIC discover the correct patterns as well but it does not help their predictive accuracy.

Table 5.5.4: SE2 - Test RMSE for $n = 1000$

|  | Mean | Ridge | Lasso | RFF | SPAM | HSIC | SRFF |
|---|---|---|---|---|---|---|---|
| RMSE | 2.216 | 2.216 | 2.216 | 2.216 | 2.162 | 2.357 | **1.603** |
| std | 0.105 | 0.105 | 0.105 | 0.104 | 0.110 | 0.141 | 0.104 |

Predictive performance in terms of root mean squared error (RMSE) over independent test sets for the SE2 dataset with training size $n = 1000$. The std line is the standard deviation of the RMSE across the 30 resamples.

Table 5.5.5: SE2 - Test RMSE for increasing train size $n$

| $n$ | Mean | Ridge | Lasso | RFF | SRFF |
|---|---|---|---|---|---|
| 1k | 2.216 (0.105) | 2.216 (0.105) | 2.216 (0.105) | 2.216 (0.105) | **1.603** (0.104) |
| 5k | 2.211 (0.079) | 2.211 (0.079) | 2.211 (0.079) | 2.211 (0.079) | **1.278** (0.076) |
| 10k | 2.224 (0.115) | 2.224 (0.115) | 2.224 (0.115) | 2.224 (0.115) | **1.272** (0.138) |
| 50k | 2.224 (0.082) | 2.224 (0.082) | 2.224 (0.082) | 2.224 (0.082) | **1.273** (0.080) |

Predictive performance in terms of root mean squared error (RMSE) over independent test sets for the SE2 dataset with increasing training size $n$. The standard deviation of the RMSE across the 30 resamples is in the brackets.

**SE3:**

In this final synthetic experiment we increase the dimensionality to $d = 1000$ to further stretch our SRFF method. There are only 10 relevant input variables in this problem. The first 5 were generated as random perturbations of the random variable $z_1$, e.g. $x_1 = z_1 + N(0, 0.1)$, the second 5 by the same procedure from $z_2$, e.g. $x_5 = z_2 + N(0, 0.1)$. The remaining 990 input variables were generated by the same process from the other 198 standard normal $z$'s.

We summarise the results for the 1k and 50k training instances in Table 5.5.6. As in the other synthetic experiments, the baseline methods are not able to capture the nonlinear relationships of this extremely sparse problem and instead predict a simple mean. Our SRFF method achieves significantly better accuracy for the 1k training set, and it further considerably improves with 50k samples to train on. These predictive gains are possible due to SRFF correctly discovering the set of relevant variables. In the 1k case, the medians across the 30 data resamples of the learned $\gamma$ parameters are between $0.37 - 0.71$ for the 10 relevant variables and maximally $0.05$ for the remaining 990 irrelevant variables. In the 50k case, the differences are even more clearly demarcated: $1.19 - 1.64$ for the relevant, and $0.03$ maximum for the irrelevant (bearing in mind that the total sum over the vector $\boldsymbol{\gamma}$ is the same in both cases).

Table 5.5.6: SE3 - Test RMSE for increasing train size $n$

| $n$ | Mean | Ridge | Lasso | RFF | SRFF |
|---|---|---|---|---|---|
| 1k | 0.676 (0.002) | 0.676 (0.002) | 0.676 (0.002) | 0.676 (0.002) | **0.478** (0.031) |
| 50k | 0.677 (0.002) | 0.677 (0.002) | 0.677 (0.002) | 0.677 (0.002) | **0.206** (0.004) |

Predictive performance in terms of root mean squared error (RMSE) over independent test sets for the SE3 dataset with increasing training size $n$. The standard deviation of the RMSE across the 30 resamples is in the brackets.

### 5.5.2 Real Data Experiments

Table 5.5.7: Summary of real-data experiments

| Data source | Dataset name | Exp code | Train size | Test size | Total dims |
|---|---|---|---|---|---|
| LIAC | Computer Activity | RCP | 6k | 1k | 21 |
| LIAC | F16 elevators | REL | 6k | 1k | 17 |
| LIAC | F16 ailernos | RAI | 11k | 1k | 39 |
| Kaggle | Ore mining impurity | RMN | 50k | 10k | 21 |

We use the same size for the test and validation samples.

We experiment on four real datasets: three from the LIACC[7] regression repository, and one Kaggle dataset[8]. The summary of these is presented in Table 5.5.8. The RFF results illustrate the advantage nonlinear modelling has over simple linear models. Our sparse nonlinear SRFF method clearly outperforms all the linear as well as the non-sparse nonlinear RFF method. Moreover, it is the only nonlinear sparse learning method that can handle problems of these large-scale datasets.

---

[7] http://www.dcc.fc.up.pt/~ltorgo/Regression/DataSets.html
[8] https://www.kaggle.com/edumagalhaes/quality-prediction-in-a-mining-process

Table 5.5.8: Real-data experiments - Test RMSE for increasing train size $n$

| $n$ | Mean | Ridge | Lasso | RFF | SRFF |
|-----|------|-------|-------|-----|------|
| RCP | 18.518 (0.988) | 9.686 (0.705) | 9.689 (0.711) | 8.194 (0.635) | **2.516** (0.184) |
| REL | 1.044 (0.050) | 0.514 (0.210) | 0.468 (0.178) | 0.446 (0.036) | **0.314** (0.032) |
| RAI | 1.013 (0.034) | 0.430 (0.018) | 0.430 (0.017) | 0.498 (0.038) | **0.407** (0.022) |
| RMN | 1.014 (0.006) | 0.987 (0.006) | 0.987 (0.006) | 0.856 (0.009) | **0.716** (0.008) |

Predictive performance in terms of root mean squared error (RMSE) over independent test sets for the real datasets. The standard deviation of the RMSE across the 30 resamples is in the brackets.

## 5.6 Summary and Conclusions

We present here a new kernel-based method for learning nonlinear regression function with relevant variable subset selection. The method is unique amongst the state of the art as it can scale to tens of thousands training instances, way beyond what any of the existing kernel-based methods can handle. For example, while none of the tested sparse method worked over datasets with more than 1k instances, the CPU version of our SRFF finished the *full* validation search over 50 hyper-parameters $\lambda$ in the 50k SE1 experiment within two hours on a laptop with a Dual Intel Core i3 (2nd Gen) 2350M / 2.3 GHz and 16GB RAM.

We focus here on nonlinear regression but the extension to classification problems is straightforward by replacing appropriately the objective loss function. We used the Gaussian kernel for our experiments as one of the most popular kernels in practice. But the principals hold for other shift-invariant kernels as well, and the method and the algorithm can be applied to them directly as soon as the corresponding probability measure $\mu(\boldsymbol{\omega})$ is recovered and the reparametrization of $\boldsymbol{\omega}$ explained in section 5.3.1 can be applied.

# Appendix

**Expectation of the complex exponential related to the positive definite kernel can be decomposed into an inner product as in equation (5.8) with the feature transformations**

A)

$$\psi_{\boldsymbol{\omega}}(\mathbf{x}) = [\cos(\boldsymbol{\omega}^T \mathbf{x}) \, \sin(\boldsymbol{\omega}^T \mathbf{x})]^T$$

B)

$$\varphi_{\boldsymbol{\omega},b}(\mathbf{x}) = \sqrt{2}\cos(\boldsymbol{\omega}^T \mathbf{x} + b)$$

*Proof part A.* Since the kernel function and the probability distribution are both real, the complex exponential $e^{i\boldsymbol{\omega}^T(\mathbf{x}-\bar{\mathbf{x}})}$ can be replaced by $\cos(\boldsymbol{\omega}^T(\mathbf{x}-\bar{\mathbf{x}})) = \cos(\boldsymbol{\omega}^T \mathbf{x})\cos(\boldsymbol{\omega}^T \bar{\mathbf{x}}) + \sin(\boldsymbol{\omega}^T \mathbf{x})\sin(\boldsymbol{\omega}^T \bar{\mathbf{x}})$ so that

$$\mathrm{E}_{\boldsymbol{\omega}}\Big(\cos(\boldsymbol{\omega}^T \mathbf{x})\cos(\boldsymbol{\omega}^T \bar{\mathbf{x}}) + \sin(\boldsymbol{\omega}^T \mathbf{x})\sin(\boldsymbol{\omega}^T \bar{\mathbf{x}})\Big) = \mathrm{E}_{\boldsymbol{\omega}}\left\langle \psi_{\boldsymbol{\omega}}(\mathbf{x}), \psi_{\boldsymbol{\omega}}(\bar{\mathbf{x}})\right\rangle = \mathrm{E}_{\boldsymbol{\omega}}(e^{i\boldsymbol{\omega}^T(\mathbf{x}-\bar{\mathbf{x}})}) \ .$$

$\square$

*Proof part B.* We first take the expectation with respect to $b$

$$\mathrm{E}_b\left\langle \varphi_{\boldsymbol{\omega}}(\mathbf{x}), \varphi_{\boldsymbol{\omega}}(\bar{\mathbf{x}})\right\rangle = \int_0^{2\pi} \sqrt{2}\cos(\boldsymbol{\omega}^T \mathbf{x} + b)\sqrt{2}\cos(\boldsymbol{\omega}^T \bar{\mathbf{x}} + b)\frac{1}{2\pi}\,\mathrm{d}b$$

$$= \frac{1}{\pi}\int_0^{2\pi} \cos(\boldsymbol{\omega}^T \mathbf{x} + b)\cos(\boldsymbol{\omega}^T \bar{\mathbf{x}} + b)\,\mathrm{d}b$$

$$= -\frac{1}{\pi}\cos(\boldsymbol{\omega}^T \mathbf{x} + b)\sin(\boldsymbol{\omega}^T \mathbf{x} + b)\Big|_0^{2\pi} + \frac{1}{\pi}\int_0^{2\pi} \sin(\boldsymbol{\omega}^T \mathbf{x} + b)\sin(\boldsymbol{\omega}^T \bar{\mathbf{x}} + b)\,\mathrm{d}b$$

$$= 0 + \frac{1}{\pi}\int_0^{2\pi} \cos(\boldsymbol{\omega}^T \mathbf{x} + b - \boldsymbol{\omega}^T \bar{\mathbf{x}} - b) - \cos(\boldsymbol{\omega}^T \mathbf{x} + b)\cos(\boldsymbol{\omega}^T \bar{\mathbf{x}} + b)\,\mathrm{d}b$$

$$(5.16)$$

The above used the integration by parts with

$$u = \cos(\boldsymbol{\omega}^T \mathbf{x} + b) \qquad\qquad \frac{dv}{db} = \cos(\boldsymbol{\omega}^T \bar{\mathbf{x}} + b)$$

$$\frac{du}{db} = -\sin(\boldsymbol{\omega}^T \mathbf{x} + b) \qquad\qquad v = \sin(\boldsymbol{\omega}^T \bar{\mathbf{x}} + b)$$

From (5.16) we have

$$\frac{2}{\pi} \int_0^{2\pi} \cos(\boldsymbol{\omega}^T \mathbf{x} + b)\cos(\boldsymbol{\omega}^T \bar{\mathbf{x}} + b)\,\mathrm{d}b = \frac{1}{\pi} \int_0^{2\pi} \cos(\boldsymbol{\omega}^T \mathbf{x} - \boldsymbol{\omega}^T \bar{\mathbf{x}})\,\mathrm{d}b$$

$$\frac{2}{\pi} \int_0^{2\pi} \cos(\boldsymbol{\omega}^T \mathbf{x} + b)\cos(\boldsymbol{\omega}^T \bar{\mathbf{x}} + b)\,\mathrm{d}b = \frac{2\pi}{\pi} \cos(\boldsymbol{\omega}^T (\mathbf{x} - \bar{\mathbf{x}}))$$

$$\frac{1}{\pi} \int_0^{2\pi} \cos(\boldsymbol{\omega}^T \mathbf{x} + b)\cos(\boldsymbol{\omega}^T \bar{\mathbf{x}} + b)\,\mathrm{d}b = \cos(\boldsymbol{\omega}^T (\mathbf{x} - \bar{\mathbf{x}}))$$

$$\mathrm{E}_b \big\langle \varphi_{\boldsymbol{\omega}}(\mathbf{x}), \varphi_{\boldsymbol{\omega}}(\bar{\mathbf{x}}) \big\rangle = \cos(\boldsymbol{\omega}^T (\mathbf{x} - \bar{\mathbf{x}}))$$

$$\mathrm{E}_{\boldsymbol{\omega},b} \big\langle \varphi_{\boldsymbol{\omega}}(\mathbf{x}), \varphi_{\boldsymbol{\omega}}(\bar{\mathbf{x}}) \big\rangle = \mathrm{E}_{\boldsymbol{\omega}} \cos(\boldsymbol{\omega}^T (\mathbf{x} - \bar{\mathbf{x}})) = \mathrm{E}_{\boldsymbol{\omega}}(e^{i\boldsymbol{\omega}^T (\mathbf{x} - \bar{\mathbf{x}})}) \qquad (5.17)$$

$\square$

# Chapter 6

# Final Thoughts

Throughout this thesis we have discussed several new methods for learning with sparsity and structured sparsity. We introduced the problems, we summarised existing approaches and their respective shortcomings, we explained the main ideas of the new methods and how these advance the state of the art, detailed the algorithms, tested them empirically and commented on the results. Driven to a large degree by the current pressure for positive results in research dissemination, we very much focused on the positive properties of the methods, on their advantages over their competitors, on their superiority documented on a set of controlled experiments.

We will now step back a little, away from this somewhat myopic view. There are many questions we asked ourselves during the development of the methods that we have not been able to answer and address satisfactorily. We take the liberty to discuss some of those now.

## 6.1   Granger modelling over non-stationary time series

The core assumption for classical time series analysis methods, including ours, is that the time series are stationary. The assumption is perfectly understandable from a statistical learning point of view. There is little hope to learn anything from the training data about the future data points unless we assume some stability in the generating process.

The difficulty in practical forecasting is that most of the series we get to work with do not seem stationary at all. Their mean or variance (or even higher moments, though these are rarely really considered) may be changing substantially with time. There may be jumps, step changes, fluctuations and other sorts of irregularities that complicate the analysis and make any forecasting efforts virtually impossible.[1]

When these changes are somehow smooth or regular, a standard approach is to take time into account as an explanatory variable and model for it explicitly (e.g. for trend and seasonality or via differencing). This is habitually done in practice and, in fact, often caters for most of the variation in the data.

If after removing the dependency on time the remainders of the series (the residuals) do not behave like a pure white noise but exhibit signs of autocorrelation, there is room for further modelling. This is where the autoregressive models (ARs) enter the game. Their goal is to capture the remaining variability in the data as well as possible in order to arrive at yet better forecasts.

Should the preprocessing steps change (e.g. use differencing instead of detrending, or fit the trend by a different function), the starting point and therefore the outcomes of the AR modelling will change accordingly. As a result, the AR contribution to the forecast accuracy will be different as will the performance of the whole modelling process over the original non-stationary time series.

From this perspective, treating the stationarizing transformations and the ensuing AR modelling as two disconnected steps is likely to degrade the overall quality of the forecasts. Instead, the two should be viewed together as a single composite in the model selection process. While this may be understood from the theoretical point of view, it is rarely followed in practice where the time dependency typically by large margin overweights the AR contribution to the total predictive accuracy. Most often therefore the AR modelling follows from an independently chosen set of transformations taking those for granted.

When discussing AR methods from a purely forecasting point of view, this may not seem too important. The goal is simply to come up with the best possible

---

[1]See for example Kuznetsov (2015) for recent efforts on establishing learning bounds based on measures of discrepancy between the pre- and post-forecasting distributions.

forecasts over the data presented to the predictive method. Whatever these may be.

However, for interpretation purposes, this is in our view far more critical. We claim in sections 2 and 3 that our methods are able to discover some structures in the underlying time series data. And as we illustrate on experiments, they indeed are. But are the structures we recover really relevant for the original time series or are these rather artefacts of the preprocessing transformations? What if we preprocessed the data somewhat differently? Would we discover the same or at least similar structures working over this different set of residuals? Under these circumstances, how shall the domain experts interpret the structural information exposed by the models? Can they rely on it when forming their understanding of the underlying generative processes?

As far as we can say, these questions tend to be either completely ignored or silently brushed aside in the current research on Granger causality discovery. Such an attitude, in our view, unfortunately greatly damages the credibility of the results and undermines the trustworthiness of the research.

The first step towards improvement is acknowledging these interpretational caveats. The next would be in finding answers to some of the questions raised above. Exploring them either empirically via stability studies and controlled simulated experiments or theoretically through establishing conditions for consistent inference of Granger causality graphs through the stationarizing transformations. Finally, the insights gained from the above analysis could be utilised for development of new methods for discovering Granger causality in non-stationary time series.

## 6.2 Derivative based regularisation

In chapter 4 we formulate regularisers using partial derivatives of the learned nonlinear function to promote sparsity and variable selection in the models. The intuition behind the derivative regularisers is based on the smoothness of the function and parallels the linear case. Obviously, this intuition holds irrespective of the function learning machine, be it kernel method or for example neural network, and goes beyond the particular case of variable selection. The derivative regularisation therefore naturally lends itself to an extension to a variety of learning problems.

A major drawback of the derivative regularisation in the kernel setting is its poor scalability. As we show in proposition 1 of chapter 4, the minimiser of a function learning problem containing the derivative regulariser can no longer be represented as a linear combination of the kernel sections centred at the $n$ training examples, but needs to be combined with additional $nd$ kernel derivative functions, where $d$ is the dimensionality of the problem. This means that to solve the learning problem we need to evaluate not only the first but also the second order derivatives of the kernel functions which brings along substantial computational as well as memory costs. Finding a way around these scaling limitations is therefore of utmost importance if the derivative regularisers shall be applied to any of the modern large or at least medium large learning problems.

Without having elaborated any details, we can speculate that an approach similar in spirit to the random approximations of the kernel functions (Rahimi and Recht 2007) might yield the necessary simplifications. This would entail finding suitable approximations for the kernel derivative functions and using those instead of the derivative terms in the regularisation.

Another idea worth considering is the random perturbations approach used in Mollaysa et al. (2017) for training a regularised neural network. There the similarity-promoting regulariser defined in terms of the function partial derivatives is replaced by a term operating over randomly perturbed data. This replacement, akin to finite difference approximation, substantially reduces the computational complexity of the algorithm. The gradient evaluations now involve only first order derivatives of the function as opposed to the second order derivatives stemming from the original formulation.

Links between data perturbation and regularisation have been shown before. For example, Bishop (1995) demonstrates that training a neural network with least squares or cross entropy error over a dataset with added random noise is equivalent to training over the original dataset with an extra Tikhonov regularisation term. The equivalence arises from the function Taylor approximation and is valid up to an error in the order of the random noise variance.

Building on the above ideas we believe that many problems with derivative regularisers should in principle be amenable to reformulations using data perturbations as finite approximations. These reformulations may be easier to handle by gradient based algorithms and therefore may be more suitable when working with realistically large datasets. To what extent this is possible? When could it be useful? What should the approximating terms look like? Should the perturbations or the random noise take some specific form? These all are questions that remain to be answered.

## 6.3 Extended interpretations of SRFF

The method for learning sparse models with Fourier random features (SRFF) we present in chapter 5 has two additional interpretations that help deeper understanding of the method and lead to ideas that may be worth following up in future work. The first of these is as a kernel learning method, the second as a strongly regularised shallow neural network.

**Kernel learning view**

We use the Gaussian kernel as an example to develop the ideas for the kernel learning view

$$k(\mathbf{x}, \mathbf{x}') = \exp\left(-\frac{\|\mathbf{x} - \mathbf{x}'\|_2^2}{2\sigma^2}\right) = \exp\left(-\frac{1}{2}(\mathbf{x} - \mathbf{x}')^T \boldsymbol{\Gamma}(\mathbf{x} - \mathbf{x}')\right), \quad \boldsymbol{\Gamma} = 1/\sigma^2 \mathbf{I} \ . \quad (6.1)$$

We generalize $\boldsymbol{\Gamma}$ to a diagonal matrix $\boldsymbol{\Gamma} = \mathrm{diag}(\boldsymbol{\gamma})$, $\gamma_i = 1/\sigma_i^2$

$$k(\mathbf{x}, \mathbf{x}') = \exp\left(-\frac{1}{2}(\mathbf{x} - \mathbf{x}')^T \boldsymbol{\Gamma}(\mathbf{x} - \mathbf{x}')\right) = \exp\left(-\sum_i^d \frac{(x_i - x_i')_2^2}{2\sigma_i^2}\right)$$

$$= \prod_i^d \exp\left(-\frac{(x_i - x_i')_2^2}{2\sigma_i^2}\right) = \prod_i^d k_i(x_i, x_i') \ , \quad (6.2)$$

where each $k_i : \mathcal{X}_i \times \mathcal{X}_i \to \mathcal{H}_{K_i}$ is the reproducing kernel over one dimension of the space $\mathcal{X} = \mathcal{X}_1 \times \ldots \times \mathcal{X}_d$, and $k : \mathcal{X} \times \mathcal{X} \to \mathcal{H}_K$ is the product kernel associated with the product reproducing kernel Hilbert space $\mathcal{H}_K = \mathcal{H}_{K_1} \otimes \ldots \otimes \mathcal{H}_{K_d}$.

From Bochner's theorem we have for each $k_i$

$$k_i(x_i, x_i') = \int_{\mathbb{R}} e^{i\,\omega_i^T(x_i - x_i')} \,\mathrm{d}\mu_i(\omega_i) \;, \tag{6.3}$$

where for the Gaussian kernel with scale parameter $\sigma_i^2$ the corresponding probability distribution $\mu_i$ is the Gaussian $N(0, \gamma_i = 1/\sigma_i^2)$. The probability distribution $\mu(\boldsymbol{\omega})$ corresponding to the product kernel (see also equation 5.12)

$$k(\mathbf{x}_i, \mathbf{x}_i') = \prod_i^d k_i(x_i, x_i') = \prod_i^d \int_{\mathbb{R}} e^{i\,\omega_i^T(x_i - x_i')}\,\mathrm{d}\mu_i(\omega_i) = \int_{\mathbb{R}^d} e^{i\,\boldsymbol{\omega}^T(\mathbf{x}_i - \mathbf{x}_i')}\,\mathrm{d}\mu(\boldsymbol{\omega}) \tag{6.4}$$

is the multivariate Gaussian $N(\mathbf{0}, \boldsymbol{\Gamma})$.

In chapter 5 we propose to learn the scale parameters $\gamma_i$ of the distributions $\mu_i(\omega_i)$. with the aim to recover a sparse model over a reduced set of input variables. From the relation $\gamma_i = 1/\sigma_i$ we observe that this is equivalent to learning the product kernel in equation 6.2.

Upon examination of the experimental results in section 5.5 we believe that a substantial contribution to the very good predictive performance of our SRFF method comes, in fact, from this *kernel learning* effect. In consequence, it appears that the methodology we propose in chapter 5 is beneficial even without the additional sparsity and variable selection goals.

When formulated solely as a kernel learning method, the non-smooth simplex constraint on the $\boldsymbol{\gamma}$ vector could be replaced by a smooth and convex $\ell_2$-ball constraint. This should lead to computational speed ups and potentially to further improvements in predictive performance, namely in cases where the analysed problem is not really sparse (in parallel to the non-sparse ridge vs sparse lasso formulations). Preliminary results of our initial investigation confirm these conjectures.

**Shallow network view**

Kernel methods can be seen as neural networks with one hidden layer. Unlike neural networks, however, the kernel methods do not learn the feature repre-sentation of the hidden layer. Instead, they fix it by choosing the kernel. The kernel feature space is typically higher dimensional than the original input space, sometimes even infinite dimensional[2]. The kernel machine network is thus shallow but may be very wide, potentially even infinitely wide.

The ability to learn the representation instead of coming up with it through some form of feature engineering is considered to be one of the main reasons for the network success story. At the same time, it has been shown on multiple occasions that randomization can boost the performance of neural networks, see e.g. surveys of Huang et al. (2015) and Zhang and Suganthan (2016). The idea of random Fourier features (further elaborated by Rahimi and Recht (2009) under the stage name "random kitchen sinks") replaces the optimisation in the first layer of the network to learn the hidden representation by randomization.

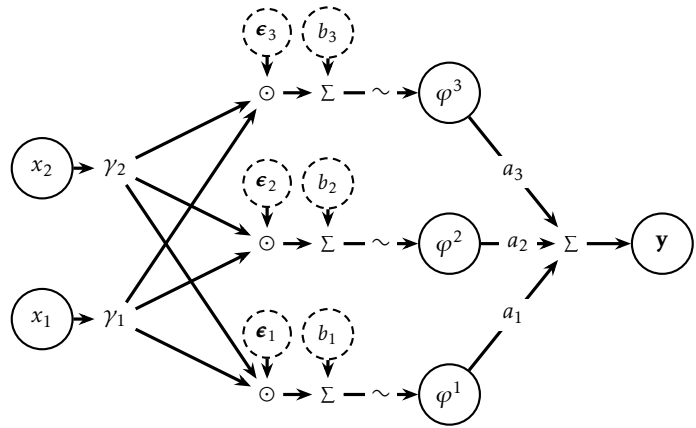

Figure 6.3.1: Network view of the SRFF method. The hidden features $\varphi$ are nonlinear functions (cosines) of the affine transformations of the inputs $\mathbf{x}$ with fixed random paramers $\boldsymbol{\epsilon}$ and $\mathbf{b}$ and learned parameters $\boldsymbol{\gamma}$. In reality, the size of the input and hidden layers is $d$ and $D$ respectively (using the notation of chapter 5).

Our SRFF method stands somewhere in between the complete optimisation approach of a standard neural network and the fully randomized approach of Rahimi and Recht (2009). As we show in figure 6.3.1, the hidden features $\varphi$ are functions of both the randomly fixed parameters $\boldsymbol{\epsilon}$ and $\mathbf{b}$ and the learned parameters $\gamma$. The feature representation in our case is therefore not fixed as in the other kernel methods, nor it is fully randomized. It is learned under very strong regularization. Out of the three parameters of the nonlinear feature maps, only one can be learnt and that with a further constraint of equality across neurons in the hidden layer and simplex constraint on the total size of the $\gamma$ vector.

---

[2]Though the true dimensionality of the hidden representations used in learning is upper bounded by the number of instances.

A more detailed analysis of the SRFF algorithm properties from the learning theory perspective under this shallow network view could bring some interesting insights not only relevant to the specific settings of our method but possibly also with a potential to contribute to the broader discussion of neural networks functioning in the vein of Poggio et al. (2017) or Belkin et al. (2018).

## 6.4 Other ideas

There are many more questions that remain unanswered and problems worth investigating. But this manuscript has to finish somewhere ...

# Index